\RequirePackage[svgnames]{xcolor}
\documentclass[11pt,letterpaper]{mystyle}
\usepackage[all]{hypcap}
\usepackage[svgnames]{xcolor}
\usepackage[comma,authoryear,compress]{natbib}
\usepackage{hyperref}[citecolor=lightblue]
\usepackage{amsmath}
\usepackage{etoolbox}

\hypersetup{
    colorlinks = true,
    citecolor = {YaleBlue},
}

\usepackage{algorithm}
\usepackage{algorithmicx}
\usepackage{algpseudocode}
\usepackage{microtype}
\usepackage{graphicx}
\expandafter\def\csname ver@subfig.sty\endcsname{}
\usepackage{booktabs} %
\usepackage{float}
\usepackage{bigstrut}
\usepackage{tikz}
\usetikzlibrary{positioning, arrows.meta, shapes.geometric}

\usepackage{comment}
\usepackage{amsmath}
\usepackage{amssymb}
\usepackage{mathtools}
\usepackage{amsthm}
\usepackage{mathrsfs}
\usepackage{nicefrac}
\usepackage{dsfont}
\usepackage{enumitem}
\usepackage{subcaption}
\usepackage{graphicx,subfig}
\usepackage{cleveref}
\usepackage{bxcoloremoji}
\usepackage{float}
\usepackage{algpseudocode}

\usepackage[utf8]{inputenc} %
\usepackage[T1]{fontenc}    %
\usepackage{url}            %
\usepackage{booktabs}       %
\usepackage{amsfonts}       %
\usepackage{nicefrac}       %
\usepackage{microtype}      %
\usepackage{graphicx}
\usepackage{amssymb}
\usepackage{wrapfig}
\usepackage{lipsum}
\usepackage{enumitem}
\usepackage{stackengine}
\usepackage[font=small,labelfont=bf]{caption}
\usepackage{color}
\usepackage{adjustbox}
\usepackage{physics}
\usepackage{siunitx}

\usepackage{rotating}
\usepackage{makecell}

\usepackage{multirow}

\newcommand{\x}{\mathbf{x}}

\definecolor{blanchedalmond}{rgb}{1.0, 0.92, 0.8}
\definecolor{carmine}{rgb}{0.59, 0.0, 0.09}
\definecolor{lightblue}{rgb}{0.22,0.45,0.70}%

\newtheorem{theorem}{Theorem}[section]

\newtheorem{definition}[theorem]{Definition}

\renewcommand{\mathbf}{\boldsymbol}

\makeatletter
\def\Ddots{\mathinner{\mkern1mu\raise\p@
\vbox{\kern7\p@\hbox{.}}\mkern2mu
\raise4\p@\hbox{.}\mkern2mu\raise7\p@\hbox{.}\mkern1mu}}
\makeatother

\definecolor{amaranth}{rgb}{0.9, 0.17, 0.31}
\definecolor{antiquebrass}{rgb}{0.8, 0.58, 0.46}
\definecolor{antiquefuchsia}{rgb}{0.57, 0.36, 0.51}
\definecolor{chromeyellow}{rgb}{0.31, 0.47, 0.26}

\newtcolorbox{AIbox}[2][]{aibox,title=#2,#1}
\definecolor{lightblue}{rgb}{0.22,0.45,0.70}%
\definecolor{Gray}{gray}{0.95}
\definecolor{Cornsilk}{rgb}{1.0, 0.97, 0.86}

\newcommand{\steplabel}[1]{\textnormal{\textbf{#1}}\;}
\usepackage{url}
\usepackage{xcolor}
\usepackage{pifont}
\usepackage{soul}
\tcbuselibrary{skins, breakable}

\usepackage{amsmath}

\newtcolorbox{cotquote}[1][]{%
  enhanced,
  breakable,
  colback=gray!5,           
  colframe=gray!50,         
  boxrule=0pt,              
  leftrule=2.5pt,           
  arc=1pt,
  left=8pt, right=6pt, top=4pt, bottom=4pt,
  fontupper=\itshape,       
  before skip=4pt, after skip=4pt,
  #1
}

\newtcolorbox{examplebox}{
    enhanced,
    breakable,             
    colframe=blue!60!black,  
    colback=blue!5,          
    boxrule=0.8pt,         
    arc=3pt,               
    before skip=10pt,      
    after skip=10pt,       
    left=5pt, right=5pt, top=5pt, bottom=5pt, 
    check odd page=true,   
    toggle enlargement=none
}

\newtcolorbox{humanbox}{
    colback=blue!5, 
    colframe=blue!75!black, 
    coltext=black, 
    boxsep=5pt, left=5mm, right=5mm, top=1mm, bottom=1mm
}

\newtcolorbox{R1box}{
    colback=green!5,         
    colframe=green!60!black, 
    coltext=black,           
    boxsep=5pt, 
    left=5mm, right=5mm, 
    top=1mm, bottom=1mm
}

\newtcolorbox{wrongbox}{
    colback=red!5,         
    colframe=red!60!black, 
    coltext=black,           
    boxsep=5pt, 
    left=5mm, right=5mm, 
    top=1mm, bottom=1mm
}

\newtcolorbox{questionbox}{
    colback=yellow!5,         
    colframe=yellow!60!black, 
    coltext=black,           
    boxsep=5pt, 
    left=5mm, right=5mm, 
    top=1mm, bottom=1mm
}

\newcounter{takeawaycounter}
\newtcolorbox{takeawaybox}[1][]{%
  enhanced,
  breakable,
  colback=blue!4,
  colframe=blue!55!black,
  boxrule=0.8pt,
  arc=2pt,
  left=12pt, right=10pt, top=8pt, bottom=8pt,
  fonttitle=\bfseries,
  before skip=8pt, after skip=8pt,
  attach boxed title to top left={xshift=10pt, yshift=-8pt},
  boxed title style={
    colback=blue!55!black,
    colframe=blue!55!black,
    boxrule=0pt,
    arc=2pt,
    left=6pt, right=6pt, top=2pt, bottom=2pt,
  },
  coltitle=white,
  before=\stepcounter{takeawaycounter},
  title=Takeaway \arabic{takeawaycounter},
  #1
}

\usepackage[all]{hypcap}

\usepackage{algpseudocode}
\usepackage{authblk}   
\usepackage{amssymb}      
\usepackage{bbm}          
\usepackage{optidef}  

\usepackage[utf8]{inputenc}
\usepackage[T1]{fontenc}

\usepackage{mathpazo}
\usepackage[svgnames]{xcolor}

\usepackage{lipsum}

\newtcolorbox{simpleElegantQuote}{
    colback=AliceBlue!50!White,   
    colframe=RoyalBlue!75!Black,  
    boxrule=0.5pt,                
    arc=2mm,                     
    boxsep=4pt,                   
    left=10pt, right=10pt,        
    top=8pt, bottom=8pt,         
    fontupper=\itshape,          
}

\title{
A Comprehensive Anatomy of Human and DeepSeek-R1 LLM Mathematical Reasoning
}

\runningtitle{A Comprehensive Anatomy of Human and DeepSeek-R1 LLM Mathematical Reasoning}

\author{Yuxiang Chen and Jun Wang\\
UCL Centre for Artificial Intelligence}

\correspondingauthor{Jun Wang: jun.wang@cs.ucl.ac.uk}

\begin{document}

\begin{abstract}

\textbf{Abstract}
\vspace{0.5em}

The emergence of ``Aha moments'' in large language models, particularly in DeepSeek-R1-0120, has raised the question of whether these systems genuinely reason or merely imitate the appearance of reasoning. To investigate this issue, we conduct a comprehensive empirical comparison between model and human reasoning across all 30 problems from AIME 2025. We exhaustively annotate 10,247 reasoning steps into five functional categories: Analysis, Inference, Branch, Backtrace, and Reflection, and systematically compare the model-generated reasoning traces with human reference solutions.
We find a clear structural difference between human and model reasoning. Human solutions maintain a compact alternation between analysis and deduction, whereas DeepSeek-R1 frequently revisits intermediate results, performs shallow (sometime unnecessary) verification, and loops through local checks without meaningful logical progress. We describe this behaviour as \emph{topological mimicry}: reproducing the surface form of reasoning without its functional role.
Despite the differences, our study identifies two main signals of genuine reasoning. First, successful traces exhibit stable use of branching and backtracking, while failed traces either underuse or overuse exploratory actions. Second, reflection is only effective when placed within deductive inference; reflections trapped in analysis loops often focus on local numerical details while missing global logical errors.
These findings suggest that current long-CoT models may be rewarded more for the appearance of reasoning than for genuine deductive progress. Based on our observations, we discuss several possible directions for improving both evaluation and training, including measuring cross-trace stability, penalising ``spinning-wheel'' traces, encouraging deeper logical correction, and reallocating inference-time compute toward deduction and backtracking. Overall, our analysis suggests that reasoning quality depends not simply on how much reflection occurs, but on whether reflection appears consistently and at the appropriate logical scale.
\end{abstract}

\maketitle

\begin{figure*}[h] 
    \centering
    \includegraphics[width=\textwidth]{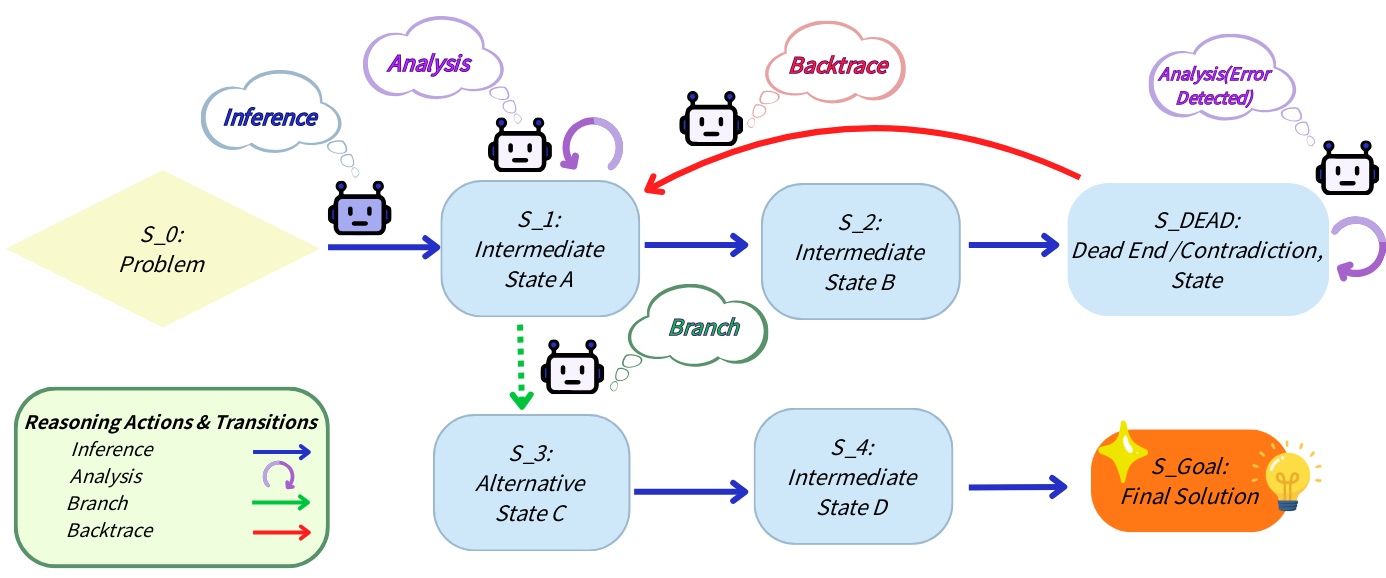}
    \caption{State CoT: A transition diagram illustrating the discrete reasoning states and meta-cognitive actions within a trajectory.}
    \label{fig:State_COT}
\end{figure*}

\section{Introduction}

\begin{figure}[h]
    \centering
    \begin{subfigure}[c]{0.48\textwidth}
        \centering
        \includegraphics[width=\textwidth]{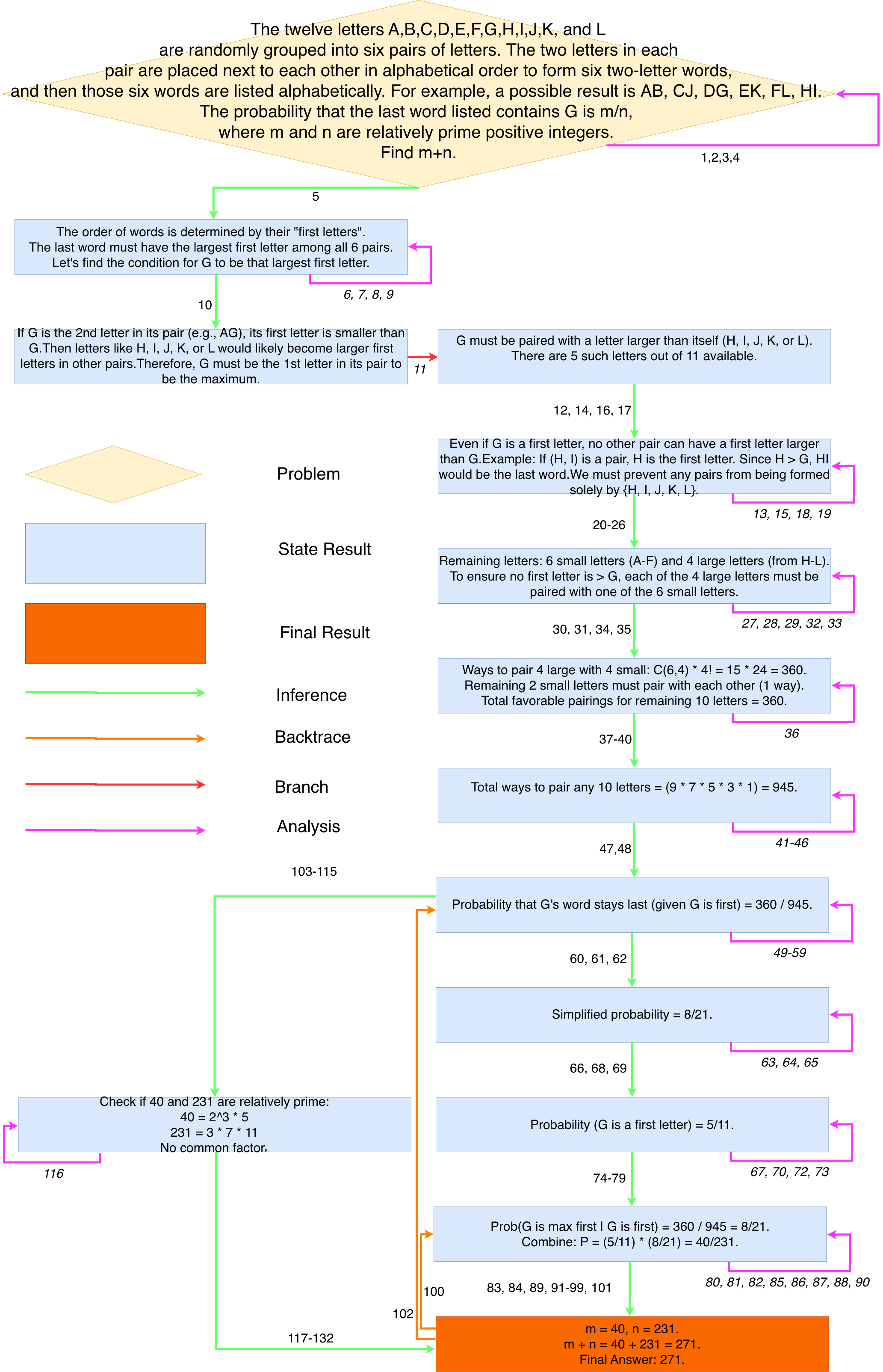}
        \caption{DeepSeek-R1's 132-step trace.}
        \label{fig:case_study_model}
    \end{subfigure}
    \hfill
    \begin{subfigure}[c]{0.48\textwidth}
        \centering
        \includegraphics[width=\textwidth]{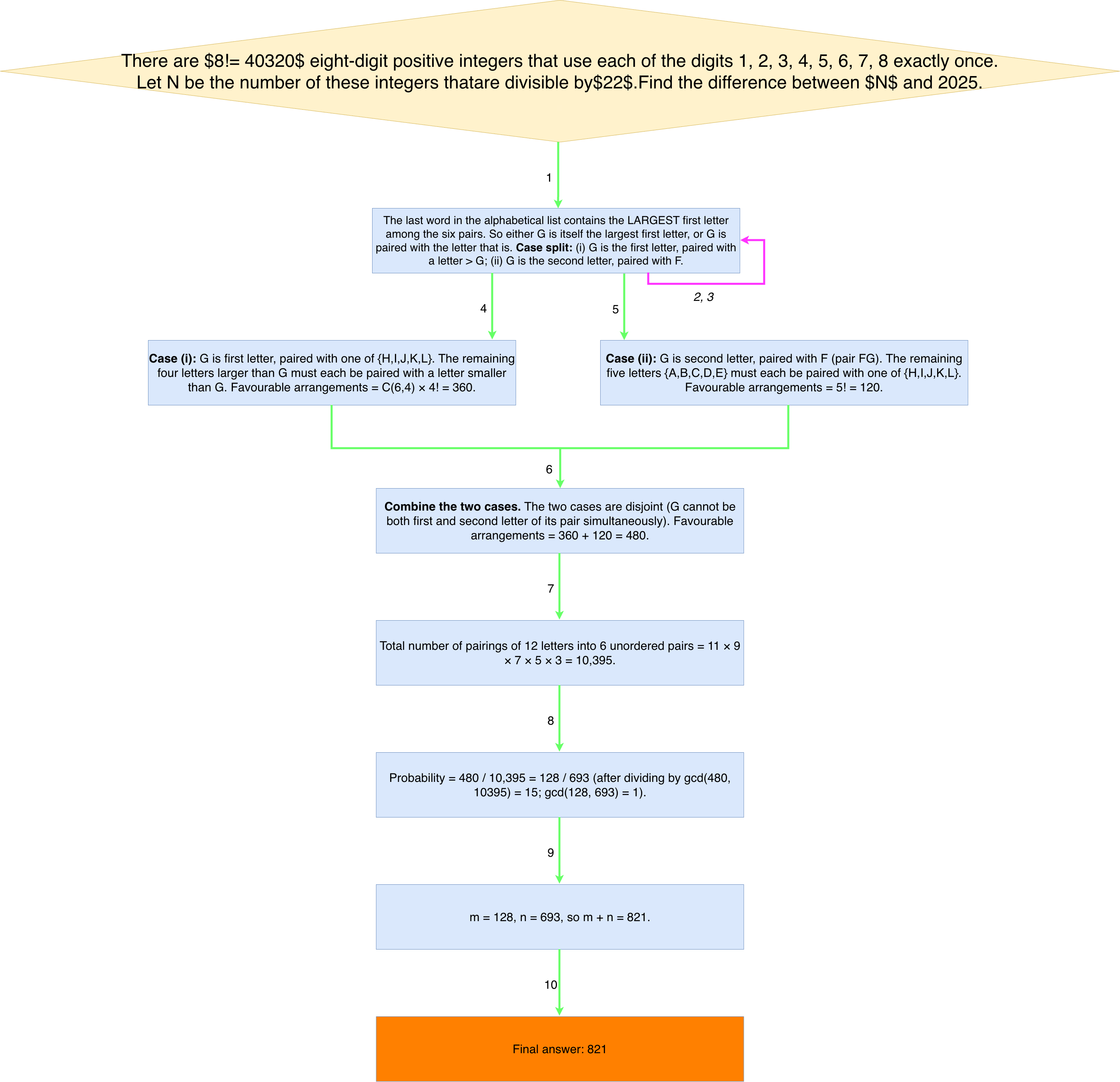}
        \caption{Human reference solution.}
        \label{fig:case_study_human}
    \end{subfigure}
    \caption{Logic flow graphs for AIME 1.7 (analysed in full in Section~\ref{sec:case_study}). \textbf{Left:} the logic flow extracted from DeepSeek-R1's 132-step trace, in which the same problem produces a large graph with multiple dead-end branches and shallow correction loops, and in which the second of the two cases needed for a correct answer is never reached. \textbf{Right:} the human reference solution, where each node is a single mathematical claim (a derived condition, a case decision, a computed quantity); the trace consists of roughly ten such nodes, organised as a brief shared preamble, a clean two-way case split, and a single convergence point. Colour coding: green = Inference, purple = Analysis, red = Branch, orange = Backtrace.}
    \label{fig:case_study_viz}
\end{figure}

Large Language Models (LLMs) have demonstrated remarkable progress in solving complex reasoning tasks, largely enabled by the Chain-of-Thought (CoT) paradigm~\citep{wei2022chain}. By generating explicit intermediate steps, CoT allows models to emulate human deductive reasoning, achieving impressive performance on mathematical and logical problems~\citep{kojima2022large, lewkowycz2022solving}. The evolution towards Long Chain-of-Thought has further pushed models to generate more detailed thinking processes, offering unprecedented opportunities to investigate the nature of machine inference~\citep{lightman2023lets}. In this context, \textbf{DeepSeek-R1-0120}, released in early 2025~\citep{deepseek-math}, has emerged as a representative open-source model capable of extended reasoning traces. The emergence of apparent ``Aha moments'' within these traces has renewed a critical debate: \textit{Does the model's elaborate reasoning structure reflect genuine logical correction, or is it merely ``parroting'' the form of reasoning without the substance, a phenomenon we term topological mimicry?}

A natural way to sharpen this question is to compare DeepSeek-R1's traces directly against the way a human solver approaches the same problem. The two are superficially alike: both produce extended written reasoning, both alternate between exploratory planning and concrete deduction, and both occasionally pause to verify or reconsider. The differences, however, are systematic. On the complete 30 problems of AIME 2025 studied in this paper, an experienced human solver typically devotes only about 15\% of steps to problem framing and planning, drives the solution forward through a lean rhythm of analysis and inference, backtracks rarely, and reserves explicit re-evaluation for the small number of points where the strategy genuinely branches. DeepSeek-R1 displays the same surface vocabulary of reasoning but distributes these activities very differently: roughly 40\% of its steps are spent in Analysis, intermediate conclusions are repeatedly re-derived rather than carried forward, and reflective interjections (``wait\dots'', ``let me check\dots'') are emitted frequently but predominantly as shallow arithmetic checks rather than as scrutiny of the underlying logic. The model's trace mimics the structural form of human reasoning whilst diverging from its functional content. Figure~\ref{fig:case_study_viz} makes this contrast concrete on a single AIME problem: the human reaches the answer in roughly ten load-bearing claims organised as a clean two-way case split, whereas DeepSeek-R1 produces a 132-step trace of the same overall shape that nonetheless never reaches the second case the proof requires, and terminates with the wrong answer. We dissect this example step by step in Section~\ref{sec:case_study}.

Existing evaluations of CoT reasoning remain limited in scope. Most studies measure only final-answer correctness, which overlooks the structural integrity and logical dependencies spanning an entire reasoning trajectory~\citep{turpin2023language, golowneko2024roscoe}. Recent surveys argue for moving beyond surface-level metrics to distinguish whether a reasoning chain is logically sound or structurally deceptive: failures from different underlying causes, such as calculation errors versus topological mimicry, are often conflated, hindering targeted improvements~\citep{lanham2023measuring}. Such coarse evaluation is particularly inadequate for benchmarks like the American Invitational Mathematics Examination (AIME)~\citep{hendrycksmath2021}, where problems demand sustained symbolic manipulation and rigorous case analysis rather than performative complexity.

In this work, we conduct a step-level empirical study of DeepSeek-R1 on AIME mathematics problems. We annotate every reasoning step into one of five functionally defined modes: \textbf{Analysis} (problem setup and planning, producing no new mathematical conclusion), \textbf{Inference} (deductive progression, producing a new equation, value, or logical claim along the current line), \textbf{Branch} (proposing a parallel solution path without abandoning the current one), \textbf{Backtrace} (explicitly retracting prior work and returning to an earlier state), and meta-cognitive \textbf{Reflection} (a triggered re-evaluation of the trace so far, signalled by markers such as ``wait'' or ``let me check''). Each mode is defined through verifiable textual signatures rather than latent representations, enabling direct human annotation by mathematics competition participants. This framework lets us address a central research question: \textit{How do the structural dynamics and control mechanisms within CoT differentiate genuine problem-solving from superficial imitation?}

Our analysis reveals that much of what appears to be self-reflection is better understood as \textbf{topological mimicry}, and uncovers a series of structural properties that distinguish genuine reasoning from imitation. To make these properties easy to locate and reuse, we condense each one into a numbered \emph{takeaway} placed at the end of the section that establishes it. These takeaways collectively identify two core properties that separate genuine reasoning from mimicry: the \textbf{cross-trace variance} of exploratory action (Section~\ref{sec:StructuralAnalysis}), where failed traces either barely use branching and backtracking or repeatedly overuse them; and the heavily context-dependent impact of reflection, which we frame as \textbf{contextual placement and functional targeting} (Sections~\ref{sec:Metacognition}--\ref{sec:SequentialDynamics}). A closing visual survey across all 30 problems (Section~\ref{sec:visual_analysis}) reinforces that surface complexity in a long Chain-of-Thought is not, on its own, evidence of reasoning quality.

Section~\ref{sec:discussion} draws out the implications of these findings for the \textbf{evaluation and training} of long-CoT reasoning models. We argue that evaluation must move beyond trace length and reflection density to include cross-trace variance and contextual placement. Furthermore, these structural distinctions translate into concrete training targets: leveraging backtrace depth as a quality filter for preference pairs to ensure deep correction ($\eta > 0.1$, Section~\ref{sec:case_study}), applying direct preference optimization (DPO) against ``spinning-wheel'' traces, using context-aware regularization of reflection triggers, and implementing inference-time compute reallocation. Together, these findings suggest that reasoning fidelity in LLMs depends not on how much reflection occurs, but on whether that reflection is applied consistently and at the correct logical scale.
\section{Related Works}

\subsection{Paradigms in LLM Reasoning}
Chain-of-Thought (CoT) prompting, introduced by~\citep{wei2022chain}, marked a significant milestone by demonstrating that eliciting intermediate steps substantially improves multi-step reasoning in large language models (LLMs). This foundational work has since inspired a variety of more complex reasoning paradigms. For example, self-consistency decoding improves robustness by sampling multiple reasoning paths and selecting the most frequent answer~\citep{wang2022self}. Research has also explored non-linear reasoning structures: Tree-of-Thought (ToT)~\citep{yao2023tree} prompting expands CoT into a tree-based search over possible solution steps, and Graph-of-Thought (GoT)~\citep{besta2024got} generalises this further by representing the reasoning process as a graph that can branch and merge paths. Beyond static prompting, new iterative frameworks enable models to refine their reasoning through feedback~\citep{bogdan2025thought}. Notably, Reflexion and Self-Refine methods prompt the model to critique and revise its own intermediate outputs, yielding improved solutions through multi-pass reasoning~\citep{shinn2023reflexion, madaan2023self, jacovi2024chain}.

However, as reasoning traces become more elaborate, distinguishing genuine reasoning from performative mimicry becomes critical. The predominant focus remains on final-answer correctness, which is a metric blind to the quality or logical validity of the underlying reasoning process~\citep{kim2024prometheus, liu2023evaluating}. This limitation is particularly acute in complex domains like advanced mathematics, where a correct answer might be reached through flawed logic (a ``false positive'')~\citep{uesato2022solving}. Recent studies have begun to address this gap. For instance, researchers advocate for more granular, process-oriented evaluation, showing that outcome-based rewards are unreliable when reasoning steps are incorrect~\citep{dziri2023faith}. In the same vein, efforts in process supervision (e.g., training reward models on chains of thought rather than answers) and Reinforcement Learning from AI Feedback (RLAIF) have emerged to explicitly reward the correctness of the reasoning process itself, not just the final outcome~\citep{lee2024rlaif}.

\subsection{Structural and Mechanistic Analysis of Reasoning}
Parallel to the development of new prompting paradigms, another line of research focuses on analysing the internal structure of reasoning traces themselves. This approach treats CoT outputs not as monolithic text blocks, but as structured artefacts that can be decomposed and functionally analysed. Early work in this direction modelled reasoning as formal program execution, allowing step-by-step verification of each inference~\citep{gao2023pal, chen2022program}. More recently, researchers have identified recurring patterns or ``motifs'' within reasoning chains that correlate with success or failure~\citep{huang2023large}. Other work has investigated the dynamics of self-correction: a model's ability to identify and fix its own errors during reasoning is a key component of robust problem-solving~\citep{welleck2022generating}. 

At a more fundamental level, the field of mechanistic interpretability seeks to understand how these functional reasoning behaviours are implemented within the transformer architecture itself~\citep{elhage2021mathematical, rauker2023toward}. Recent studies treat models as computational graphs to be reverse-engineered, revealing, for example, specialised attention heads that implement iterative ``scratchpad''-like computations or discovering interpretable circuits for navigating structured problem spaces~\citep{olsson2022context, stolfo2023mechanistic, wang2022interpretability, li2022emergent}. Such findings begin to bridge the gap between high-level reasoning strategies and the learned weights of the model.

In summary, prior analysis has largely bifurcated into either high-level prompting techniques (macro-structure) or low-level circuit mechanisms (micro-structure). A crucial mesoscopic level of analysis, focusing on the functional dynamics of reasoning steps within a discrete state space, is often missing. Our contribution fills this gap by introducing a functional taxonomy (Analysis, Inference, Branch, Backtrace, Reflection) grounded in observable textual features to systematically analyse the interplay between these modes. We bridge the gap between structure and performance by showing that it is the \textbf{strategic stability} and the \textbf{generative steering} of these modes that defines successful reasoning, providing a principled lens to separate genuine logical progression from topological mimicry.

\section{Methodology}
\label{sec:method}

\subsection{Conceptual Framework}
\label{sec:preliminaries}

To analyse the reasoning dynamics of DeepSeek-R1, we treat each Chain-of-Thought (CoT) trace as a sequence of functionally distinct steps, where each step is assigned to one of five categories based on observable textual features (Section~\ref{sec:taxonomy}). This step-level decomposition allows us to examine how reasoning unfolds over time without committing to any latent geometric representation: the categorisation is grounded entirely in surface-level evidence that human annotators can verify independently. Throughout the paper we adopt informal notions such as \emph{deductive progression}, \emph{strategic shift}, and \emph{trajectory} as intuitive descriptors of how reasoning evolves.

This framing enables us to identify a central failure mode that motivates the rest of the paper:

\begin{definition}[Topological Mimicry]
\label{def:mimicry}
A reasoning trace exhibits \emph{topological mimicry} when its surface-level structure, including length, mode composition, frequency of strategic shifts, and density of self-reflection, closely resembles that of successful traces, yet the trace makes little net progress towards the correct answer. The form of logical correction is reproduced without its function.
\end{definition}

The remainder of this section operationalises the five reasoning modes (Section~\ref{sec:taxonomy}) and details the annotation protocol (Section~\ref{sec:annotation}) used to label every step in our dataset.

\subsection{Operational Taxonomy of Reasoning Steps}
\label{sec:taxonomy}
 
We classify each reasoning step using two complementary layers. The first layer captures the \textbf{primary action} the model performs while generating the Chain-of-Thought trace: producing new content, committing to a sideways alternative, or undoing prior work. Four mutually exclusive primary modes, namely Analysis, Inference, Branch, and Backtrace, cover this layer. The second layer captures \textbf{meta-cognitive intervention}: textual signals such as ``wait'', ``hmm, is that right?'', or ``let me double-check'' that interrupt the current line of reasoning and redirect it. We call any such intervention a \emph{Reflection}. Reflection is not a separate primary action but an overlay: a single step is always assigned exactly one primary mode and may additionally carry one Reflection subtype, denoted by the combined label \texttt{Mode\_Ref\_X} (e.g., \texttt{Inf\_Ref\_L}).
 
This two-layer separation matters because Branch and Backtrace describe what the model is doing in the forward generation of the trace, whereas Reflection describes a triggered re-evaluation of what has just been generated. The two can co-occur. For example, a Logical reflection embedded within an Inference step is exactly what produces the \texttt{Inf\_Ref\_L} label.
 
Each mode and subtype below is defined through verifiable textual signatures, allowing direct human annotation. Short subscripted symbols are attached for concise reference in the N-gram analysis (Section~\ref{sec:SequentialDynamics}).
 
\subsubsection*{Primary Action Modes}
 
\paragraph{Analysis ($\mathcal{A}_{\text{ana}}$, Problem Setup and Planning)}
Steps that produce no new mathematical conclusion. The model restates or rephrases the problem, sets up notation or coordinates, lists known constraints, names a candidate strategy, or describes what it intends to do next. The output is preparatory: it organises information already given but does not derive a new equation, a new value, or a new logical claim. Markers commonly include phrases such as ``let me set up\dots'', ``so we need to find\dots'', ``one approach is to\dots'', and explicit case enumerations stated without yet being solved.
 
\begin{examplebox}
\textbf{Example (Problem Framing).}\quad
The problem asks for the area of triangle $BGE$, where six labelled points lie on a line at given pairwise distances and a seventh point $G$ lies off the line. The model's opening output is:
\begin{cotquote}
Okay, so I need to find the area of triangle $BGE$ given these points on a straight line and some distances. Let me start by visualising the problem. There are six points $A, B, C, D, E, F$ on a straight line in that order. Point $G$ is not on the line, so it forms a triangle with any three non-collinear points. The area we need is of triangle $BGE$.
\end{cotquote}
The step contains no new mathematical object: no coordinate is assigned, no equation is written, no length is computed. The model is restating the problem and visualising the configuration before any deduction begins. We label the step \texttt{Analysis}.
\end{examplebox}
 
\paragraph{Inference ($\mathcal{A}_{\text{inf}}$, Deductive Progression)}
Steps that produce a new mathematical fact along the current line of reasoning. The signature is the appearance of a new equation, a new numerical value, a new geometric relation, an applied theorem, or a discharged case: some object that did not exist before this step and that follows from prior content. Inference steps are what carry the trace forward toward the answer.
 
\begin{examplebox}
\textbf{Example (Algebraic Execution).}\quad
Continuing the same problem, the model has set up coordinates and used the conditions $CG = 40$ and $DG = 30$ to write two equations in the unknown horizontal coordinate $x$ of $G$:
\[
(x - 26)^2 + y^2 = 1600, \qquad (x - 40)^2 + y^2 = 900.
\]
It now eliminates $y^2$ by subtraction:
\begin{cotquote}
Subtract equation 2 from equation 1 to eliminate $y^2$: $[(x - 26)^2 - (x - 40)^2] = 1600 - 900$. Expand both squares: $(x^2 - 52x + 676) - (x^2 - 80x + 1600) = 700 \dots \implies 28x = 1624 \implies x = 58$.
\end{cotquote}
The step performs an explicit algebraic manipulation that yields a new numerical fact, the coordinate $x = 58$, which did not exist in the trace before. We label the step \texttt{Inference}.
\end{examplebox}
 
\paragraph{Branch ($\mathcal{A}_{\text{bra}}$, Strategic Diversification)}
Steps in which the model proposes a parallel solution path while leaving the current path intact. The intent is exploratory rather than corrective: the model has not declared the current approach wrong, but offers an additional method (a different formula, a different decomposition, a different parametrisation) to be pursued in parallel or as a cross-check. Marker phrases include ``alternatively'', ``another way to see this'', ``we could also\dots''.
 
\begin{examplebox}
\textbf{Example (Method Switch).}\quad
The problem asks for the number of subsets of size 8 from a row of 16 chairs that contain no three consecutive chairs. Just before this step, the model has derived a recurrence $f(n,k) = f(n-1,k) + f(n-2,k-1) + f(n-3,k-2)$ together with its base cases, and has decided to compute $f(16,8)$ by dynamic programming. It then writes:
\begin{cotquote}
Alternatively, maybe there's a generating function approach. Let me think.
\end{cotquote}
The step opens a parallel line of attack on generating functions but does not abandon the recurrence-plus-DP plan that was just established; that plan remains the working path and is in fact returned to a few steps later. The model is widening the search rather than retracting prior work. We label the step \texttt{Branch}.
\end{examplebox}

\paragraph{Backtrace ($\mathcal{A}_{\text{bac}}$, Strategic Reversal)}
Steps in which the model abandons a recently pursued path and explicitly returns the trace to an earlier point, typically a previously established intermediate result, a previously committed strategy, or a base case from which it can restart. The defining feature is the retraction itself: a side approach, a candidate computation, or a working assumption is dropped and the trace re-enters earlier content. The retraction is sometimes signalled with strong markers (``that is wrong'', ``I made a mistake at\dots'') but in practice is often implicit, surfacing only as a quiet return to prior territory after an exploration has failed to pay off. Unlike Branch, Backtrace removes a path from the working agenda rather than adding one.
 
\begin{examplebox}
\textbf{Example (Return to the Established Plan).}\quad
Continuing the chair-subset problem above, the model has by this point explored two alternative approaches (generating functions, then inclusion-exclusion) without committing to either, and the original DP plan still sits in the background. The next step reads:
\begin{cotquote}
Alternatively, using the recursion as we set up earlier.
\end{cotquote}
Despite the soft phrasing, the function of this step is a strategic reversal: the two recently proposed alternatives are dropped, and the trace returns to the recurrence-plus-DP plan that had been established earlier. The next step then begins concretely computing entries of the DP table. The model has not added a path. It has removed two and re-entered an older one. We label the step \texttt{Backtrace}.
\end{examplebox}
 
\subsubsection*{Reflection: Meta-Cognitive Triggers}
 
A \emph{Reflection} is a textual interruption in which the model momentarily steps out of the forward generation of the trace to evaluate what has just been produced. Linguistically, reflections are signalled by hedging or self-questioning markers such as ``wait'', ``hmm'', ``is that right?'', ``let me check'', and ``actually'', which briefly redirect the trace before it resumes. Reflections do not stand alone: every reflection co-occurs with a primary mode (Analysis, Inference, Branch, or Backtrace), yielding combined labels such as \texttt{Inf\_Ref\_L} or \texttt{Ana\_Ref\_N}. We distinguish four reflection subtypes by the scope of what the model is checking, ordered from the most local to the most strategically consequential. Intuitively, the subtypes can be read as redirecting the reasoning by progressively larger amounts: a Numerical check barely perturbs the direction, whereas a Logical reflection can reverse it entirely.
 
\begin{itemize}

    \item \textbf{Numerical ($\mathcal{R}_{\text{num}}$).} A localised arithmetic check. The model briefly pauses to recompute or verify a specific numerical result, such as a product, a sum, a division, or a modular reduction, without questioning the surrounding logic. The scope is one calculation, and the surrounding strategy is taken for granted.
    \begin{examplebox}
    \textbf{Example (Arithmetic Verification).}\quad
    Continuing the AIME 1.1 trace used in the previous examples, the model has just divided $1624 / 28$ by simplifying it as $406/7 = 58$, concluding $x = 58$. Before moving on to substitute this value back into the original system, it pauses:
    \begin{cotquote}
    Wait, $28 \times 58 = 1624$? Let's check: $28 \times 50 = 1400$, $28 \times 8 = 224$. $1400 + 224 = 1624$. Yes. So $x = 58$.
    \end{cotquote}
    The interrupt re-derives the multiplication $28 \times 58$ from a different decomposition ($28 \times (50 + 8)$) purely to confirm a single arithmetic fact. The strategy of substituting $x = 58$ back into equation 2 is not revisited; the equations themselves are not revisited; only the one product is checked. Because this is a localised arithmetic check sitting within a planning-level context (``Let me compute that''), the step is annotated \texttt{Ana\_Ref\_N}: an Analysis step carrying a Numerical reflection.
    \end{examplebox}
 
    \item \textbf{Formal ($\mathcal{R}_{\text{form}}$).} A check against the explicit output requirements of the problem rather than against the mathematics itself. AIME problems impose unusually strong output constraints. The answer must be an integer between $000$ and $999$, two named integers must be coprime, a length must be expressible as $m + n\sqrt{p}$ for specified types of $m, n, p$, and so on. The model occasionally interrupts itself to confirm that its current candidate satisfies these constraints. The underlying derivation is not re-examined; only the surface form of the result is checked.
    \begin{examplebox}
    \textbf{Example (Format Constraint Verification).}\quad
    After a chain of geometric computations, the model arrives at a candidate length and pauses before treating it as the final answer:
    \begin{cotquote}
    Wait, the problem requires the final answer to be an integer between $000$ and $999$, but my current result is $17\sqrt{3}$. This format is incorrect.
    \end{cotquote}
    The interrupt does not question how $17\sqrt{3}$ was derived. The geometric reasoning that produced it is left untouched. What the model is checking is whether the result fits the AIME output format, and the answer is no. The candidate is rejected on formal grounds alone. A step of this kind is annotated \texttt{Inf\_Ref\_F}: an Inference step (the geometric computation) carrying a Formal reflection (the format check).
    \end{examplebox}
 
    \item \textbf{Supplementary ($\mathcal{R}_{\text{sup}}$).} A sideways pause for stocktaking. Rather than verifying a single computation, the model takes inventory of \emph{multiple} previously derived facts at once: it lists what has been pinned down, what is still open, and how the pieces fit together, before resuming forward derivation. Unlike Numerical reflection, the scope is the global state of the trace rather than a single arithmetic operation, and the activity is consolidation rather than verification of a contested value.
    \begin{examplebox}
    \textbf{Example (Global State Assessment).}\quad
    Continuing the AIME 1.1 trace, the model has separately derived $f = 73$, $d = 40$, $b = 18$, $c = 26$ across several preceding steps using the constraints $AF$, $DF$, $BD$, $AC$. Before computing the remaining coordinate $e$, it interrupts the derivation to lay out everything obtained so far:
    \begin{cotquote}
    Now, we can check the positions so far: $A: 0$, $B: 18$, $C: 26$, $D: 40$, $E: ?$, $F: 73$.
    \end{cotquote}
    The interrupt does not recompute any individual value (in contrast to Numerical reflection, where one specific product would be re-derived) and does not check any of them against an output requirement (in contrast to Formal reflection). Instead, it surveys the global state of the derivation, marks the one coordinate ($E$) still missing, and gives the model a structured summary to resume from. A step of this kind is annotated \texttt{Ana\_Ref\_S}: an Analysis step carrying a Supplementary reflection.
    \end{examplebox}
 
    \item \textbf{Logical ($\mathcal{R}_{\text{log}}$).} A scrutiny of the foundational logic of the current path. The model questions whether an assumption is justified, whether a case split is exhaustive, whether the chosen strategy is in fact applicable, or whether two derived statements are actually consistent. Unlike the three preceding subtypes, Logical reflection targets the structural premises of the derivation rather than its surface arithmetic, output format, or progress state, and it can trigger a wholesale reversal of the trace's direction.
    \begin{examplebox}
    \textbf{Example (Case-Completeness Check).}\quad
    A different AIME problem asks for the number of integer pairs $(x, y) \in [-100, 100]^2$ satisfying $12x^2 - xy - 6y^2 = 0$. The model has factorised the equation into two lines, $y = -\tfrac{3}{2}x$ and $y = \tfrac{4}{3}x$, counted $67$ integer solutions on the first and $51$ on the second, and is about to add them. Just as it commits the sum, it interrupts itself:
    \begin{cotquote}
    Therefore, total solutions would be $67 + 51 = 118$. But wait, need to check if there's any overlap between the two cases. For example, is there a pair $(x,y)$ that satisfies both $3x + 2y = 0$ and $4x - 3y = 0$?
    \end{cotquote}
    The interrupt does not recompute $67$ or $51$, does not check the format of $118$, and does not summarise progress. What it questions is a structural premise of the derivation itself, namely whether the two cases the model has been counting are genuinely disjoint. The reflection turns out to be consequential: solving the joint system reveals the single overlap $(0, 0)$, and the final answer is corrected from $118$ to $117$. Without this Logical reflection, the entire derivation would have terminated at the wrong number despite every individual sub-count being correct. A step of this kind is annotated \texttt{Ana\_Ref\_L}: an Analysis step carrying a Logical reflection.
    \end{examplebox}
\end{itemize}
 
\paragraph{Notation summary.}
For convenience we collect the symbols used throughout the paper. We write $\mathcal{A} = \{\mathcal{A}_{\text{ana}}, \mathcal{A}_{\text{inf}}, \mathcal{A}_{\text{bra}}, \mathcal{A}_{\text{bac}}\}$ for the set of primary action modes and $\mathcal{R} = \{\mathcal{R}_{\text{num}}, \mathcal{R}_{\text{form}}, \mathcal{R}_{\text{sup}}, \mathcal{R}_{\text{log}}\}$ for the set of reflection subtypes. The combined label \texttt{Mode\_Ref\_X} (e.g., \texttt{Inf\_Ref\_L}) denotes a step whose primary mode is \texttt{Mode} and which simultaneously carries a reflection of subtype \texttt{X}. One additional quantity, the \emph{normalised jump amplitude} $\eta \in (0, 1]$, is introduced when needed in Section~\ref{sec:StructuralAnalysis}; it measures the depth of a Backtrace event relative to the trace.

\subsection{Experimental Design}

\subsubsection{Dataset}
We analysed 10\,247 reasoning steps from DeepSeek-R1 solving 30 AIME 2025 problems. AIME problems demand sustained symbolic manipulation, thereby eliciting rich reasoning dynamics whilst minimising guessing effects.

\subsubsection{Annotation Protocol}
\label{sec:annotation}

The 10\,247 reasoning steps across 30 AIME problems were annotated by five mathematics competition participants, including Mathematical Olympiad medallists. We chose to rely on expert human annotators rather than on automated rules or pretrained classifiers because, as we explain below, the distinction between several modes is fundamentally context-dependent and resists shallow pattern matching.

\paragraph{Why surface markers are not enough.}
Each mode in our taxonomy (Section~\ref{sec:taxonomy}) carries a set of typical linguistic signals such as ``Let me start by\dots'', ``Subtract equation 2 from equation 1\dots'', ``Alternatively\dots'', and ``Wait, that is wrong\dots'', but in practice these signals are at best partial cues, not decisive evidence. The same surface phrase can correspond to genuinely different functional roles depending on what surrounds it. ``Alternatively\dots'' opens both Branch steps (the model widens the search whilst keeping the current path live) and Backtrace steps (the model quietly retreats from a recent excursion to a previously established plan); the two are distinguishable only by whether the path being entered is being added to or removed from the working agenda. ``Wait\dots'' precedes Numerical reflections (a single arithmetic re-check), Formal reflections (a check against the problem's output requirements), Supplementary reflections (a stocktaking pause), and Logical reflections (a scrutiny of foundational assumptions); deciding which one applies requires understanding the scope of what the model has just chosen to interrogate. A keyword-based or LLM-based classifier acting on individual sentences misses this scope information by construction.

\paragraph{Why human annotators are necessary.}
Reliable labelling therefore requires three forms of judgement that, in our experience, only a domain-trained human reader can supply consistently: \emph{(i)} reading a small window of preceding steps to determine whether a proposed alternative is the model's current working path or a freshly added one (essential for Branch vs.\ Backtrace); \emph{(ii)} recognising the mathematical content under inspection in a reflection step, whether arithmetic, formatting, global progress, or strategic foundation, which itself requires understanding the problem (essential for the four reflection subtypes); and \emph{(iii)} adjudicating steps that blend modes, such as a single sentence that both performs a computation and questions whether the broader strategy is sound. AIME problems exacerbate all three difficulties: the reasoning is long, the strategies are non-trivial, and the same step can carry several functional roles at once. Our annotators were selected so that each could read DeepSeek-R1's output the way a competition coach reads a student's scratch work, tracking the live state of the argument rather than scanning for keywords.

\paragraph{Disagreement resolution.}
To quantify inter-annotator agreement, we performed cross-validation on two representative problems (734 steps), selected to span diverse reasoning patterns including combinatorial counting and geometric optimisation. All five annotators independently labelled every step in this subset following the criteria in Section~\ref{sec:taxonomy}. The exact match rate was 87.3\%, with Fleiss' $\kappa = 0.81$ (95\% CI [0.76, 0.85]), indicating strong agreement beyond chance despite the modest sample size and the genuine difficulty of the task.

The remaining 13\% disagreements clustered around the predictable hard cases described above: Branch versus Backtrace distinctions where the surrounding context was ambiguous, and steps blending calculation with planning. We resolved these through a pre-defined hierarchy: \texttt{Inference} was prioritised when explicit computation was present; \texttt{Analysis} served as the default otherwise; and Branch versus Backtrace ties were broken by whether the trace returned to that line within the next ten steps. This protocol was validated during cross-validation, where it reduced disagreement by 62\%.

\paragraph{Single-annotator coverage of the remaining problems.}
For the remaining 26 problems (9\,513 steps), each step was annotated by a single annotator assigned based on domain expertise (e.g., geometry specialists for geometry problems). We acknowledge that the cross-validation subset (734 steps) is modest relative to the full dataset. Nevertheless, the high agreement ($\kappa = 0.81$) on a strategically selected subset spanning distinct problem types provides evidence that our criteria (Section~\ref{sec:taxonomy}), once supported by trained human judgement, are sufficiently precise to yield consistent annotations across the complete dataset.
\section{Macro-Structural Analysis: From Aggregate Composition to Control Stability}
\label{sec:StructuralAnalysis}

In this section, we analyse empirical properties of reasoning traces under the operational taxonomy of Section~\ref{sec:method}. We adopt a progressive analytical strategy: moving from \textit{aggregate composition} (mode counts and transition frequencies averaged over a trace) to \textit{control stability} (the variance of exploratory actions across traces). By contrasting structural signatures of successful and unsuccessful traces, we aim to distinguish genuine logical progression from \textit{topological mimicry}, the failure mode introduced in Definition~\ref{def:mimicry}, in which failed traces replicate the surface structure of successful ones whilst failing to make corresponding progress towards the answer.

\subsection{The Illusion of Structure: Aggregate Mimicry}
\label{sec:AggregateFailure}

A prevailing hypothesis is that correct reasoning is characterised by a higher frequency of meta-cognitive interventions~\citep{shinn2023reflexion}, that is, more frequent application of \textit{Backtrace} or \textit{Branch} steps. We tested this by comparing \textit{Correct} and \textit{Incorrect} groups along two dimensions: how often each mode is occupied within a trace, and how the trace transitions between modes locally.

\begin{figure}[htbp]
    \centering
    \includegraphics[width=0.85\linewidth]{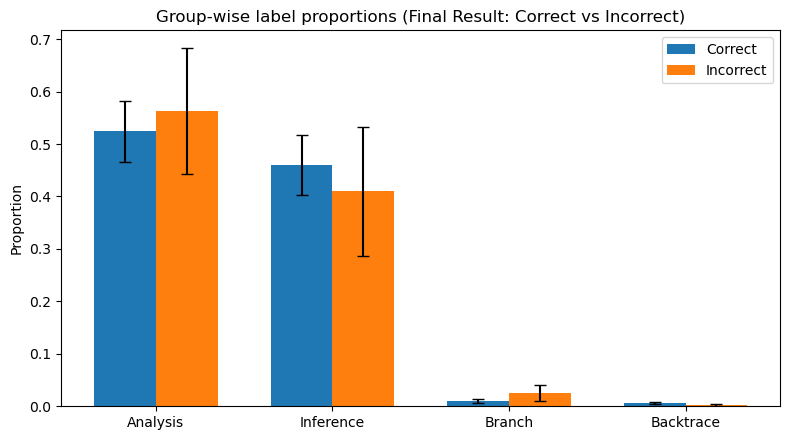}
    \caption{Mode occupancy ($\mathbf{P}$) for correct versus incorrect solutions. The lack of significant differences indicates that aggregate compositional structure is a poor predictor of success, which is consistent with \textbf{topological mimicry}.}
    \label{fig:group-proportions}
\end{figure}

\textbf{Indistinguishable mode occupancy.}
Contrary to expectations, our analysis returns a robust negative result at the aggregate level. As shown in Figure~\ref{fig:group-proportions}, Welch's $t$-tests detect no statistically significant difference in the mean occupancy of any mode (\textit{Analysis}, \textit{Inference}, \textit{Branch}, or \textit{Backtrace}) between the two groups (all $p > 0.3$). Crucially, the rate of \textit{Backtrace} steps is statistically identical across success and failure ($p \approx 0.33$), so the mere presence of strategic reversals does not predict whether the trace will reach the correct answer.

\textbf{Indistinguishable transition dynamics.}
Beyond mode occupancy, we examined the local dynamics of how traces move between modes. We considered two indicators: (1) the empirical mode-to-mode transition matrix, treated as a first-order Markov kernel; and (2) the Analysis--Inference alternation rate, which captures how tightly exploration and deduction are interleaved.

The results (Table~\ref{tab:dynamic_metrics}) again show striking similarity. The Jensen--Shannon divergence between transition matrices is negligible ($< 0.01$) for the high-frequency source modes, indicating that the local switching behaviour is the same regardless of outcome. Alternation rates ($0.278$ versus $0.227$) likewise show no statistically significant difference. These findings reinforce the topological-mimicry hypothesis: failed traces are \textit{structurally indistinguishable} from successful ones in length, mode composition, and transition behaviour. They reproduce the form of reasoning without its function.

\begin{table}[htbp]
    \centering
    \resizebox{0.95\linewidth}{!}{
    \begin{tabular}{lcccc}
        \toprule
        \multicolumn{5}{c}{\textbf{Part A: Mode-to-Mode Transitions (Markov Kernel)}} \\
        \midrule
        \textbf{Source State} & \textbf{JS Divergence} & \textbf{Counts (Correct)} & \textbf{Counts (Incorrect)} & \textbf{Interpretation} \\
        \midrule
        Analysis  & 0.0044 & 2346 & 1540 & Near-identical \\
        Inference & 0.0024 & 2032 & 1172 & Near-identical \\
        Branch    & 0.0085 & 55   & 63   & Similar \\
        Backtrace & 0.0425 & 18   & 9    & Similar \\
        \midrule
        \multicolumn{5}{c}{\textbf{Part B: Analysis--Inference Coupling}} \\
        \midrule
        \textbf{Metric} & \textbf{Correct Group} & \textbf{Incorrect Group} & \textbf{Diff} & \textbf{Significance} \\
        \midrule
        Alternation Rate  & $0.278 \pm 0.158$ & $0.227 \pm 0.123$ & +0.051 & Not Sig. \\
        Mean Run-Length   & $4.57 \pm 2.53$   & $5.25 \pm 2.86$   & -0.68  & Not Sig. \\
        \bottomrule
    \end{tabular}
    }
    \caption{Comparison of local dynamic metrics. The low JS divergence indicates that incorrect traces reproduce the transition dynamics of correct ones, consistent with \textbf{topological mimicry}.}
    \label{tab:dynamic_metrics}
\end{table}

\begin{figure}[htbp]
    \centering
    \includegraphics[width=0.85\linewidth]{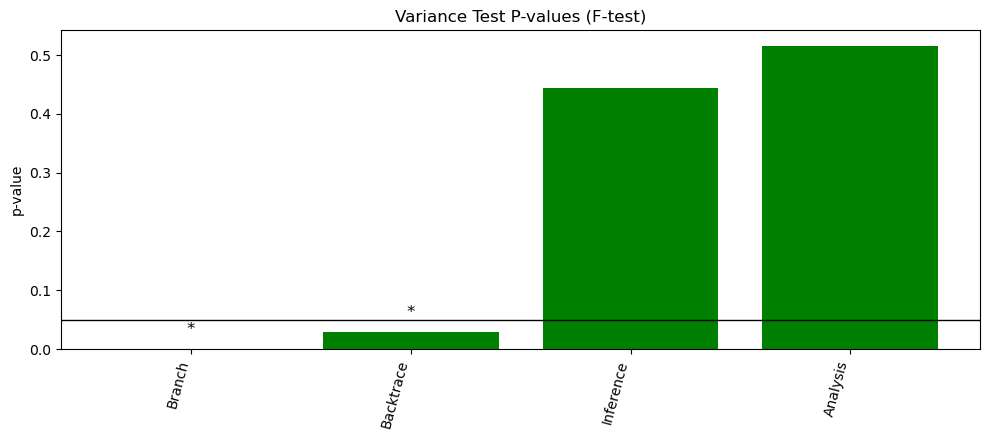}
    \caption{$p$-values from $F$-tests for equality of variances. Highly significant differences for \textit{Branch} and \textit{Backtrace} reveal that incorrect traces are marked by \textbf{control instability}: the rate at which the model deploys exploratory steps fluctuates erratically across traces.}
    \label{fig:f-test-plot}
\end{figure}

\subsection{The Reality of Instability: Control Chaos and Shallow Backtracking}
\label{sec:VarianceAndDepth}

Having established that aggregate composition is not predictive, we turn to a variance-level analysis of how consistently each control action is deployed across traces.

\textbf{High variance in exploratory steps reveals control chaos.}
We employed $F$-tests for equality of variances on the per-trace rates of each mode. The results (Figure~\ref{fig:f-test-plot}) show highly significant differences for the two exploratory modes. Successful traces apply \textit{Branch} and \textit{Backtrace} steps with consistent intensity from one trace to the next, whereas incorrect traces show significantly higher variance ($p < 0.001$ for Branch, $p = 0.030$ for Backtrace). This pattern is what we call \textit{control chaos}: in the absence of a stable policy for when to widen the search and when to retreat, the model's exploratory behaviour varies erratically between traces. Some traces are over-committed to a single line, whilst others are pulled apart by frequent strategy switches.

\textbf{The ``spinning wheel'' pathology.}
Within an individual failed trace, this instability surfaces as a pattern we informally call the \textit{spinning wheel} effect: frequent directional shifts (high Branch variance) without substantive forward progress. The model changes its proposed approach repeatedly but does not produce enough new deductive content for any of those approaches to advance the solution. The trace becomes stuck in a small region of the reasoning space, cycling between alternative proposals rather than committing to one and pushing it through.

\begin{figure}[htbp]
    \centering
    \includegraphics[width=0.85\linewidth]{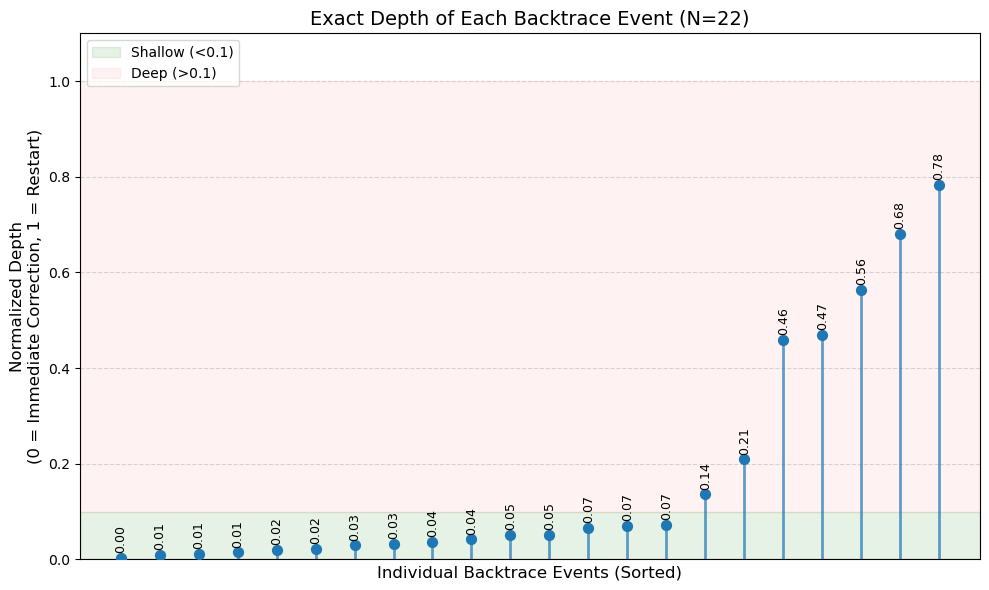}
    \caption{Distribution of the normalised jump amplitude $\eta$ defined in~\eqref{eq:eta}. The prevalence of shallow events ($\eta < 0.1$) indicates that the model primarily performs local one- or two-step rewinds rather than the deeper resets required to undo upstream errors.}
    \label{fig:backtrace-depth}
\end{figure}

\textbf{Shallow backtracking and the normalised jump amplitude $\eta$.}
The hypothesis of insufficient progress is corroborated by examining the depth of \textit{Backtrace} events. For each Backtrace step, we define the \emph{normalised jump amplitude}
\begin{equation}
    \eta \;=\; \frac{t - t_{\mathrm{back}}}{T} \;\in\; (0, 1],
    \label{eq:eta}
\end{equation}
where $t$ is the index of the Backtrace step, $t_{\mathrm{back}}$ is the earliest step to which the trace returns, and $T$ is the total length of the trace. Intuitively, $\eta$ measures how far back, as a fraction of the trace, a Backtrace event reaches: $\eta$ close to $0$ means the model retreats only one or two steps, whilst $\eta$ close to $1$ means it returns essentially to the beginning of the trace. Figure~\ref{fig:backtrace-depth} shows that the overwhelming majority of Backtrace events in failed traces have $\eta < 0.1$. Most self-corrections are therefore \textbf{local}: the model rewinds a step or two to re-check arithmetic, but does not perform the deeper resets required to escape an incorrect premise established further upstream. It changes direction often, but it does not move far back, producing a cycle of superficial reflection without functional correction.

\begin{takeawaybox}
The aggregate composition of a long Chain-of-Thought trace, including mode occupancy, transition frequencies, and mean reflection rates, does not separate correct from incorrect reasoning in our data; the cross-trace variance of exploratory action does. Successful traces deploy Branch and Backtrace at a stable rate from one problem to the next ($p < 0.001$ for Branch, $p = 0.030$ for Backtrace); failed traces alternate between barely deploying these actions and deploying them in bursts. The signal is in the variance, not in the mean.
\end{takeawaybox}
\section{Micro-Structural Analysis: The Functional Role of Meta-Cognitive Reflection}
\label{sec:Metacognition}

Whilst Section~\ref{sec:StructuralAnalysis} established that the aggregate structure of incorrect traces is indistinguishable from that of correct ones, the macro-level expression of topological mimicry, it leaves a fundamental question unanswered: \textit{why} does the model devote a substantial fraction of its steps to reflection without converging on a correct solution?

In this section we move from the trace level to the step level, examining the contextual placement and functional role of reflection. We re-summarise the four reflection subtypes operationalised in Section~\ref{sec:taxonomy}, characterise where in a trace each subtype tends to appear, and document a recurring failure mode we call \textit{performative reflection}, in which the model produces the linguistic surface of self-correction without any of its corrective effect.

\subsection{Reflection Subtypes Revisited}
\label{sec:ReflectionTaxonomy}

Section~\ref{sec:taxonomy} introduced four reflection subtypes; here we briefly recall what each one operationally checks, since this distinction underpins the analysis that follows. \textit{Numerical reflection} ($\mathcal{R}_{\text{num}}$) re-derives a single arithmetic fact, such as a product, a sum, or a modular reduction. \textit{Formal reflection} ($\mathcal{R}_{\text{form}}$) checks the candidate result against an explicit output requirement of the problem (an integer in $[000,999]$, a coprimality constraint, a prescribed surd form). \textit{Supplementary reflection} ($\mathcal{R}_{\text{sup}}$) is a stocktaking pause in which the model lays out multiple already-derived facts at once to consolidate the global state of the trace. \textit{Logical reflection} ($\mathcal{R}_{\text{log}}$) scrutinises the structural premises of the derivation, including an assumption, a case split, or the applicability of the chosen strategy, and is the only subtype capable of triggering a wholesale change of direction. Reflection co-occurs with a primary mode rather than replacing it; for example, a step labelled \texttt{Inf\_Ref\_L} carries an Inference action together with a Logical reflection.

\subsection{Contextual Asymmetry: Reflection Embedded in Planning}
\label{sec:ReflectionDistribution}

A common assumption is that reflection acts as a post-hoc verification mechanism applied after a deductive step has been taken~\citep{turpin2023language}. Our data tells a different story: in DeepSeek-R1's traces, reflection is overwhelmingly embedded inside planning steps rather than deductive ones.

Table~\ref{tab:reflection_distribution} shows the distribution. Across all traces, \textbf{78.8\%} of reflection instances co-occur with an \textit{Analysis} primary mode and only 18.0\% with an \textit{Inference} primary mode. The bias holds for Numerical (80.7\%), Logical (82.1\%) and Supplementary (80.4\%) subtypes alike. The single exception is Formal reflection, of which 31.5\% sits within Inference steps, typically when an intermediate computation has just produced a value that violates the problem's output constraints (a non-integer where an integer is required, a negative quantity where a positive one is required) and the model immediately flags the violation. The overall picture is that DeepSeek-R1's reflection apparatus is loaded mostly during planning, not during deduction: the model thinks twice when it is preparing to compute, and rarely when it is actually computing.

\begin{table}[htbp]
    \centering
    \resizebox{0.8\linewidth}{!}{
    \begin{tabular}{lccc}
        \toprule
        \textbf{Reflection Type} & \textbf{in Analysis} & \textbf{in Inference} & \textbf{in Branch/Backtrace} \\
        \midrule
        Numerical ($\mathcal{R}_{\text{num}}$) & \textbf{80.7\%} & 13.2\% & 6.1\% \\
        Logical ($\mathcal{R}_{\text{log}}$)    & \textbf{82.1\%} & 16.0\% & 1.9\% \\
        Formal ($\mathcal{R}_{\text{form}}$)    & 68.5\% & \textbf{31.5\%} & 0.0\% \\
        Supplementary ($\mathcal{R}_{\text{sup}}$) & \textbf{80.4\%} & 9.8\% & 9.8\% \\
        \midrule
        \textbf{Overall Weighted} & \textbf{78.8\%} & 18.0\% & 3.2\% \\
        \bottomrule
    \end{tabular}
    }
    \caption{Conditional distribution of reflection subtypes across primary modes. Reflection in DeepSeek-R1 is predominantly embedded within Analysis (planning) steps rather than Inference (deduction) steps.}
    \label{tab:reflection_distribution}
\end{table}

\subsection{Density and Stability are Two Different Things}
\label{sec:ReflectionStats}

Does higher reflection density correlate with success? We tested whether either the mean per-trace rate or the cross-trace stability of each reflection subtype distinguishes correct from incorrect solutions. The two answers turn out to be very different (Table~\ref{tab:reflection_stats}).

On mean density, Welch's $t$-tests show no statistically significant difference for any subtype between correct and incorrect solutions (all $p > 0.29$). Reflecting more often does not, on its own, make the model more likely to succeed. On variance across traces, however, $F$-tests find significant differences for three of the four subtypes: Numerical ($p = 0.040$), Formal ($p = 0.0002$) and Supplementary ($p = 0.0013$). The disparity is most extreme for Formal reflection, whose cross-trace variance in failed traces ($\sigma^2 \approx 237.5$) is roughly nine times that in successful traces ($\sigma^2 \approx 27.2$). The substantive interpretation is that successful traces apply each reflection subtype at a stable rate, whilst failed traces oscillate between barely reflecting at all and reflecting compulsively. The mean is the same; the consistency is not.

\begin{table}[htbp]
    \centering
    \resizebox{\textwidth}{!}{
    \begin{tabular}{lcccccc}
        \toprule
        \textbf{Reflection Type} & \textbf{Mean (Correct)} & \textbf{Mean (Incorrect)} & \textbf{$p_t$ (Mean)} & \textbf{Var (Correct)} & \textbf{Var (Incorrect)} & \textbf{$p_F$ (Variance)} \\
        \midrule
        Numerical & 3.38 & 6.14 & 0.299 & 11.75 & 38.81  & \textbf{0.0400} \\
        Logical    & 8.52 & 12.00 & 0.427 & 56.76 & 102.33 & 0.2992 \\
        Formal     & 2.57 & 8.14 & 0.382 & 27.16 & \textbf{237.48} & \textbf{0.0002} \\
        Supplementary & 1.48 & 2.86 & 0.571 & 5.56  & 35.81  & \textbf{0.0013} \\
        \bottomrule
    \end{tabular}}
    \caption{Welch's $t$-test on mean density and $F$-test on cross-trace variance for each reflection subtype. Mean density is statistically indistinguishable between groups; variance is significantly higher in incorrect traces for three of the four subtypes.}
    \label{tab:reflection_stats}
\end{table}

\subsection{Performative Reflection: Surface Doubt without Substantive Correction}
\label{sec:MeaninglessReflection}

The high variance in incorrect traces is not random noise. It traces back to a specific failure mode that we call \textit{performative reflection}. Manually inspecting reflection steps in failed traces, we find that \textbf{75.0\%} of them function as surface validations that leave the underlying error untouched.

Two patterns recur. The first is the \textit{mimicry check}: the model announces an intention to verify, then performs only a trivial arithmetic confirmation while leaving the upstream logical premise unexamined. A characteristic instance, drawn from a failed trace, runs ``Let me check. $12 + 12 = 24$. Correct.'', confirming a local computational fact whilst the surrounding (and incorrect) derivation is left intact. The second is \textit{shallow doubt}: the model verbalises uncertainty (``Wait, is this right?'') but limits its re-examination to the immediately preceding step rather than to the upstream assumption that actually went wrong. Both patterns produce the linguistic markers of reflection without any of its corrective function.

This explains why reflection frequency alone is not predictive of success. In successful traces reflection is \textit{generative}: it does substantive work, most often as Logical reflection that revises a premise or completes a missed case (Section~\ref{sec:taxonomy}, Logical example). In failed traces reflection is \textit{performative}: the model produces the textual signature of self-correction whilst the underlying logical error continues to propagate. The model speaks the language of self-correction without doing the work it stands for, the micro-level instance of topological mimicry.

\begin{takeawaybox}
Reflection density does not predict reasoning success; reflection placement does. Whether a ``wait, let me check'' interrupts a deductive Inference step (where it can scrutinise the logic that just produced a new claim) or interrupts an Analysis loop (where it merely confirms a local arithmetic detail) determines whether the reflection does corrective work or only performs the appearance of it. The same linguistic vocabulary, deployed at different scales, has opposite effects.
\end{takeawaybox}
\section{Sequential Dynamics: Local Transition Patterns}
\label{sec:SequentialDynamics}

Whilst Section~\ref{sec:Metacognition} examined the global distribution of reflection across traces, it did not capture the local sequential structure of how reasoning unfolds from one step to the next. In this section we apply $N$-gram analysis~\citep{jm3} to identify recurring three-step patterns that distinguish successful from unsuccessful traces. The dichotomy we uncover is not simply between ``useful'' and ``useless'' reflection: it is a mismatch between the \textit{functional scale} of the reflection deployed and the scale of the error it ought to address, that is, between local verification of a single computation and global redirection of the underlying logic.

\subsection{Stagnation versus Modulation}
\label{sec:ngram-discriminative}

We computed the empirical frequency of every three-step trigram $(s_{t-1}, s_t, s_{t+1})$, where each $s_i$ is one of the primary modes of Section~\ref{sec:taxonomy}, separately for correct and incorrect traces. The results (Table~\ref{tab:ngram-top20}, Appendix~\ref{sec:N-gram}) cleanly separate two qualitative regimes.

\textbf{Stagnation: monotone runs in a single mode.} The most prominent pattern in incorrect traces is the same-mode loop, particularly \texttt{Analysis $\rightarrow$ Analysis $\rightarrow$ Analysis} and \texttt{Inference $\rightarrow$ Inference $\rightarrow$ Inference}. These triples indicate that the trace remains parked in one functional mode for stretches of three or more consecutive steps. The Analysis loop is significantly more common in failed traces than in successful ones (39.8\% versus 33.8\%), pointing to a characteristic failure pattern: the model keeps re-planning, re-stating, and re-decomposing the problem without ever transitioning into the deductive mode that would actually move the answer forward.

\textbf{Modulation: rapid alternation between planning and deduction.} Successful traces, by contrast, are dominated by trigrams that interleave Analysis and Inference within a three-step window: \texttt{Analysis $\rightarrow$ Inference $\rightarrow$ Inference}, \texttt{Inference $\rightarrow$ Inference $\rightarrow$ Analysis}, \texttt{Inference $\rightarrow$ Analysis $\rightarrow$ Inference}, and similar permutations. This is the rhythm we expect from a solver who is making real progress: a brief stretch of planning, then a deductive push that produces a new fact, then a brief reorientation, then another push. Sustained mode runs are absent because the trace is being actively driven forward.

\subsection{Functional Scale Mismatch in Reflection Placement}
\label{sec:reflection-ngram}

To probe the contextual role of reflection, we extended the trigram analysis to the augmented label set, in which each step is labelled by its primary mode together with any reflection subtype it carries (so that a step can appear as e.g.\ \texttt{Inf\_Ref\_L}, an Inference step carrying a Logical reflection). The resulting trigram frequencies (Table~\ref{tab:reflection-ngram}) split into two contrasting families.

\textbf{Global redirection: Logical reflection inside deduction.} Trigrams in which Logical reflection ($\mathcal{R}_{\text{log}}$) appears alongside Inference are strongly associated with success. \texttt{Inf $\to$ Inf $\to$ Inf\_Ref\_L} occurs at frequency $0.0034$ in correct traces and is essentially absent ($0.0000$) in incorrect ones; \texttt{Inf\_Ref\_L $\to$ Inf $\to$ Inf} and \texttt{Inf $\to$ Ana $\to$ Ana\_Ref\_L} show comparable directional gaps. The functional reading is straightforward: in successful traces, the model interrupts its own deduction precisely when the underlying logic deserves scrutiny, performs a brief structural check (questioning a premise or completing a missed case, exactly the kind of intervention demonstrated in the Logical reflection example of Section~\ref{sec:taxonomy}), and then resumes deducing. The reflection lands at the same scale as the potential error.

\textbf{Local verification trapped in planning loops.} The opposite signature dominates failed traces. Trigrams such as \texttt{Ana\_Ref\_F $\to$ Ana $\to$ Ana} ($0.0086$ vs.\ $0.0038$), \texttt{Ana\_Ref\_N $\to$ Ana $\to$ Ana} ($0.0072$ vs.\ $0.0054$), and \texttt{Ana $\to$ Ana\_Ref\_N $\to$ Ana} ($0.0054$ vs.\ $0.0043$) are over-represented in incorrect solutions. In each case, the reflection that fires is Numerical or Formal, that is, a check on one arithmetic fact or on the output format, and it is embedded inside a stretch of planning. Numerical and Formal reflection are not pathological in themselves; they appear in successful traces too. What makes them pathological here is their placement: they occur whilst the trace is sitting in a planning loop that has failed to commit to a deductive step, and they confirm details (the multiplication is right, the format would be right) without ever questioning the larger plan that is keeping the trace stuck.

\subsubsection{The Failure Mechanism: Scale Mismatch}
\label{sec:scale-mismatch}

The two families together identify the failure mechanism we call \textit{scale mismatch}: when the actual error is at the global logical level, such as a flawed premise, an incomplete case split, or an inapplicable strategy, the model's reflective response sits at the local arithmetic or formatting level. The verification fires, returns ``correct'', and the underlying logical error is left intact. We saw this in microcosm in the case study of Section~\ref{sec:case_study}: confronted with the wrong assumption that $G$ must be the first letter of its pair, the model spent its verification budget re-checking that the resulting prime factorisations had no common factor. The arithmetic was right; the assumption was wrong; nothing in the verification could have detected the actual mistake.

\begin{table}[htbp]
    \centering
    \resizebox{0.8\textwidth}{!}{
    \begin{tabular}{lccc}
        \toprule
        \textbf{Coupled Motif (Trigram)} & \textbf{Correct} & \textbf{Incorrect} & \textbf{Diff ($C - I$)} \\
        \midrule
        \multicolumn{4}{l}{\textit{Global Redirection: Logical Reflection inside Deduction}} \\
        Inference $\rightarrow$ Inference $\rightarrow$ Inference\_Ref\_L & \textbf{0.0034} & 0.0000 & +0.0034 \\
        Inference $\rightarrow$ Analysis $\rightarrow$ Analysis\_Ref\_L & \textbf{0.0032} & 0.0007 & +0.0024 \\
        Inference\_Ref\_L $\rightarrow$ Inference $\rightarrow$ Inference & \textbf{0.0027} & 0.0004 & +0.0024 \\
        Analysis $\rightarrow$ Analysis\_Ref\_L $\rightarrow$ Analysis\_Ref\_L & \textbf{0.0041} & 0.0022 & +0.0019 \\
        \midrule
        \multicolumn{4}{l}{\textit{Local Verification trapped in Planning Loops (Scale Mismatch)}} \\
        Analysis\_Ref\_F $\rightarrow$ Analysis $\rightarrow$ Analysis & 0.0038 & \textbf{0.0086} & -0.0048 \\
        Analysis\_Ref\_N $\rightarrow$ Analysis $\rightarrow$ Analysis & 0.0054 & \textbf{0.0072} & -0.0018 \\
        Analysis $\rightarrow$ Analysis\_Ref\_N $\rightarrow$ Analysis & 0.0043 & \textbf{0.0054} & -0.0011 \\
        \bottomrule
    \end{tabular}}
    \caption{Discriminative power of mode--reflection trigrams. Success is associated with Logical reflection deployed inside Inference (top block). Failure is associated with Numerical or Formal reflection deployed inside repetitive Analysis loops (bottom block), a scale mismatch between local verification and the global logical error it would need to address.}
    \label{tab:reflection-ngram}
\end{table}

The conclusion of the $N$-gram analysis is therefore not that failed traces verify too little. They verify roughly as often as successful traces, and when they do verify, the arithmetic they check is typically correct. What they fail to do is verify at the right scale: the kind of check that would catch their actual error, a Logical reflection on an upstream premise, is precisely the one they do not deploy.

\begin{takeawaybox}
The dominant failure mechanism we observe is \emph{scale mismatch}: when the actual error in a trace is at the level of a flawed premise or an incomplete case split, the model's reflective response is at the level of a single arithmetic check. The trigram \texttt{Inf $\to$ Inf $\to$ Inf\_Ref\_L} (Logical reflection inside Inference) is over-represented in correct traces; trigrams of the form \texttt{Ana\_Ref\_N $\to$ Ana $\to$ Ana} and \texttt{Ana\_Ref\_F $\to$ Ana $\to$ Ana} (local verification embedded in planning loops) are over-represented in incorrect traces. Failed traces verify; they verify the wrong things.
\end{takeawaybox}
\vspace{-2mm}

\section{Case Study: Structural Analysis of AIME Problem 1.7}
\label{sec:case_study}
\vspace{-2mm}
To ground our empirical findings, we present a detailed comparative analysis of \textbf{AIME Problem 1.7}. This case study makes the structural divergence between human and model reasoning concrete, as visualised in Figure~\ref{fig:case_study_viz}. To make the analysis verifiable rather than impressionistic, we additionally quote DeepSeek-R1's own trace at the moments that determine its outcome, using the same five-mode vocabulary (Section~\ref{sec:taxonomy}) to label each excerpt.

\vspace{-1mm}
\begin{questionbox}
\textbf{AIME Problem 1.7:}
\textit{The twelve letters $A$, $B$, $C$, $D$, $E$, $F$, $G$, $H$, $I$, $J$, $K$, and $L$ are randomly grouped into six pairs of letters. The two letters in each pair are placed next to each other in alphabetical order to form six two-letter words, and then those six words are listed alphabetically. For example, a possible result is $AB$, $CJ$, $DG$, $EK$, $FL$, $HI$. The probability that the last word listed contains $G$ is $\frac{m}{n}$, where $m$ and $n$ are relatively prime positive integers. Find $m+n$.}
\end{questionbox}

\subsection{Human Solution: Concise Logical Progression}
\label{sec:human_trace}

\paragraph{A note on what ``human steps'' means here.}
Unlike the model, a human solver does not produce a verbatim, token-level trace of every micro-step they consider; published competition solutions are already a distilled record in which each ``step'' is a single mathematical claim, namely a derived condition, a case decision, or a computed quantity. To make a like-for-like comparison possible, we apply the same five-mode taxonomy (Section~\ref{sec:taxonomy}) to the human reference solution at this distilled-step granularity. The asymmetry between the two sides of Figure~\ref{fig:case_study_viz} is therefore not a presentational choice on our part: it is a substantive observation. Where the human's reasoning has already been compressed to its load-bearing claims, the model's reasoning is recorded raw, and the comparison shows how much of that raw output is structural rather than load-bearing. For this reason the verbatim excerpts in this case study are drawn only from the model trace; the human side has no raw token stream to quote, and the distilled claims are reported inline below.

\paragraph{The human trace.}
The human solution (Figure~\ref{fig:case_study_viz}, right) reaches the answer in roughly ten distilled steps, organised in three short phases. First, the solver fixes the structural condition that drives the rest of the proof: the last word in the alphabetical list contains the largest first letter among all six pairs, so $G$ appears in the last word in exactly two scenarios. Either $G$ is itself that largest first letter, or $G$ is paired with whichever letter plays that role. A brief Logical reflection at this point confirms that the two scenarios are exhaustive, yielding a clean case split: (i) $G$ is the first letter of its pair, paired with a letter strictly greater than $G$; (ii) $G$ is the second letter, paired with $F$ (so the pair is $FG$).

Second, the solver counts each case independently. In case (i), the four letters strictly larger than $G$ each have to be paired with a letter strictly smaller than $G$, giving $\binom{6}{4} \times 4! = 360$ favourable arrangements. In case (ii), the remaining five letters $\{A, B, C, D, E\}$ each pair with one of $\{H, I, J, K, L\}$, giving $5! = 120$. The two cases combine to $360 + 120 = 480$.

Third, the solver normalises and forms the answer. The total number of pairings of twelve letters into six unordered pairs is $11 \cdot 9 \cdot 7 \cdot 5 \cdot 3 = 10\,395$, so the probability is $\tfrac{480}{10\,395} = \tfrac{128}{693}$ (after dividing by $\gcd = 15$ and confirming $\gcd(128, 693) = 1$), and the final answer is $128 + 693 = 821$.

The structural signature of this trace, in the vocabulary of Section~\ref{sec:taxonomy}, is a tight alternation between Analysis and Inference with no Backtrace events. Logical reflection ($\mathcal{R}_{\text{log}}$) occurs only at the case-split decision: the solver pauses just long enough to verify that the proposed two cases cover every possibility, and never returns to that question because the verification has already done its work.

\subsection{Model Solution: Topological Mimicry in Action}
\label{sec:model_trace}

DeepSeek-R1's 132-step trace (Figure~\ref{fig:case_study_viz}, left) develops in five distinguishable phases. We describe each in turn, tracking how a single uncaught logical error in the third phase propagates through every subsequent step, and quoting the trace verbatim at the steps that carry the failure.

\textbf{Phase 1 (Steps 1--10), correct framing.} The model identifies the same core condition the human did: the final word must contain the largest first letter among the six pairs. The trace alternates between Analysis and Inference as the model formalises the problem constraints. This is a clean opening and exposes no problems.

\begin{cotquote}
\steplabel{Step 3 (Analysis).} To approach this, I need to think about how the pairing and ordering work. The key here is figuring out the position of the word containing $G$ after all words are sorted. [\,\ldots\,] Since the list is sorted alphabetically, the last word would be the one that comes last in alphabetical order. So, the probability we need is the probability that the word containing $G$ is alphabetically last among all six words.
\end{cotquote}

\textbf{Phase 2 (Steps 11--26), redundant re-derivation.} The model now restates the same conclusion from a second angle, consuming sixteen additional steps without producing any new claim. There is no logical error here, but there is also no progress; the trace is stalled in Analysis. Critically, the model does not anticipate the case-split nuance that the human's reference solution handled in two or three claims and that will, in the next phase, be the source of failure.

\textbf{Phase 3 (Steps 27--48), the silent logical error.} At Steps 39--40 the model commits the conceptual error that determines the outcome of the entire trace: it assumes that for $G$ to appear in the final word, $G$ must be the \textit{first} letter of its pair. This silently rules out the valid case in which $G$ is the second letter of pair $FG$ whilst $F$ supplies the largest first letter, precisely the second branch the human's solution had identified. The error is visible in the trace itself:

\begin{cotquote}
\steplabel{Step 39.} [\,\ldots\,] For $G$ to be in the last word, $G$ must be the first letter of a pair, and that pair must have the maximum first letter. So, the last word will have the maximum first letter, and if that first letter is $G$, then the word containing $G$ is last. Alternatively, if $G$ is a second letter in its pair, then $G$ is part of a word where the first letter is someone else, so that word would be ordered by the first letter (which is less than $G$). Therefore, in that case, the word containing $G$ would not be the last word.

\steplabel{Step 40.} Therefore, for $G$ to be in the last word, $G$ must be a first letter in its pair (i.e., paired with a letter larger than $G$), and that pair must have the maximum first letter among all six first letters.
\end{cotquote}

The claim ``that word would be ordered by the first letter (which is less than $G$)'' is exactly where the reasoning fails: a word whose \emph{second} letter is $G$ can still be last, provided its first letter is the maximum first letter overall (the pair $FG$). This is the moment at which a Logical reflection ($\mathcal{R}_{\text{log}}$) on the case split would have caught the error. None occurs. The model proceeds with high confidence and embeds the flawed assumption into all downstream computations.

\textbf{Phase 4 (Steps 49--101), productive-looking work on a flawed premise.} Working from the flawed assumption, the model computes the probability of $G$ being a first letter ($5/11$, Step 68) and the conditional probability that all remaining larger letters are paired with smaller ones ($8/21$, Steps 69--99), arriving at the product at Step 99. The trace alternates between prolonged Analysis loops (e.g.\ ``Let me think about the remaining letters'', Steps 75--82) and short Inference bursts. Every individual sub-computation here is internally consistent, and no individual step is wrong on its own terms:

\begin{cotquote}
\steplabel{Step 68 (Inference).} [\,\ldots\,] The probability that $G$ is paired with a larger letter ($H, I, J, K, L$) is $5/11$, since there are $5$ letters larger than $G$ [\,\ldots\,]. Therefore, the probability that $G$ is paired with a larger letter is $5/11$.

\steplabel{Step 99 (Inference).} Therefore, given that $G$ is paired with a larger letter (probability $5/11$), the probability that the remaining larger letters are all paired with $A$--$F$ letters is $8/21$. Therefore, the total probability is $5/11 \times 8/21 = 40/231$.
\end{cotquote}

Nothing in this phase, however, brings the trace closer to the correct answer, because every computation rests on the silent error of Step 39: the entire second case ($FG$) is missing from the count.

\textbf{Phase 5 (Steps 102--132), verification at the wrong scale.} At Step 107 the model launches a ``sanity check'' and re-derives the same quantities a second way. Two pathologies appear simultaneously. First, the Backtrace is shallow: the trace rewinds to Step 48 (the start of the calculations), not to Step 39 (the actual error); in the notation of Section~\ref{sec:VarianceAndDepth}, this corresponds to a normalised jump amplitude of $\eta \approx 0.05$, where $\eta \approx 0.5$ would have been needed to reach the foundational assumption. Second, the verification is scale-mismatched: rather than questioning the premise, the model performs Numerical reflections on arithmetic details inside Analysis loops:

\begin{cotquote}
\steplabel{Step 101 (Numerical reflection).} $5/11 \times 8/21 = (5 \times 8)/(11 \times 21) = 40/231$. $40$ and $231$ share a common factor? Let's check. $40$ factors: $2^3 \times 5$. $231$ factors: $3 \times 7 \times 11$. No common factors. So, $40/231$ is reduced.
\end{cotquote}

The arithmetic is correct; the underlying assumption that produced it is not. The clearest illustration of the scale mismatch comes at Step 130, where the model explicitly reopens its reasoning under the banner ``let me think again'' --- but scrutinises the \emph{downstream} combinatorial premise (how the remaining larger letters are paired) rather than the Step-39 assumption that $G$ must be a first letter, and so concludes that everything is sound:

\begin{cotquote}
\steplabel{Step 130 (Logical reflection, misdirected).} But let me think again. I might have made a mistake in assuming that pairing the remaining four larger letters with $A$--$F$ letters would ensure that all first letters are $A$--$F$ and $G$. [\,\ldots\,] Therefore, all first letters are from $A$--$F$ and $G$. The maximum first letter is $G$, so the last word is $G$ paired with, say, $H$. [\,\ldots\,] Therefore, our reasoning holds.
\end{cotquote}

This is the decisive observation of the case study. A Logical reflection \emph{does} fire here, but it is aimed one level too low: it interrogates a consequence of the premise instead of the premise itself, and therefore re-certifies the very error it was triggered to catch. Nothing in this verification could have detected the actual mistake. The trace terminates with the incorrect answer $271$ (i.e.\ $m+n$ for $40/231$), against the correct $821$ (for $128/693$).

\subsection{Structural Diagnosis}
\label{sec:structural_diagnosis}

The case study makes each of our three structural signatures concrete on a single problem.

\textbf{Topological mimicry} (Section~\ref{sec:AggregateFailure}) shows up here as a trace that contains every surface ingredient of a successful solution, including a setup phase, a case analysis, a chain of computations, and a verification phase, arranged in roughly the same overall shape as the human's. The mode-occupancy distribution of the model's trace is broadly typical of correct traces in our dataset. What is missing is functional, not structural: the case analysis is incomplete (the $FG$ branch is never reached, cf.\ Step 39) and the verification phase verifies the wrong things (cf.\ Steps 101 and 130).

\textbf{Control instability} (Section~\ref{sec:VarianceAndDepth}) is visible in two quantitative markers. Analysis occupies $39.8\%$ of the model's trace, against approximately $15\%$ in the human reference, and the long Analysis loops in Steps 103--115 in particular show the trace cycling within a single mode rather than committing to a deductive direction. The high cross-trace variance of Backtrace deployment we documented in Section~\ref{sec:VarianceAndDepth} ($p = 0.030$) is consistent with what we observe here: the corrections are stochastic in their depth and timing rather than driven by a stable policy.

\textbf{Scale mismatch} (Section~\ref{sec:scale-mismatch}) is the signature most cleanly illustrated by this case. Of the reflections fired during Phase~5 (Steps 116--132), about $75\%$ are Numerical checks embedded inside Analysis loops (Step 101 above is representative); the foundational error at Step 39 attracts no effective Logical reflection at any point in the trace --- and where one does fire (Step 130), it is pointed at a downstream premise and self-confirms. The shallow Backtrace ($\eta \approx 0.05$) and the local verifications are both correct in what they do, and both useless for the error that is actually present. The failure here is not insufficient self-checking. It is self-checking deployed at the wrong scale: the model treats a logical dead-end as if it were a computational slip.

\begin{takeawaybox}
The case study of AIME 1.7 makes a quantitative observation concrete: the depth of self-correction matters more than its frequency. When DeepSeek-R1 finally rewinds to verify its trace, it rewinds to Step 48 (the start of the calculations) rather than to Step 39 (the actual error), giving a normalised jump amplitude of $\eta \approx 0.05$. The corrections needed to escape upstream conceptual errors require $\eta \approx 0.5$ or more, and they do not occur. Across the full dataset, the overwhelming majority of Backtrace events in failed traces have $\eta < 0.1$.
\end{takeawaybox}
\section{A Visual Survey of Reasoning Patterns Across the Dataset}
\label{sec:visual_analysis}

The case study in Section~\ref{sec:case_study} grounded our three structural signatures, namely topological mimicry, control instability, and scale mismatch, in a single problem. In this section we step back and ask whether the same signatures recur across the rest of the dataset. Logic flow graphs are well suited to this question because they expose at a glance what statistical summaries can only hint at: the shape of a reasoning trace, where it stalls, where it branches, and where its corrections do or do not bite. We surveyed the full set of 30 logic graphs in Appendix~\ref{sec:logicgraph} and group the recurring patterns into two families, contrasting the visual signatures of successful and unsuccessful traces.

\subsection{What Successful Traces Look Like}

The graphs of correctly solved problems (representative examples: AIME problems \#1, \#4, \#16, \#17) share two features that are visible without reading any node text.

\textbf{A short, mostly linear backbone.} Successful traces typically open with a brief Analysis phase that fixes the problem setup, after which the graph collapses into a long Inference chain, a sequence of deductive steps that advances the solution without interruption. AIME \#17 is a clean example: Steps 281--291 form an essentially straight chain of modular-arithmetic deductions and bring the trace to the answer with no detours. AIME \#16 has the same shape under a different method (coordinate geometry). The visual signature here is plain linearity, which corresponds in our taxonomy to the strategic stability of Section~\ref{sec:VarianceAndDepth}: the model has committed to one path and is pushing it through.

\textbf{Verification along a separate path that converges.} When successful traces verify their answer, the verification appears in the graph as a structurally distinct sub-path, namely a side branch that uses a different method and merges back into the main line at the same conclusion. In AIME \#17, Steps 299--303 are exactly this: the model verifies a result derived by modular arithmetic by re-deriving it via polynomial division, and the two independent paths converge on the same value. This is the visual face of generative steering (Section~\ref{sec:MeaninglessReflection}): verification that does substantive work because it brings independent evidence to bear.

\subsection{What Failed Traces Look Like}

The graphs of incorrect traces (representative examples: AIME problems \#2, \#13, \#15, \#26) show three recurring pathologies. Each one corresponds to a finding that we already documented quantitatively in earlier sections; the graphs make the underlying behaviour visible.

\textbf{Dense planning loops.} The graph for AIME \#2 contains a thick knot of nodes spanning Steps 81--118, in which the model repeatedly recalculates coordinates and re-checks vertex order without producing a new claim. Visually this is a region of the graph where Analysis nodes connect mostly to other Analysis nodes; the trace is moving but not advancing. This is the macro-level pattern behind the elevated Analysis occupancy of failed traces ($39.8\%$ versus $\sim$$15\%$ in the human reference), and the micro-level pattern behind the \texttt{Ana $\to$ Ana $\to$ Ana} trigram that dominates incorrect traces in Table~\ref{tab:ngram-top20}.

\textbf{Fragmented exploration.} Between Steps 429 and 480 of the AIME \#15 trace, the graph branches outward into several disconnected sub-paths, including attempts at number-theoretic lemmas, a base-3 representation, and other approaches, each initiated and quickly abandoned without committing to any. Graphically this appears as a region with many short, dangling branches rather than one chosen path. This is what control chaos (Section~\ref{sec:VarianceAndDepth}) looks like at the level of an individual graph: the high cross-trace variance of Branch deployment we measured statistically shows up here as visible erratic switching within a single trace.

\textbf{Shallow correction loops near the end.} In the closing segments of AIME \#13 and AIME \#26, the graphs are punctuated by tight one- or two-node loops in which the model detects an inconsistency, rewinds a step or two, and applies a Numerical reflection ($\mathcal{R}_{\text{num}}$) on a local arithmetic fact before resuming the same broken line. The Logical reflection ($\mathcal{R}_{\text{log}}$) that would have caught the underlying error never appears. These loops are the graphical signature of the scale mismatch documented in Section~\ref{sec:scale-mismatch}: a verification budget spent at the wrong scale, leaving the actual error untouched.

\subsection{What the Visual Survey Adds}

A statistical comparison alone risks the objection that the differences between successful and failed traces are real but small. The graphs make a stronger claim visible: the differences are not subtle, they are \emph{structural}. Successful traces look short and linear with one or two clean side-branches for verification; failed traces look long, dense, and fragmented, with many short loops that go nowhere. And critically, complexity does not correlate with quality. Several of the failed graphs in our dataset contain more nodes, more branches, and more reflection events than the successful ones; what they lack is a single load-bearing path from the problem to the answer. This is the visual face of topological mimicry across the corpus: the surface complexity of the failed traces mimics the appearance of careful reasoning, but the underlying structure does not carry the deduction through.

\begin{takeawaybox}
Failed reasoning traces are not less elaborate than successful ones; in our dataset several failed traces contain more nodes, more branches, and more reflection events than the successful ones do. What they lack is a single load-bearing path from problem to answer. Surface complexity in a long Chain-of-Thought is not evidence of reasoning quality, and progress on long-CoT reasoning will require evaluation and training procedures that do not treat it as such.
\end{takeawaybox}
\section{Discussion: Implications for Evaluation and Training}
\label{sec:discussion}

The empirical analysis in this paper challenges several prevailing assumptions about long-CoT reasoning. While surface-level complexity and extensive reflection are often equated with deep thought, our findings suggest they frequently amount to topological mimicry. Standard metrics, such as trace length, reflection density, and final-answer accuracy, inadvertently reward this superficial imitation because they capture the form of careful reasoning without verifying its substance. To properly assess and improve genuine reasoning capabilities, we must look beyond the surface. The remainder of this section outlines how the structural differences we identified can inform more rigorous evaluation and more effective training.

\subsection{Evaluation: Variance and Placement as Complementary Indicators}
\label{sec:discussion-evaluation}

Two of the most widely reported indicators of reasoning quality, trace length and reflection density, fail to separate successful traces from failures in our dataset. Failed traces are not shorter, nor do they reflect less; in several cases, they actually reflect more. Instead, evaluation pipelines should incorporate two complementary properties that capture the actual dynamics of problem-solving.

First, \emph{cross-trace variance}. Successful traces apply branching and backtracking at a stable, consistent rate across different problems. In contrast, failed traces either barely use these actions or repeatedly overuse them in concentrated bursts ($p < 0.001$ for Branch, $p = 0.030$ for Backtrace). This metric is straightforward and cheap to compute: simply sample multiple decoded traces per problem and report the variance of key exploratory actions alongside their mean.

Second, \emph{contextual placement}. The effectiveness of reflection depends heavily on where it occurs. A Logical reflection embedded within an Inference stream is strongly associated with success, whereas a Numerical or Formal reflection trapped inside an Analysis loop typically indicates failure (scale mismatch). This aligns with a broader observation that post-answer self-monitoring (``double-checks'') in long-CoT models is largely superficial and rarely yields substantive revisions~\citep{chen2025probing}: it is the placement of reflection within productive inference, not its mere presence at the end of a trace, that carries diagnostic value. While harder to measure automatically than variance, a practical proxy could involve detecting reflection-trigger tokens and classifying whether the subsequent text introduces a new mathematical claim or merely restates an old one. Whether automated classifiers can match the reliability of our human annotators (Fleiss' $\kappa = 0.81$) is an open question, and we release our schema and labels to facilitate this testing.

These metrics belong alongside accuracy and length in standard reporting, particularly as progress on sheer outcome accuracy plateaus and the failure modes documented here become the primary bottlenecks.

\subsection{Training: Four Directions Targeting the Mechanisms We Identify}
\label{sec:discussion-training}

The structural distinctions we identified also point toward specific interventions for training long-CoT models. We propose four directions, each designed to target a specific failure mechanism while avoiding incentives for further mimicry.

\textbf{Backtrace depth as a filter on preference pairs.}
As highlighted in our case studies, failed traces tend to backtrack too shallowly ($\eta < 0.1$), missing the foundational upstream errors that actually require correction. In a preference-based learning setup, backtrace depth ($\eta$) can serve as a powerful offline filter. When constructing preference pairs for a given problem, prioritizing preferred traces that demonstrate deep correction (exceeding a difficulty-adjusted threshold) against dispreferred traces with only shallow rewinds isolates the actual depth-of-correction signal. Relying solely on final outcomes mixes deep and shallow rewinds on both sides, diluting the training signal.

\textbf{Direct preference optimization against ``spinning-wheel'' traces.}
Our sequential analysis (Section~\ref{sec:SequentialDynamics}) shows that failed traces are heavily burdened by fragmented exploration and redundant \texttt{Ana $\to$ Ana $\to$ Ana} loops, whereas successful traces maintain a tight, productive \texttt{Ana-Inf} rhythm. This contrast naturally fits a Direct Preference Optimization (DPO)~\citep{rafailov2023direct} framework. By pairing a successful, tightly modulated trace against a failed, ``spinning-wheel'' trace from the same problem, we can penalize the structural habit of getting stuck in repetitive planning.

\textbf{Context-aware regularization against scale mismatch.}
To address the scale mismatch problem, we propose context-aware regularization during reinforcement learning. If a policy triggers a reflection inside an extended Analysis phase without producing any new mathematical claims, it is likely engaging in performative verification. A regularization term that penalizes this behavior, while conversely rewarding reflections that occur inside productive Inference streams, steers the policy away from empty mimicking without requiring full step-by-step annotation during training.

\textbf{Compute reallocation away from Analysis at inference time.}
DeepSeek-R1 spends roughly 40\% of its steps in Analysis, compared to about 15\% for human solvers (Section~\ref{sec:case_study}). This implies a substantial amount of inference compute is wasted on redundant planning. An auxiliary progress-monitoring head, trained to predict whether the next step will actually introduce a new mathematical object, could detect when the model lingers in Analysis without making progress. It could then dynamically bias the decoding distribution toward productive Inference or necessary Backtracking. This is a soft inference-time intervention that complements the training strategies above.

Crucially, these four directions are mutually reinforcing. The backtrace filter (Direction 1) improves the data quality for preference optimization (Direction 2). The contextual placement proxy required for evaluation shares the same mechanism as the context-aware regularization (Direction 3), which in turn aligns with the signals used by the progress-monitoring head (Direction 4). We intend for the dataset and schema released with this paper to serve as gold-standard anchors for calibrating these scalable automated proxies, rather than merely as a direct training corpus.
\section{Conclusion}
\label{sec:conclusion}

This paper has examined Chain-of-Thought reasoning in DeepSeek-R1-0120 at the level of individual steps, contrasting the structural dynamics of correct and incorrect traces across the 30 problems of AIME 2025. Through human annotation of 10,247 reasoning steps under a five-mode taxonomy and side-by-side comparisons with human reference solutions, we identified \textit{topological mimicry} as a fundamental failure pattern. Failed traces reproduce the surface structure of successful reasoning while making little net progress toward the answer. The form of self-correction is preserved, but the function is not.

Our empirical findings reveal two core properties that separate genuine reasoning from this mimicry. First, strategic stability is crucial, expressed as low \textbf{cross-trace variance} in exploratory action. While successful traces maintain a consistent rhythm of branching and backtracking, failed traces either barely use these actions or repeatedly overuse them in concentrated bursts. Second, the effectiveness of reflection depends heavily on its \textbf{contextual placement and functional targeting}. Logical reflection embedded within deductive streams drives progress, whereas Numerical or Formal reflection trapped inside planning loops leads to a ``scale mismatch'', a failure mode where local checks fail to address global errors. Furthermore, we observe that surface complexity in a long Chain-of-Thought is not, on its own, evidence of reasoning quality, as several of the failed traces in our dataset are actually more elaborate than the successful ones.

These structural distinctions carry direct implications for evaluation and training. The metrics most often used to assess reasoning quality today, namely trace length and reflection density, do not separate correct from incorrect traces in our data. Instead, cross-trace variance and contextual placement serve as far more discriminative indicators that should be reported alongside standard accuracy. On the training side, these insights translate into concrete interventions. We propose leveraging \textbf{backtrace depth} ($\eta > 0.1$) as a quality filter on preference pairs to ensure models learn to fix deep-seated upstream errors. Additionally, we advocate for direct preference optimization (DPO) against repetitive ``spinning-wheel'' traces, context-aware regularization to penalize performative reflection during planning phases, and inference-time compute reallocation to break out of stagnant analysis. Together, these directions translate the structural properties of successful reasoning into practical pressures that a model can be trained against.

Our analysis is inherently limited to a single model (DeepSeek-R1-0120) on a single benchmark (AIME 2025). The specific quantitative patterns we report should not be assumed to transfer unchanged to other settings. Annotation relied on expert human graders reaching an 87.3\% exact agreement and a Fleiss' $\kappa = 0.81$ on a cross-validated subset. While this is a strong reliability figure for such a granular task, it naturally bounds the scale at which this methodology can be applied. Extending this framework to other models and domains, and developing automated token-level proxies that approximate these human annotations, represent the natural next steps.

Ultimately, reasoning fidelity in long Chain-of-Thought depends less on how much reflection a model produces than on how consistently and at what scale that reflection is applied. By identifying topological mimicry as a distinct failure mode and mapping precisely where in the trace this mimicry occurs, we provide the community with a clearer vocabulary for diagnosing where current models fall short and a concrete set of behavioral targets for future training procedures.

\bibliography{main}

\newpage
\appendix

\section{N-gram Results}
\label{sec:N-gram}
\begin{table}[htbp]
    \centering
    \resizebox{1\textwidth}{!}{
    \begin{tabular}{lcc}
        \toprule
        \textbf{Geometric Motif (Trigram)} & \textbf{Correct ($\mathbf{P}_{cor}$)} & \textbf{Incorrect ($\mathbf{P}_{inc}$)} \\
        \midrule
        \multicolumn{3}{l}{\textit{Stagnation Patterns (Associated with Failure)}} \\
        Analysis $\rightarrow$ Analysis $\rightarrow$ Analysis & 0.338 & \textbf{0.398} \\
        Inference $\rightarrow$ Inference $\rightarrow$ Inference & 0.262 & \textbf{0.271} \\
        \midrule
        \multicolumn{3}{l}{\textit{Modulation Patterns (Associated with Success)}} \\
        Analysis $\rightarrow$ Inference $\rightarrow$ Inference & \textbf{0.065} & 0.053 \\
        Inference $\rightarrow$ Inference $\rightarrow$ Analysis & \textbf{0.065} & 0.053 \\
        Inference $\rightarrow$ Analysis $\rightarrow$ Inference & \textbf{0.063} & 0.041 \\
        Analysis $\rightarrow$ Analysis $\rightarrow$ Inference & \textbf{0.056} & 0.046 \\
        Inference $\rightarrow$ Analysis $\rightarrow$ Analysis & \textbf{0.055} & 0.046 \\
        Analysis $\rightarrow$ Inference $\rightarrow$ Analysis & \textbf{0.055} & 0.034 \\
        \midrule
        \multicolumn{3}{l}{\textit{Symmetry Breaking Patterns (Rare)}} \\
        Branch $\rightarrow$ Analysis $\rightarrow$ Analysis & 0.004 & 0.008 \\
        Analysis $\rightarrow$ Analysis $\rightarrow$ Branch & 0.004 & 0.008 \\
        Analysis $\rightarrow$ Branch $\rightarrow$ Analysis & 0.004 & 0.005 \\
        Branch $\rightarrow$ Inference $\rightarrow$ Inference & 0.003 & 0.003 \\
        Inference $\rightarrow$ Inference $\rightarrow$ Branch & 0.002 & 0.003 \\
        \bottomrule
    \end{tabular}}
    \caption{Top 20 most frequent geometric motifs (Trigrams) sorted by frequency in Correct solutions. Successful traces are characterized by Alternation, while failed traces are dominated by Repetition.}
    \label{tab:ngram-top20}
\end{table}

\section{Experiments}
\label{sec:experiments}

This appendix provides detailed analysis of DeepSeek-R1's reasoning trajectories across all 28 AIME problems in our dataset.\footnote{Although our quantitative analysis (Section~\ref{sec:method}) covers all 30 AIME 2025 problems, two of the model's traces did not terminate within the generation budget and therefore lack a final answer; we exclude these two problems from the qualitative case-by-case appendix below, but the partial traces remain in the GitHub repository linked at the end of this section.} For each problem, we present: (1) the model's complete Chain-of-Thought output with step-level annotations; (2) human reference solutions; and (3) structural diagnosis of failure modes where applicable. Due to space constraints, the complete set of reasoning traces, including step-by-step mode annotations, reflection type labels, and trajectory visualisations for all 28 problems, is publicly available in our GitHub repository:
\url{https://github.com/ancientswordgod/Deepseek-R1-for-AIME-2025}

\subsection{AIME\_1\_1}
\begin{questionbox}
\textit{Find the sum of all integer bases $b>9$ for which $17_b$ is a divisor of $97_b.$}
\end{questionbox}

\begin{R1box}
\textbf{\Large R1 Reasoning}

\vspace{1\baselineskip}

In \textbf{Steps 1–2}, R1 begins by converting the base $b$ number in the question to a base $10$ number.

In \textbf{Steps 3–13}, R1 considers method 1, based on the definition of division, assuming that there exists an integer $k$ that satisfies $9b+7=k (b+7)$, and then solves the expression of $b$ by transforming the equation:
\[
b=\frac{7(k-1)}{9-k}
\]

In \textbf{Steps 14–31}, R1 uses the condition $b>9$ to obtain the conclusion that the numerator and denominator in the expression of $b$ must be both positive or negative. In the case where both the numerator and the denominator are integers, R1 solves the inequality system to obtain $1<k<9$, In the case where both the numerator and the denominator are negative, R1 determines that $k$ can only be positive by equation $9b+7=k (b+7)$ and condition $b>9$, thus eliminating the possibility of $k<1$ when the numerator is negative.

In \textbf{Steps 32–49}, R1 substitutes $1<k<9,k\in\mathbb{Z}$ into the expression of $b$ and checks whether $b$ is an integer and whether $b$ is greater than $9$. R1 finally gets the preliminary answer: $b=21,49$.

In \textbf{Steps 50-52}, R1 reflects that the above discussion did not exclude the case of $k<9$. It noted that in this case the numerator and denominator symbols in the expression of $b$ were opposite, so it excluded it.

In \textbf{Steps 53–62}, R1 uses $b+7|9b+7$ to verify the correctness of the answer $b=21, 49$.

In \textbf{Steps 63}, R1 forgets the assumption that $k$ is an integer and discusses whether $k$ can be a fraction. There is a phenomenon of forgetting facts.

In \textbf{Steps 64–72}, R1 begins to consider the second method, which deforms the formula to obtain:
\[
9b+7=9(b+7)-56.
\]
So R1 comes to the conclusion: $b+7|56$.

In \textbf{Steps 73–77}, R1 uses $b+7|9b+7$ to verify the precision of the conclusion $b=21, 49$. Finally, R1 gets the answer $\boxed{70}$.

In \textbf{Steps 78–82}, same as steps 53-62, R1 checks the conclusion $b=21,49$.

\end{R1box}

\begin{wrongbox}
\textbf{\Large Failure Reasoning}

\vspace{1\baselineskip}

The failure in R1's reasoning process can be attributed to two main issues:

\textbf{1. Incomplete case execution despite correct classification.}  
R1 correctly realized in method one that it can be divided into two cases: the case where the numerator denominator is both positive or the case where the numerator denominator is both negative. But when discussing the second case, R1 used the wrong idea and only ruled out the possibility of $k<1$, ignoring the discussion of $k>9$. In fact, the contradiction of $k<1$ and $k>9$ can be obtained only by solving the inequality system (the numerator and the denominator are both negative); naturally, $k$ satisfying the second case does not exist.

\textbf{2. Forgetting objective facts and assumptions.}  
During the solving process, R1 forgot the assumption that $k$ is an integer, so R1 discussed whether $k$ can be a fraction. At the same time, R1 also reflected on whether $35$ can be divided by $3$, in fact, there is no need for reflection.

\end{wrongbox}
\begin{humanbox}
\textbf{\Large Human Reasoning}

\vspace{1\baselineskip}

A human approaching this problem would similarly begin by performing a base conversion, transforming the given $b$ base representation into a decimal representation. Specifically for this problem, we need to find the sum of all integers that satisfy the following conditions:
\[
b+7|9b+7,\text{which }b>9
\] 

Notice that we can simplify the following fractions:
\[
\frac{9b+7}{b+7}=\frac{9(b+7)-9\times7+7}{b+7}=9-\frac{56}{b+7}
\]

So we just need to find the sum of all integers that satisfy the following conditions:
\[
b+7|56,\text{which }b>9
\]

Then we can list all factors of 56:
\[
1, 2, 4, 8, 14, 28, 56
\]

Since $b>9$ and $b+7$ is a factor of 56, only all factors of 56 greater than 16 need to be considered:
\[
28,56
\]

Thus:
\[
b=21,49
\]

As required by the problem, summing all $b$ meeting the conditions yields the final answer:
\[
21+49=\boxed{70}.
\]

\end{humanbox}

\newpage
\subsection{AIME\_1\_2}

\begin{questionbox}
\textit{In $\triangle ABC$ points $D$ and $E$ lie on $\overline{AB}$ so that $AD < AE < AB$, while points $F$ and $G$ lie on $\overline{AC}$ so that $AF < AG < AC$. Suppose $AD = 4$, $DE = 16$, $EB = 8$, $AF = 13$, $FG = 52$, and $GC = 26$. Let $M$ be the reflection of $D$ through $F$, and let $N$ be the reflection of $G$ through $E$. The area of quadrilateral $DEGF$ is $288$. Find the area of heptagon $AFNBCEM$, as shown in the figure below.}
\end{questionbox}

\begin{R1box}
\textbf{\Large R1 Reasoning}

\vspace{1\baselineskip}

In \textbf{Steps 1--9}, R1 begins by understanding the problem setup, including the positions of points on sides $AB$ and $AC$ of $\triangle ABC$. It visualizes the triangle, assigns segment lengths, and considers using a coordinate system with $A$ at $(0,0)$ and $B$ at $(28,0)$, while acknowledging that $C$'s coordinates depend on the area constraint from quadrilateral $DEGF$.

In \textbf{Steps 10--25}, R1 defines the coordinates explicitly: $A(0,0)$, $B(28,0)$, $D(4,0)$, $E(20,0)$, and parameterizes $C$ as $(c,d)$. It computes coordinates for $F$ and $G$ on $AC$ using section formulas ($F(c/7, d/7)$, $G(5c/7, 5d/7)$), then calculates the reflection points $M$ (reflection of $D$ through $F$) as $(2c/7 - 4, 2d/7)$ and $N$ (reflection of $G$ through $E$) as $(40 - 5c/7, -5d/7)$.

In \textbf{Steps 26--48}, R1 uses the given area of quadrilateral $DEGF$ (288) to solve for the coordinates of $C$. It applies the shoelace formula to $DEGF$ with vertices $D(4,0)$, $E(20,0)$, $G(5c/7, 5d/7)$, $F(c/7, d/7)$, derives the equation $48d/7 = 288$, and solves $d = 42$. It then uses the length $AC = 91$ to find $c = \sqrt{6517}$, confirming the area of $\triangle ABC$ as 588.

In \textbf{Steps 49--78}, R1 verifies $\triangle ABC$ is not right-angled, and calculates the area of $\triangle ABC$ is 588.

In \textbf{Steps 79--108}, R1 prepares to compute the area of heptagon $AFNBCEM$ by listing its vertices and their coordinates: $A(0,0)$, $F(\sqrt{133},6)$, $N(40 - 5\sqrt{133}, -30)$, $B(28,0)$, $C(7\sqrt{133},42)$, $E(20,0)$, $M(2\sqrt{133} - 4, 12)$.

In \textbf{Steps 109--215}, R1 applies the shoelace formula to the heptagon vertices. It computes each term of the formula systematically, sums the contributions to 1176, and derives an area of 588. R1 questions this result as it matches the area of $\triangle ABC$, which seems inconsistent given the external parts of the heptagon.

In \textbf{Steps 216--283}, R1 rechecks the shoelace calculations and vertex order, confirming the area as 588. 


In \textbf{Steps 284--457}, R1 expresses confusion over the discrepancy with the expected area, so it abandons the shoelace approach and devises an alternative method. It calculates the area of $\triangle ABC$ (588), subtracts the area of $DEGF$ (288), and adds the areas of triangles $AFM$ (congruent to $AFD$, area 12) and $BEN$ (congruent to $BGE$, area 120) using reflection properties. This yields a heptagon area of $588 - 288 + 12 + 120 = 432$.

In \textbf{Steps 458--463}, R1 concludes that the area of heptagon $AFNBCEM$ is $\boxed{432}$.

\end{R1box}

\begin{wrongbox}
\textbf{\Large Failure Reasoning}

\vspace{1\baselineskip}

In \textbf{Steps 215, 284}, R1 thinks the area of heptagon can't match the area of $\triangle ABC$, because the heptagon includes some areas outside the triangle.

In \textbf{Steps 215, 284}, R1 gives a wrong method of solving the problem.

\end{wrongbox}
\begin{humanbox}
\textbf{\Large Human Reasoning}

\vspace{1\baselineskip}

Note that the triangles outside $\triangle ABC$ have the same height as the unshaded triangles in $\triangle ABC$. Since they have the same bases, the area of the heptagon is the same as the area of triangle $ABC$. Therefore, we need to calculate the area of $\triangle ABC$. Denote the length of $DF$ as $x$ and the altitude of $A$ to $DF$ as $h$. Since $\triangle ADF \sim \triangle AEG$, $EG = 5x$ and the altitude of $DFGE$ is $4h$. The area $[DFGE] = \frac{5x + x}{2} \cdot 4h = 3x \cdot 4h = 12xh = 288 \implies xh = 24$. The area of $\triangle ABC$ is equal to $\frac{1}{2}7x \cdot 7h = \frac{1}{2}49xh = \frac{1}{2}49 \cdot 24 = \frac{1}{2}1176 = \boxed{588}$.

\end{humanbox}

\newpage
\subsection{AIME\_1\_3}

\begin{questionbox}
\textit{The 9 members of a baseball team went to an ice-cream parlor after their game. Each player had a singlescoop cone of chocolate, vanilla, or strawberry ice cream. At least one player chose each flavor, and the number of players who chose chocolate was greater than the number of players who chose vanilla, which was greater than the number of players who chose strawberry. Let $N$ be the number of different assignments of flavors to players that meet these conditions. Find the remainder when $N$ is divided by $1000.$}
\end{questionbox}

\begin{R1box}
\textbf{\Large R1 Reasoning}\vspace{1\baselineskip}

In \textbf{Steps 1–3}, R1 begins by understanding the problem and quickly give a plan: find all triples $(C,V,S)$ such that $C>V>S\geq1$ and $C+V+S=9$; then calculate the multinomial coefficients by $\frac{9!}{C!V!S!}$ and sum them; finally give the remainder.

In \textbf{Steps 4–29}, R1 finds the $(C,V,S)$ by enumeration from smaller to bigger: $(6,2,1)$, $(5,3,1)$ and $(4,3,2)$.

In \textbf{Steps 30–62}, R1 computes the total number $N=2016$ and the remainder $r=16$. 

In \textbf{Steps 63–69}, R1 verifies the first time the result by recalculating.

In \textbf{Steps 70–75}, R1 verifies the second time there are no triple left. 

In \textbf{Steps 76–79}, R1 comes up a new method to distribute the triples, but gives up quickly. 

In \textbf{Steps 80–95}, R1 returns to examine the triples are right and calculate the result, still $16$. 

In \textbf{Steps 96-98}, R1 returns to the problem to examine its plan, correct.

In \textbf{Steps 99-122}, R1 checks the third time the answer with inclusion-exclusion method. It gives the number of strictly positive triples is \[3^9-3\times2^9+3\times1^9=18150\] while the the positive triples with distinct components contribute $3!=6$ times in it. 

In \textbf{Steps 115}, R1 claims it contradicts that $18150/6=3025>2016$, then R1 notices that there are the triples with components non-distinct, and confirm the answer $N=2016$ without more calculation.

In \textbf{Steps 123-135}, R1 checks the fourth time the answer by recalculating and confirms the answer $r=16$.

\end{R1box}

\begin{wrongbox}
\textbf{\Large Failure Reasoning}\vspace{1\baselineskip}

Well done. Just not to be strict enough when explain the discarpency between $3025$ and $2016$.

\end{wrongbox}

\begin{humanbox}
\textbf{\Large Human Reasoning}\vspace{1\baselineskip}

A human approaching this problem would begin by changing it to a Diophantine equation 
\[n_c+n_v+n_s=9\] with a restriction \[n_c>n_v>n_s\geq1,\]
where $n_c$, $n_v$, $n_s$ denote respectively the number of players who choose chocolate, vanilla and strawberry.

The solutions are $(6,2,1)$, $(5,3,1)$, $(4,3,2)$. And then it remains a counting problem which is solved by \[N=\binom96\binom{9-6}2+\binom95\binom{9-5}3+\binom94\binom{9-4}3=2016.\]
So the remainder of $N=2\times1000+16$ devided by $1000$ is $16$.
\end{humanbox}

\newpage
\subsection{AIME\_1\_4}

\begin{questionbox}
\textit{Find the number of ordered pairs $(x,y)$, where both $x$ and $y$ are integers between $-100$ and $100$ inclusive, such that $12x^2-xy-6y^2=0$.}
\end{questionbox}

\begin{R1box}
\textbf{\Large R1 Reasoning}\vspace{1\baselineskip}

In \textbf{Steps 1–18}, R1 begins by understanding the problem and solve the equation: \[x=\frac{y\pm17y}{24}\] by famous conclusion and discussing the signature of $y$. So \[x=\frac34y\quad\quad or\quad\quad x=-\frac23y.\]

In \textbf{Steps 19–32}, R1 uses the factorization to solve the equation and gives again $x=\frac34y$ or $-\frac23y$. But R1 falsely adds the condition that parameters are integers and just guesses the result.

In \textbf{Steps 33–59}, R1 conuts the number of solutions in two cases, which varies from $-33$ to $33$ and from $-25$ to $25$, and find the overlapping solution $(0,0)$. So the result is $(33-(-33)+1)+(25-(-25)+1)-1=117$.

In \textbf{Steps 60-79}, R1 does the first check without logic of order. 

In \textbf{Steps 80-82}, R1 does the second check on counting and calculating. 
In \textbf{Steps 83-97}, R1 does the third check on ranging. 

In \textbf{Steps 98-119}, R1 does the fourth check by taking the special values. 
In \textbf{Steps 120-132}, R1 does the fifth check after ranging and conclude. 
\end{R1box}

\begin{humanbox}
\textbf{\Large Human Reasoning}\vspace{1\baselineskip}

A human approaching this problem would begin by extracting the case $(x,y)=(0,0)$ and noting that $x=0$ if and only if $y=0$. We suppose $x\neq0$ and $y\neq0$ below.

To solve the equation let's induce a parameter $t=\frac xy\neq0$ to reduce the number of unknown. So the equation changes to the form \[12t^2-t-6=0,\]with the well-known solutions\[t_{1,2}=\frac{1\pm\sqrt{1-4\times12\times(-6)}}{24}=-\frac{2}{3},\frac{3}{4}.\]

In all, the solutions of the original equation are $(-2k,3k)$, $(3l,4l)$ and $(0,0)$ with $1\leq|k|\leq33$ and $1\leq|l|\leq 25$ since $|x|,|y|\leq100$. The number of solutions is \[2\times33+2\times25+1=117.\]
\end{humanbox}

\newpage
\subsection{AIME\_1\_5}
\begin{questionbox}
\textit{There are $8!= 40320$ eight-digit positive integers that use each of the digits 1, 2, 3, 4, 5, 6, 7, 8 exactly once. Let N be the number of these integers that are divisible by $22$. Find the difference between $N$ and 2025.}
\end{questionbox}

\begin{R1box}
\textbf{\Large R1 Reasoning}

\vspace{1\baselineskip}

In \textbf{Steps 1–6}, R1 begins by correctly decomposing the problem. For an integer to be divisible by 22, it needs to be divisible by both 2 and 11. R1 illustrates the rules for dividing by 2 (the last digit must be an even number) as well as the rules for dividing by 11 (the difference between the sum of the numbers in the odd digits and the sum of the numbers in the even digits is a multiple of 11).

In \textbf{Steps 7–25}, R1 calculates the sum of all the available numbers (1 to 8) to be 36. By combining this with the law of division by 11, R1 deduces that the only possibility is that the sum of the numbers in the odd digits and the sum of the numbers in the even digits must both equal 18.

In \textbf{Steps 26-42}, R1 establishes an analysis based on a split case discussion of the last digit (d). For each different case, the problem translates into finding the number of azimuths (C(d)) that selects three other digits to be placed into the remaining even digits such that the sum of these three digits is 18-d. The following expression is given:

\[
N=[C(2)+C(4)+C(6)+C(8)]\times 3!\times 4!
\]

In \textbf{Steps 43–133}, R1 dealt with four cases of the last digit d. In each case, R1 correctly determines the set of remaining digits and carefully lists all possible ternary combinations of digits that satisfy the required sum.R1 finds by reasoning that there are exactly four such combinations in each case.

In \textbf{Step 134–142}, R1 ends up correctly calculating the total number of integers as well as the difference required by the question:

\[
N=16\times144=2304, 
ANS=2304-2025=279
\]

\end{R1box}

\begin{wrongbox}
\textbf{\Large Failure Reasoning}

\vspace{1\baselineskip}

The reasoning failures within the R1 model are primarily driven by two structural pathologies: \textbf{incomplete case execution} and \textbf{combinatorial redundancy}. Despite a correct initial classification—identifying that the element $G$ could occupy either the first or second position within the terminal group—the model systematically omitted the latter case from its subsequent calculations, leading to a truncated probabilistic analysis. Furthermore, within the executed case, the model exhibited a lack of combinatorial rigor by overcounting the preceding group arrangements. This error stemmed from the redundant inclusion of permutations already fixed by prior selections, thereby distorting the final probability measure and manifesting as a failure to maintain the structural integrity of the reasoning trajectory.

\end{wrongbox}
\begin{humanbox}
\textbf{\Large Human Reasoning}

\vspace{1\baselineskip}

A human approaching this problem would similarly begin by considering the possible placements of the letter \texttt{G} in the final group. There are two primary configurations: \texttt{G} can either be the first or the second element of the final group.

If \texttt{G} is the first element in the final group (i.e., the group has the form \texttt{G\_}), then we must select 2 letters from the 6 letters preceding \texttt{G} (i.e., \texttt{A} through \texttt{F}) and place them together in one of the first five groups. The remaining five groups' second letters can then be arranged arbitrarily. This gives a total count of:
\[
\binom{6}{2} \times 5! = 15 \times 120 = 1800.
\]

If \texttt{G} is the second element in the final group (i.e., the group has the form \texttt{F,G}), then the six letters before \texttt{G} (\texttt{A} through \texttt{F}) must be used as the leading letters of the first five groups, with one of them (specifically \texttt{F}) paired with \texttt{G}. Since the first letters of each group are fixed, we only need to consider the permutations of the second letters, yielding:
\[
5! = 120.
\]

Combining both configurations, the total number of favorable arrangements is:
\[
\binom{6}{2} \times 5! + 5! = 1800 + 120 = 1920.
\]

The total number of all possible valid groupings is:
\[
11 \times 9 \times 7 \times 5 \times 3 = 10395,
\]
since each group must be in lexicographical order and consists of two distinct letters selected from the remaining unused ones.

Thus, the final probability is:
\[
\frac{1920}{10395} = \frac{128}{693}.
\]

As required by the problem, summing the numerator and denominator yields the final answer:
\[
128 + 693 = 821.
\]

\end{humanbox}

\newpage
\subsection{AIME\_1\_6}
\begin{questionbox}
\textit{An isosceles trapezoid has an inscribed circle tangent to each of its four sides. The radius of the circle is $3$, and the area of the trapezoid is $72$. Let the parallel sides of the trapezoid have lengths $r$ and $s$, with $r \neq s$. Find $r^2+s^2$}
\end{questionbox}

\begin{R1box}
\textbf{\Large R1 Reasoning}

\vspace{1\baselineskip}

In \textbf{Steps 1–3}, R1 starts by analyzing the problem of an isosceles trapezoid with an inscribed circle. R1 recalls the property that for a quadrilateral with an inscribed circle, the sum of the lengths of opposite sides is equal. In the case of the isosceles trapezoid with bases $r$ and $s$, R1 infers that each leg has a length of $(r + s)/2$.

In \textbf{Steps 4–6}, R1 further elaborates on the property of the tangential quadrilateral and calculates the semiperimeter.R1 determines that the length of each leg $l = 12$ and $r + s=24$.

In \textbf{Steps 7–8}, R1 calculates the height of the trapezoid. From the area formula $(1/2)(r + s)h = 72$ and $r + s = 24$, R1 finds $h = 6$. R1 then analyzes the relationship between the radius and height and conclude that the diameter of the inscribed circle (which is 6) is equal to the height of the trapezoid.

In \textbf{Steps 9–19}, R1 focuses on finding $r^{2}+s^{2}$. R1 plans to find $r - s$ since R1 already knows $r + s = 24$. By dropping perpendiculars from the ends of the shorter base to the longer base in the isosceles trapezoid, R1 forms right - triangles and use the Pythagorean theorem $((r - s)/2)^{2}+6^{2}=12^{2}$. After a series of calculations, R1 finds that $r - s = 12\sqrt{3}$.

In \textbf{Steps 20–25}, R1 solves the system of equations $\begin{cases}r + s = 24\\r - s = 12\sqrt{3}\end{cases}$ to get $r = 12 + 6\sqrt{3}$ and $s = 12 - 6\sqrt{3}$.

In \textbf{Steps 26–32}, R1 computes $r^{2}+s^{2}$ by squaring $r$ and $s$ separately and then adding the results. R1 gets $r^{2}+s^{2}=(252 + 144\sqrt{3})+(252 - 144\sqrt{3}) = 504$ and then plan to check the result.

In \textbf{Steps 33–45}, R1 verifies the use of the Pythagorean theorem step and re - calculates $r$, $s$, $r^{2}$, $s^{2}$, and $r^{2}+s^{2}$, confirming that the result of $r^{2}+s^{2}=504$ is correct.

In \textbf{Steps 46–57}, R1 explores an alternative method to calculate $r^{2}+s^{2}$ using the identity $(r + s)^{2}+(r - s)^{2}=2(r^{2}+s^{2})$. With $r + s = 24$ and $r - s = 12\sqrt{3}$, R1 calculates $24^{2}+(12\sqrt{3})^{2}=2(r^{2}+s^{2})$ and again get $r^{2}+s^{2}=504$, finally presenting the answer as $\boxed{504}$. 
\end{R1box}

\begin{humanbox}
\textbf{\Large Human Reasoning}

\vspace{1\baselineskip}

To begin with, because of tangents from the circle to the bases, the height is $2 \cdot 3 = 6$. The formula for the area of a trapezoid is $\frac{h(b_{1}+b_{2})}{2}$. Plugging in our known values we have
$$\frac{6(r+s)}{2} = 72.$$

$$r+s = 24.$$

Next, we use Pitot's Theorem which states for tangential quadrilaterals $AB+CD=AD+BC$. Since we are given $ABCD$ is an isosceles trapezoid we have $AD=BC=x$. Using Pitot's we find,

$$AB+CD=r+s=2x=24.$$

$$x=12.$$

Finally we can use the Pythagorean Theorem by dropping an altitude from D,
$$\left(\frac{r-s}{2}\right)^{2}+6^{2}=12^{2}.$$
$$
\left(\frac{r-s}{2}\right)^{2}=108.$$
$$(r-s)^{2}=432.$$
Noting that $\frac{(r+s)^{2}+(r-s)^{2}}{2}=r^{2}+s^{2}$ we find,
$$
\frac{(24^{2}+432)}{2}=\boxed{504}$$

\end{humanbox}

\newpage
\subsection{AIME\_1\_7}
\begin{questionbox}
\textit{The twelve letters $A$,$B$,$C$,$D$,$E$,$F$,$G$,$H$,$I$,$J$,$K$, and $L$ are randomly grouped into six pairs of letters. The two letters in each pair are placed next to each other in alphabetical order to form six two-letter words, and then those six words are listed alphabetically. For example, a possible result is $AB$, $CJ$, $DG$, $EK$, $FL$, $HI$. The probability that the last word listed contains $G$ is $\frac mn$, where $m$ and $n$ are relatively prime positive integers. Find $m+n$.}
\end{questionbox}

\begin{R1box}
\textbf{\Large R1 Reasoning}

\vspace{1\baselineskip}

In \textbf{Steps 1–9}, R1 begins by understanding the structural decomposition of the problem. It identifies two primary configurations to consider: the case where letter \texttt{G} is placed at the first position of the last group, and the case where \texttt{G} is placed at the second position of the last group.

In \textbf{Steps 10–11}, R1 reasons that it suffices to focus only on configurations where \texttt{G} appears in the final group, excluding other placements from consideration. This reflects an early stage of problem simplification.

From \textbf{Steps 12–44}, R1 attempts to explicitly analyze all possible permutations involving \texttt{G} being at the second position of the final group. It makes three successive attempts to reason through the remaining positions of the other letters. However, each attempt results in confusion and an inability to complete the enumeration, due to entanglement in its own reasoning steps.

In \textbf{Steps 45–64}, R1 recognizes the recurring failure pattern and backtracks to its mental state at Step 12. It reflects on the limitations of its current reasoning process and attempts to devise a more structured combinatorial approach. Nevertheless, after a brief round of planning, it fails three more times, unable to reach a coherent conclusion.

Only in \textbf{Step 65} does R1 experience a breakthrough. It realizes that a simple case distinction based on whether \texttt{G} is the first or second element in the final group suffices. Once this classification is made, the remaining letters can be handled via straightforward combinatorics.

In \textbf{Steps 65–99}, R1 computes the probability that \texttt{G} is the first element in the final group. It selects 4 letters from the 6 that are lexicographically less than \texttt{G}, and combines them with the 4 letters that come after \texttt{G} to form the remaining 4 groups. The 2 unused letters form a group themselves. This gives a favorable count of:$\binom{6}{4} \times 4!$, out of the total number of valid groupings:$\frac{10!}{2^5 \cdot 5!}.$. R1 calculates the resulting probability as: $\frac{8}{21}$.
It then considers the complementary case where \texttt{G} is the second letter in the final group, with probability \(\frac{5}{11}\), leading to a final answer of:
\[
\frac{8}{21} \times \frac{5}{11} = \frac{40}{231}.
\]

In \textbf{Steps 100–116}, R1 attempts to verify its solution. However, the first attempt fails because R1 misidentifies what should be verified. It computes all possible permutations and finds the result uninformative.

From \textbf{Steps 117–132}, R1 revisits its earlier group count and performs a more targeted verification. This time, it uses the known total number of valid groupings to cross-check the derived probability, successfully confirming the result.

\end{R1box}

\begin{wrongbox}
\textbf{\Large Failure Reasoning}

\vspace{1\baselineskip}

The reasoning failures within the R1 model are primarily driven by three structural pathologies: \textbf{incomplete case execution}, \textbf{combinatorial redundancy}, and the \textbf{absence of targeted validation}. Despite a correct initial classification, identifying that the element $G$ could occupy either the first or second position within the terminal group, the model systematically omitted the latter branch from its subsequent calculations, leading to a truncated probabilistic analysis. Furthermore, within the executed branch, the model exhibited a lack of combinatorial rigour by overcounting preceding group arrangements; this error stemmed from the redundant inclusion of permutations already fixed by prior selections, thereby distorting the final probability measure. Ultimately, the trajectory failed to converge due to a lack of post-computational reflection. The model failed to revisit its foundational assumptions or verify the completeness of its case partitioning, demonstrating that without functional meta-cognitive steering, the reasoning process remains susceptible to structural hallucinations despite local deductive accuracy.

\end{wrongbox}
\begin{humanbox}
\textbf{\Large Human Reasoning}

\vspace{1\baselineskip}

A human approaching this problem would similarly begin by considering the possible placements of the letter \texttt{G} in the final group. There are two primary configurations: \texttt{G} can either be the first or the second element of the final group.

If \texttt{G} is the first element in the final group (i.e., the group has the form \texttt{G\_}), then we must select 2 letters from the 6 letters preceding \texttt{G} (i.e., \texttt{A} through \texttt{F}) and place them together in one of the first five groups. The remaining five groups' second letters can then be arranged arbitrarily. This gives a total count of:
\[
\binom{6}{2} \times 5! = 15 \times 120 = 1800.
\]

If \texttt{G} is the second element in the final group (i.e., the group has the form \texttt{F,G}), then the six letters before \texttt{G} (\texttt{A} through \texttt{F}) must be used as the leading letters of the first five groups, with one of them (specifically \texttt{F}) paired with \texttt{G}. Since the first letters of each group are fixed, we only need to consider the permutations of the second letters, yielding:
$5! = 120.$ Combining both configurations, the total number of favorable arrangements is:$
1800 + 120 = 1920.$

The total number of all possible valid groupings is: $11 \times 9 \times 7 \times 5 \times 3 = 10395,$
since each group must be in lexicographical order and consists of two distinct letters selected from the remaining unused ones.Thus, the final probability is:
\[
\frac{1920}{10395} = \frac{128}{693}.
\]

As required by the problem, summing the numerator and denominator yields the final answer:
\[
128 + 693 = 821.
\]

\end{humanbox}

\newpage
\subsection{AIME\_1\_8}
\begin{questionbox}
\textit{Let $k$ be a real number such that the system\begin{align*} &|25+20i-z|=5\\ &|z-4-k|=|z-3i-k| \\ \end{align*}has exactly one complex solution $z.$ The sum of all possible values of $k$ can be written as $\dfrac{m}{n},$ where $m$ and $n$ are relatively prime positive integers. Find $m+n.$ Here $i=\sqrt{-1}.$}
\end{questionbox}

\begin{R1box}
\textbf{\Large R1 Reasoning}

\vspace{1\baselineskip}

In \textbf{Steps 1--4}, R1 begins by interpreting the problem geometrically. It identifies the first equation $|25 + 20i - z| = 5$ as a circle centered at $(25, 20)$ with radius $5$, and the second equation $|z - 4 - k| = |z - 3i - k|$ as a condition to be translated into Cartesian coordinates.

In \textbf{Steps 5--27}, R1 rewrites the second equation in terms of $x$ and $y$ (where $z = x + yi$), squares both sides to eliminate square roots, and simplifies it to the linear equation $8x - 6y = 7 + 8k$. This represents a straight line in the plane. R1 then recognizes that the problem reduces to finding values of $k$ for which this line is tangent to the circle.

From \textbf{Steps 28--56}, R1 applies the distance formula from the circle's center to the line, setting it equal to the radius $5$. This yields the equation $|73 - 8k| = 50$, and find two values for $k$:  
\[
k = \frac{23}{8} \quad \text{and} \quad k = \frac{123}{8}.
\]

In \textbf{Steps 57--61}, R1 sums these values to obtain $\frac{73}{4}$, confirms the fraction is irreducible, and concludes $m + n = 77$.

From \textbf{Steps 62--110}, R1 revisits the algebraic derivation of the line equation to confirm no errors were made. It rechecks the simplification of $|z - 4 - k| = |z - 3i - k|$ and reaffirms the distance condition, ultimately validating the results.

In \textbf{Steps 111--132}, R1 considers an alternative approach by solving the system of equations explicitly. It substitutes $y = \frac{8x - 7 - 8k}{6}$ into the circle's equation, derives a quadratic in $x$, and sets its discriminant to zero.

From \textbf{Steps 133--208}, R1 meticulously expands and simplifies the discriminant condition, arriving at the quadratic equation $64k^2 - 1168k + 2829 = 0$. Solving this yields the same solutions for $k$, reinforcing the correctness of the earlier geometric approach. The final answer is:  
\[
\boxed{77}.
\]

\end{R1box}

\begin{humanbox}
\textbf{\Large Human Reasoning}

\vspace{1\baselineskip}

\begin{figure}[H]
    \centering
    \includegraphics[width=0.5\linewidth]{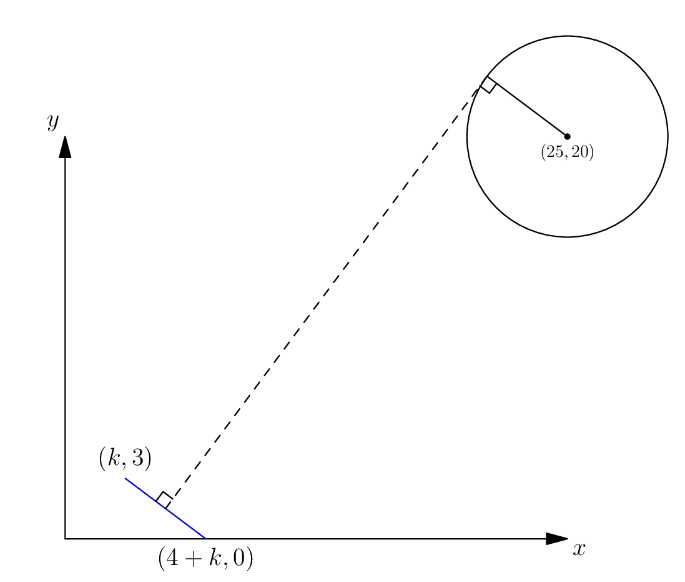}
\end{figure}

The complex number $z$ must satisfy the following conditions on the complex plane:

\begin{enumerate}
    \item The magnitude between $z$ and $(25, 20)$ is $5$. This can be represented by drawing a circle with center $(25, 20)$ and radius $5$.

    \item It is equidistant from the points $(4 + k, 0)$ and $(k, 3)$. Hence it must lie on the perpendicular bisector of the line connecting these points.
\end{enumerate}

For $z$ to have one solution, the perpendicular bisector of the segment connecting the two points must be tangent to the circle. This bisector must pass the midpoint, $(2 + k, \frac{3}{2})$, and have slope $\frac{4}{3}$. The segment connecting the point of tangency to the center of the circle has slope $-\frac{3}{4}$, meaning the points of tangency can be $(29, 17)$ or $(21, 23)$. Solving the equation for the slope of the perpendicular bisector gives
$$\frac{\frac{3}{2} - 23}{k + 2 - 21} = \frac{4}{3}$$
or
$$
\frac{\frac{3}{2} - 17}{k + 2 - 29} = \frac{4}{3},$$
giving $k = \frac{23}{8}$ or $\frac{123}{8}$, having a sum of $\frac{73}{4}\Longrightarrow\boxed{077}$

\end{humanbox}

\newpage
\subsection{AIME\_1\_9}
\begin{questionbox}
\textit{The parabola with equation $y = x^2 - 4$ is rotated $60^\circ$ counterclockwise around the origin. The unique point in the fourth quadrant where the original parabola and its image intersect has $y$-coordinate $\frac{a - \sqrt{b}}{c}$, where $a$, $b$, and $c$ are positive integers, and $a$ and $c$ are relatively prime. Find $a + b + c$.}
\end{questionbox}

\begin{R1box}
\textbf{\Large R1 Reasoning}

\vspace{1\baselineskip}

In \textbf{Steps 1–16}, R1 Assumes that the coordinates of any point on the original parabola are $(x, y)$, when the parabola is rotated counterclockwise by $60^{\circ}$, the coordinates of that point become $(X, Y)$. R1 uses rotation matrix to find the relationship between two coordinates:

\[
\left\{
\begin{aligned}
    x & =  X\cos(-60^{\circ})-Y\sin(-60^\circ) \\
    y & =  X\sin(-60^{\circ})+Y\cos(-60^\circ) \\
\end{aligned}
\right.
\]

Then, R1 substitutes the above results in the parabolic expression $y=x^2-4$ to obtain:

\[
-\frac{\sqrt{3}}{2}X + \frac{1}{2}Y = (\frac{1}{2}X + \frac{\sqrt{3}}{2}Y)^2 - 4\tag{1}
\]

In \textbf{Steps 17–82}, R1 notes that if $(X, Y)$ is the intersection of two parabolas, then $(X, Y)$ also satisfies the relationship:

\[
Y=X^2-4\tag{2}
\]

From equations (1) and (2), R1 obtains:

\[
3X^4 + 2\sqrt{3} X^3 - 25X^2 - 6\sqrt{3}X + 40 = 0\tag{3}
\]

In \textbf{Steps 83–84}, R1 wishes to parameterize equation (3) at $t=\sqrt{3}X$ for ease of subsequent calculations.

In \textbf{Steps 85–87}, R1 attempts to find an obvious solution in equation (3), but does not find it. Therefore, it thinks of factoring the left side of the equation and first considered decomposing it into four linear factors.

In \textbf{Steps 88–103}, R1 substitutes $t=\sqrt{3}X$ into equation (3) to obtain:

\[
t^4 + 2t^3 -25t^2 -18t +120 = 0\tag{4}
\]

In \textbf{Steps 104–117}, R1 reaches the conclusion that the possible rational roots of (4) are factors of 120 (both positive and negative factors need to be considered). R1 only tries to calculate a few factors and found that none of them were roots of equation (4). Therefore, it is considered that the left side of equation (4) can only be decomposed into the product of two quadratic factors.

In \textbf{Steps 118–159}, using the method of undetermined coefficients, R1 assumes the existence of parameters $a,b,c,d$ that satisfy the equation:

\[
t^4 + 2t^3 -25t^2 -18t +120=(t^2+at+b)(t^2+ct+d)\tag{5}
\]

\end{R1box}
\begin{R1box}
R1 simplifies equation (5) to obtain:

\[
\left\{
\begin{aligned}
     & a+c=2 \\
     & b+d+ac=-25 \\
     & ad+bc=-18\\
     & bd=120\\
\end{aligned}
\right.
\]
R1 only considers the case where $a$, $b$, $c$, and $d$ are rational numbers. Starting from bd=120, R1 calculates that:

\[
a=-1,b=-10,c=3,d=-12
\]

By bringing the parameter results back to equation (5) for checking, R1 obtains:

\[
t^4 + 2t^3 -25t^2 -18t +120=(t^2-t-10)(t^2+3t-12)=0\tag{6}
\]

In \textbf{Steps 160–186}, R1 solves equation (6) and uses $t=\sqrt{3}x$ to obtain:

\[
X=\frac{1\pm\sqrt{41}}{2\sqrt{3}},\frac{-3\pm\sqrt{57}}{2\sqrt{3}}
\]

Through approximate calculation, R1 checks whether all intersection coordinates meet the condition "$x>0, y<0$", so as to obtain the unique fourth quadrant intersection:

\[
(X,Y)=(\frac{-3+\sqrt{57}}{2\sqrt{3}},\frac{3-\sqrt{57}}{2})\tag{7}
\]

Thus, R1 gets the answer to the question:
\[
a=3,b=57,c=2,a+b+c=\boxed{62}.
\]

In \textbf{Steps 187–201}, R1 first checks the value of result (7) to see if it meets equations (2) and (4), then confirms that result (7) is the only fourth quadrant intersection, and finally confirms that the values of $a,b,c$ and their sum are correct.

\end{R1box}

\begin{humanbox}
\textbf{\Large Human Reasoning}

\vspace{1\baselineskip}

A human approaching this problem would similarly begin by converting rotation to symmetry. Notice that the angle made by the intersection point with the y-axis is $60/2 = 30$ degrees by symmetry, so the point lies on $y = -\sqrt{3}x$. Set up simultaneous equations: 

\[ 
\left\{
\begin{aligned}
    y & =  -\sqrt{3}x \\
    y & =  x^2-4 \\
\end{aligned}
\right.
\]

Solving, we have:

\[
x^2+\sqrt{3}-4=0 \Rightarrow x=\frac{-\sqrt{3}\pm\sqrt{19}}{2}
\]

The intersection point in the fourth quadrant has a positive x-coordinate. So, we take the positive root, which is $\frac{\sqrt{19}-\sqrt{3}}{2}$, substitute in the equation $y=-\sqrt{3}x$, we have:

\[
y=\frac{3-\sqrt{57}}{2}
\]

Thus:

\[
a=3, b=57, c=2
\]

As required by the problem, the final answer is as follows:

\[
a+b+c=\boxed{62}.
\]

\end{humanbox}

\newpage
\subsection{AIME\_1\_10}
\begin{questionbox}
\textit{The $27$ cells of a $3 \times 9$ grid are filled in using the numbers $1$ through $9$ so that each row contains $9$ different numbers, and each of the three $3 \times 3$ blocks heavily outlined in the example below contains $9$ different numbers, as in the first three rows of a Sudoku puzzle. The number of different ways to fill such a grid can be written as $p^a \cdot q^b \cdot r^c \cdot s^d$ where $p$, $q$, $r$, and $s$ are distinct prime numbers and $a$, $b$, $c$, $d$ are positive integers. Find $p \cdot a + q \cdot b + r \cdot c + s \cdot d$.}
\end{questionbox}

\begin{R1box}
\textbf{\Large R1 Reasoning}

\vspace{1\baselineskip}

In \textbf{Steps 1–46}, R1 defines the constraints: no repeated numbers per block/row, and establishes row-wise discussion order.

In \textbf{Steps 47–48}, ‌R1 realizes that the sorting method of the first row has 9! types (full arrangement of nine numbers).‌

In \textbf{Steps 49–114}, ‌R1 segments the first-row digit sequence into three distinct ordered triples designated $A$, $B$, and $C$. ‌Similarly, R1 designates the three 3-element tuples in the second row as $D$, $E$, and $F$.‌

\begin{center}
\setlength{\tabcolsep}{12pt} 
\renewcommand{\arraystretch}{1.2}
\begin{tabular}{|*{3}{c|}} 
\hline
$A$ & $B$ & $C$ \\ \hline
$D$ & $E$ & $F$ \\ \hline
$G$ & $H$ & $I$ \\ \hline
\end{tabular}
\end{center}

Under given constraints, R1 verifies $D \subset B\cup C, E \subset A\cup C, F \subset A\cup B$, and for the triple $A$, assume $a=|E \cup A|$, thus $3-a=|F \cup A|$; for the triple $B$, assume $b=|D \cup B|$, thus $3-b=|F \cup B|$; for the triple $C$, assume $c=|D \cup C|$, thus $3-c=|E \cup C|$. ‌Since $D$, $E$, and $F$ are all triples, R1 obtain the system of equations:‌

\[ 
\left\{
\begin{aligned}
    & b + c =  3 \\
    & a = c \\
    & a + b =  3 \\
\end{aligned}
\right. \Rightarrow
\left\{
\begin{aligned}
    & a = c \\
    & a + b =  3 \\
\end{aligned}
\right.
\]

Therefore, for each $a\in \{0,1,2,3\}$, the number of ways to choose the assignments is: $C_{3}^{a}C_{3}^{b}C_{3}^{c}=C_{3}^{a}C_{3}^{3-a}C_{3}^{a}$ (R1 chooses $a$ elements from $A$ to put into $E$, chooses $b$ elements from $B$ to put into $D$ and chooses $c$ elements from $C$ to put into $D$). ‌Finally, R1 sums over $a$ to obtain how many ways to assign the second row:‌

\[
\sum_{a=0}^{3}C_{3}^{a}C_{3}^{3-a}C_{3}^{a}=56.
\]

R1 performs a check on this answer and confirms it is correct (in reality, this answer is incorrect, as R1 overlooks the order of elements within the triples, resulting in undercounting).

\end{R1box}

\begin{R1box}

In \textbf{Steps 115-119}, R1 claims that when the first two rows of triples are fixed, the third row is uniquely determined (this is actually incorrect; it is not uniquely determined, as the order of the three elements within the tuple is variable).

In \textbf{Steps 120-130}, R1 reviews the entire process above and ‌concludes that the answer is $9!\times56‌$ (this answer is naturally incorrect). 

In \textbf{Steps 131-147}, R1 realizes the counting error in the second row (namely that the order of elements within the triple is variable), and calculates that:

\[
56\times(3!)^3
\]

In \textbf{Steps 148–268}, Through repeated checks, R1 proposes numerous peculiar ideas yet failed to rectify the counting issue in the third row; persistently viewing it as uniquely determined and overlooking the variability in element permutations within the triple G, H, I, ultimately arriving at the result:

\[
9!\times56\times(3!)^3\times1=2^{13} \times 3^7 \times 5 \times 7^2
\]

Based on the requirements of the problem statement, the final answer is:
$2\times13+3\times7+5\times1+7\times2=\boxed{61}$.

\end{R1box}

\begin{wrongbox}
\textbf{\Large Failure Reasoning}

\vspace{1\baselineskip}

The failure in R1's reasoning process can be attributed to two main issues:

\textbf{1. ‌Numerous ineffective logical reflections.}  
R1 initially ‌devotes‌ excessive verbiage to reiteration, essentially just ‌restating‌ the problem's constraint rules without substantive contributions.

\textbf{2. The valuable insights achieved through reflection have not been effectively transferred.}  In subsequent sections, R1 successfully ‌identifies‌ the calculation error in the ‌second row‌ but ‌does not extend this verification‌ to the ‌third row‌, despite the potential for similar mistakes. Consequently, the valuable insights derived from reflection ‌have not been prioritized‌.

\end{wrongbox}
\begin{humanbox}
\textbf{\Large Human Reasoning}

\vspace{1\baselineskip}

First, we consider filling the 9 empty cells in the first row with the digits 1–9. Clearly, this constitutes a permutation, yielding a total of 9! possible filling methods. For any given one of these 9! methods, define the labels for the digits in the first row of the first, second, and third blocks as $AAA$, $BBB$, and $CCC$ respectively.

\begin{center}
\setlength{\tabcolsep}{0pt}
\renewcommand{\arraystretch}{1.2}
\begin{tabular}{|*{9}{S|}}
\hline
$A$ & $A$ & $A$ & $B$ & $B$ & $B$ & $C$ & $C$ & $C$ \\ \hline
 & & & & & & & & \\ \hline
 & & & & & & & & \\ \hline
\end{tabular}
\end{center}

‌Secondly, we fill the nine empty cells in the second row. We focus on the three empty cells in the third block. There are two types of filling methods here.

Case 1: We can fill either $AAA$ or $BBB$ into row 2 of the third block. 

Demonstration uses $AAA$. Since the problem requires all nine digits in each block to be distinct, the remaining six cells must be filled in the pattern $BBBCCC$. At this point, the third row can also be filled according to the requirement that all numbers in each block must be unique. Then we have:

\begin{center}
\setlength{\tabcolsep}{0pt}
\renewcommand{\arraystretch}{1.2}
\begin{tabular}{|*{9}{S|}}
\hline
$A$ & $A$ & $A$ & $B$ & $B$ & $B$ & $C$ & $C$ & $C$ \\ \hline
$B$ & $B$ & $B$ & $C$ & $C$ & $C$ & $A$ & $A$ & $A$ \\ \hline
$C$ & $C$ & $C$ & $A$ & $A$ & $A$ & $B$ & $B$ & $B$ \\ \hline
\end{tabular}
\end{center}
We need only focus on the arrangement of three digits within the same block and same row. This constitutes a full permutation with $3!$ possibilities. Since there are six such row-block segments, the total permutations amount to $(3!)^6$. We demonstrated the $AAA$ case as an example; the $BBB$ case follows similarly. Combining this with the first step, we obtain the count for Case 1:$9! \times (3!)^6 \times 2$.

Case 2: We can fill either $AAB$ or $ABB$ into row 2 of the third block.(Here we disregard the order of the three digits and treat them solely as a distinction between ‌one $B$ versus two $B$s‌.)

Demonstration uses $AAB$. Due to the constraint that digits must be unique within each block, the remaining single $A$ is thus forced into the second block, while the two remaining $B$s must be placed into the first block. The three C digits are freely assigned to the remaining three cells.

\begin{center}
\setlength{\tabcolsep}{0pt}
\renewcommand{\arraystretch}{1.2}
\begin{tabular}{|*{9}{S|}}
\hline
$A$ & $A$ & $A$ & $B$ & $B$ & $B$ & $C$ & $C$ & $C$ \\ \hline
$B$ & $B$ & $C$ & $A$ & $C$ & $C$ & $A$ & $A$ & $B$ \\ \hline
 & & & & & & & & \\ \hline
\end{tabular}
\end{center}

Fundamentally, this involves assigning ‌one $B$ to the second row of Block 3‌, ‌one $A$ to the second row of Block 2‌, and ‌one $C$ to the second row of Block 1‌. This allocation yields $3^3$ possible ways‌. Subsequently, accounting for the permutations of the three digits within each block's row, there are ‌$(3!)^3$ distinct arrangements‌.

\end{humanbox}

\begin{humanbox}
The last row must be filled under the constraint that ‌all nine digits within each block remain distinct‌. Accounting for the permutations of the three digits within each block's row, there are $(3!)^3$ distinct arrangements‌ possible.

\begin{center}
\setlength{\tabcolsep}{0pt}
\renewcommand{\arraystretch}{1.2}
\begin{tabular}{|*{9}{S|}}
\hline
$A$ & $A$ & $A$ & $B$ & $B$ & $B$ & $C$ & $C$ & $C$ \\ \hline
$B$ & $B$ & $C$ & $A$ & $C$ & $C$ & $A$ & $A$ & $B$ \\ \hline
$C$ & $C$ & $B$ & $C$ & $A$ & $A$ & $B$ & $B$ & $A$ \\ \hline
\end{tabular}
\end{center}

The $ABB$ case follows similarly. Combining this with the first step, we obtain the count for Case 2:$9! \times 3^3 \times (3!)^3\times (3!)^3 \times 2$. We add the number of arrangements from Case1 and Case2, ‌yielding the total number of arrangements for the problem as‌:
\[
2^{16}\cdot3^{10}\cdot5\cdot7^2
\]
The answer to this question is $2\times16+3\times10+5\times1+7\times2=\boxed{81}$.
\end{humanbox}

\newpage
\subsection{AIME\_1\_11}
\begin{questionbox}
\textit{A piecewise linear function is defined by\[f(x) = \begin{cases} x & \text{if } x \in [-1, 1) \\ 2 - x & \text{if } x \in [1, 3)\end{cases}\]and $f(x + 4) = f(x)$ for all real numbers $x.$ The parabola $x = 34y^2$ intersects the graph of $f(x)$ at finitely many points. The sum of the $y$-coordinates of these intersection points can be expressed in the form $\tfrac{a + b\sqrt c}d,$ where $a, b, c$ and $d$ are positive integers, $a, b,$ and $d$ has greatest common divisor equal to $1,$ and $c$ is not divisible by the square of any prime. Find $a + b + c + d.$}
\end{questionbox}

\begin{R1box}
\textbf{\Large R1 Reasoning}

\vspace{1\baselineskip}

In \textbf{Steps 1--22}, R1 begins by analyzing the periodic piecewise function $f(x)$. It carefully examines the definition of $f(x)$ in different intervals and its periodicity, establishing that the function repeats every 4 units with a sawtooth pattern between $-1$ and $1$.

In \textbf{Steps 23--38}, R1 considers the intersection between the parabola $x = 34y^2$ and the periodic function $f(x)$. It recognizes that since $f(x)$ is bounded between $-1$ and $1$, the intersections must occur where $|y| \leq 1$, limiting $x$ to the interval $[0,34]$.

From \textbf{Steps 39--59}, R1 develops a systematic approach to handle the periodicity. It divides the problem into cases based on the period segments, identifying that for each integer $k$, there are ascending and descending segments where different equations apply.

In \textbf{Steps 60--94}, R1 formulates the exact equations to solve for each case:
\begin{itemize}
    \item For ascending segments: $34y^2 - y - 4k = 0$
    \item For descending segments: $34y^2 + y - (4k + 2) = 0$
\end{itemize}
It also establishes the valid ranges for solutions in each case ($y \in [-1,1)$ for ascending and $y \in (-1,1]$ for descending segments).

In \textbf{Steps 95--140}, R1 begins solving these equations for successive values of $k$. It uses Vieta's formulas to find that:
\begin{itemize}
    \item The sum of roots for ascending segments is always $\frac{1}{34}$
    \item The sum of roots for descending segments is always $-\frac{1}{34}$
\end{itemize}

From \textbf{Steps 141--188}, R1 carefully evaluates each case from $k=0$ to $k=8$, discovering that:
\begin{itemize}
    \item For $k=0$ to $k=8$ in ascending segments, all roots are valid
    \item For $k=0$ to $k=7$ in descending segments, all roots are valid
    \item For $k=8$ in descending segments, only one root is valid
\end{itemize}

\end{R1box}

\begin{R1box}
In \textbf{Steps 189--220}, R1 computes the total sum of all valid $y$-coordinates:
\[
\text{Ascending sum} = \frac{9}{34} = \frac{18}{68}
\]
\[
\text{Descending sum} = \frac{-17 + 5\sqrt{185}}{68}
\]
\[
\text{Total sum} = \frac{1 + 5\sqrt{185}}{68}
\]
and get the solution 
\[a+b+c+d=1+5+185+68=259\]

In \textbf{Steps 221--236}, R1 verifies that this expression is in the required form $\frac{a + b\sqrt{c}}{d}$ with:
\[
a = 1,\ b = 5,\ c = 185,\ d = 68
\]
and confirms that $\gcd(1,5,68) = 1$ and $185$ is square-free. The final answer is computed as:
\[
a + b + c + d = 1 + 5 + 185 + 68 = \boxed{259}
\]

\end{R1box}

\begin{humanbox}
\textbf{\Large Human Reasoning}

\vspace{1\baselineskip}

Note that $f(x)$ consists of lines of the form $y = x - 4k$ and $y = 4k + 2 - x$ for integers $k$. In the first case, we get $34y^2 = y - 4k$ and the sum of the roots is $\frac{1}{34}$ by Vieta. In the second case, we similarly get a sum of $-\frac{1}{34}$. Thus pairing $4k$ and $4k + 2$ gives a $y$-coordinate sum of $0$. This process of pairing continues until we get to $k = 8$. Then $y = x - 32$ behaves exactly as we expect, with a sum of $\frac{1}{34}$. However, $y = 34 - x$ is where things start becoming fishy, since there is one root with absolute value less than $1$ and one with absolute value greater than $1$. We get

$34 - 34y^2 = y.$

Solving with the quadratic formula (clear to take the positive root) gives

$y = \frac{-1 \pm \sqrt{68^2 + 1}}{68} = \frac{-1 + 5\sqrt{185}}{68}.$

Adding our $\frac{1}{34}$ from earlier gives the answer $\frac{1 + 5\sqrt{185}}{68} \implies \boxed{259}$.

\end{humanbox}

\newpage
\subsection{AIME\_1\_12}
\begin{questionbox}
\textit{The set of points in $3$-dimensional coordinate space that lie in the plane $x+y+z=75$ whose coordinates satisfy the inequalities$$x-yz<y-zx<z-xy$$forms three disjoint convex regions. Exactly one of those regions has finite area. The area of this finite region can be expressed in the form $a\sqrt{b},$ where $a$ and $b$ are positive integers and $b$ is not divisible by the square of any prime. Find $a+b.$}
\end{questionbox}

\begin{R1box}
\textbf{\Large R1 Reasoning}

\vspace{1\baselineskip}

In \textbf{Steps 1--32}, R1 analyzes the problem by rewriting the original inequalities $x - yz < y - zx < z - xy$ into simpler forms. It derives $(x - y)(1 + z) < 0$ and $(y - z)(1 + x) < 0$, and explores the conditions under which these products are negative by considering sign changes for each factor.

In \textbf{Steps 33--49}, R1 incorporates the plane constraint $x + y + z = 75$ by substituting $z = 75 - x - y$. This reduces the inequalities to two variables: $(x - y)(76 - x - y) < 0$ and $(2y + x - 75)(1 + x) < 0$.

In \textbf{Steps 50--75},  R1 parametrizes the plane using $u = x$ and $v = y$. It defines cases for each inequality in the $uv$-plane. For the first inequality, it specifies Case 1 ($u > v$ and $u + v > 76$) and Case 2 ($u < v$ and $u + v < 76$). For the second inequality, it defines Case A ($2v + u > 75$ and $u < -1$) and Case B ($2v + u < 75$ and $u > -1$).

In \textbf{Steps 76--125}, R1 examines all combinations of the cases. It determines that Case 1 and Case A is impossible due to contradictions. Case 1 and Case B yields an unbounded region extending to infinity. Case 2 and Case A also produces an unbounded region. 


In \textbf{Steps 126--180}, R1 focuses on Case 2 and Case B, deriving the constraints $-1 < u < 25$ and $u < v < (75 - u)/2$. It identifies this region as a triangle in the $uv$-plane with vertices at $(-1, -1)$, $(-1, 38)$, and $(25, 25)$.

In \textbf{Steps 181--218}, R1 calculates the area of the triangle in the $uv$-plane as 507 using the shoelace formula. It then recognizes that the actual area in 3D space must account for the scaling factor from the plane parametrization. The cross product of the partial derivatives $\partial \mathbf{r}/\partial u = (1, 0, -1)$ and $\partial \mathbf{r}/\partial v = (0, 1, -1)$ has magnitude $\sqrt{3}$, so the actual area is $507 \sqrt{3}$. R1 expresses this as $a \sqrt{b}$ with $a = 507$ and $b = 3$ (square-free), and computes $a + b = 510$.

In \textbf{Steps 219--471}, R1 verifies the solution by confirming the existence of three disjoint convex regions: one bounded (from Case 2 and Case B) and two unbounded (from Case 1 and Case B, and Case 2 and Case A). It reaffirms the area calculation and concludes with the final answer: 510.

\end{R1box}

\begin{humanbox}
\textbf{\Large Human Reasoning}

\vspace{1\baselineskip}

Rewriting we have $z = 75 - x - y$.
From the inequality $x - yz < y - zx$ we can rewrite to get,
$$x - y(75 - x - y) < y - x(75 - x - y).$$
$$
76x - 76y + y^2 - x^2 < 0.$$
$$(x + y + 76)(x - y) < 0.$$
Similarly from the inequality $y - zx < z - xy$ we rewrite to get,
$$
y - x(75 - x - y) < (75 - x - y) - xy.$$
$$2y + 2xy + x^2 - 74x - 75 < 0.$$
$$
(x + 1)(2y + x - 75) < 0.$$
Our next step is a visual which can be seen by roughly graphing the two inequalities. The first inequality is similar to a bow tie(you have to graph to see it lol) with bounds of $x - y = 0$ and $76 + x + y = 0$. The second one is a bow tie with edges of $x + 1 = 0$ and $2y + x - 75 = 0$.
Here is the region of our solutions.

\begin{figure}[H]
    \centering
    \includegraphics[width=0.5\linewidth]{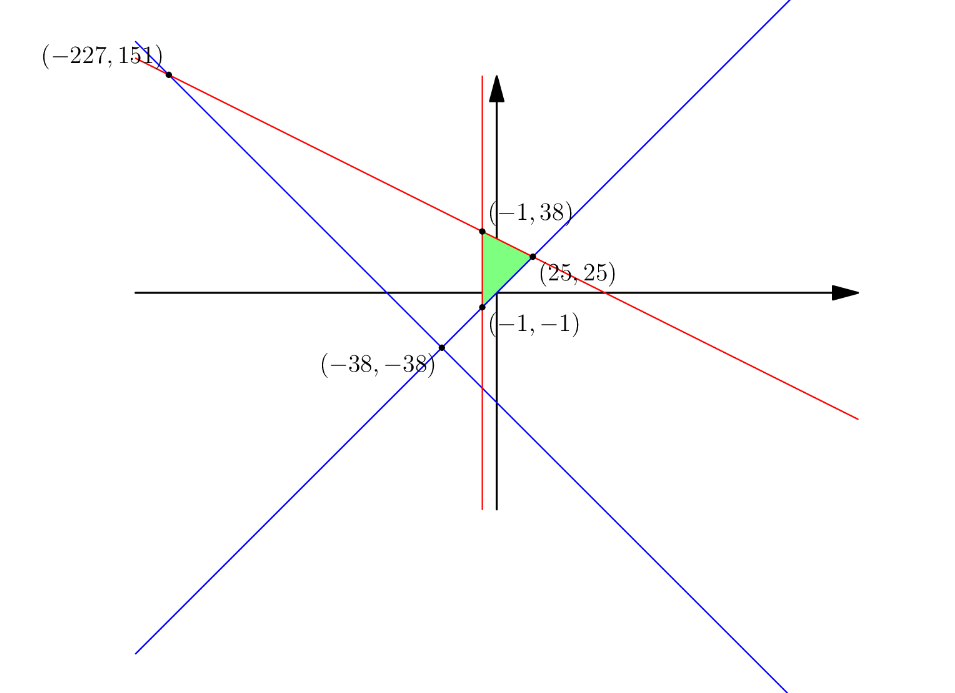}
\end{figure}
It is simple to find the intersections of those which are $(-1, 38, 38), (25, 25, 25)$ and $(-1, -1, 77)$. The sidelengths of this triangle are $39\sqrt{2}$, $26\sqrt{6}$, $13\sqrt{6}$ which is a $30-60-90$ degree triangle, gives us an area of
$$\frac{1}{2}(39\sqrt{2})(13\sqrt{6})=507\sqrt{3} \Longrightarrow \boxed{510}$$
\end{humanbox}

\subsection{AIME\_1\_13}
\begin{questionbox}
\textit{Alex divides a disk into four quadrants with two perpendicular diameters intersecting at the center of the disk. He draws $25$ more lines segments through the disk, drawing each segment by selecting two points at random on the perimeter of the disk in different quadrants and connecting these two points. Find the expected number of regions into which these $27$ line segments divide the disk.}
\end{questionbox}

\begin{R1box}
\textbf{\Large R1 Reasoning}\vspace{1\baselineskip}

In \textbf{Steps 1–9}, R1 begins by understanding the problem.

In \textbf{Steps 10–62}, R1 transforms the problem to count the intersections of lines: \[\mathbb E[R] =\mathbb E[1 + \sum_c (k_c + 1)] = 1 + \sum_c\mathbb E[k_c + 1] = 28 + \sum_c\mathbb E[k_c],\] where $k_c$ is the number of intersections that chord c has with chords drawn before it.

In \textbf{Steps 63–75}, R1 counts $\mathbb E[k_c]$. It starts by the two diameters with $k_0=0$, $k_1=1$ and produces $4$ regions. 

In \textbf{Steps 76–117}, R1 counts the intersection between lines added and the diameters fixed, but no explicit conclusion is given.

In \textbf{Steps 118–215}, R1 notes the $4$ regions produced by diameters fixed as $1$, $2$, $3$, $4$ and then obtains that the choice of chord is uniformly distributed in the space $\{12,13,14,23,24,34\}$, where $\mathbb P(adjacent)=2/3$ and $\mathbb P(opposite)=1/3$. 

In \textbf{Steps 216–296}, R1 figures out that \[\mathbb E(|l\cap\{diameters fixed\}|)=\frac13\times2+\frac23\times1=\frac43.\]

In \textbf{Steps 297–313}, R1 comes up with that \[\mathbb E[k_j] = 4/3 + (j-1) p\]for the $j$-st lines added, where $p$ is the constant probability that two arbitrary lines added intersect. In \textbf{Steps 314–338}, R1 tries to find $p$ but gives up before checking every cases.

In \textbf{Steps 339–375}, R1 turns to calculate $p$ for the case the lines are not forced to pass two different quadrants, which is totally meaningless. 

In \textbf{Steps 376–498}, R1 claims we need to respect the constraints, but goes again to proof that when choosing the chords freely, the probability for two chords to intersect is $1/3$.
In \textbf{Steps 499–543}, R1 immitates the method above to calculate $p$ when the choice of lines is restricted. It makes a fault to suppose an endpoint $A$ of chords $AB$ meets with the startpoint of the region $1$. 

\end{R1box}

\begin{R1box}

In \textbf{Steps 544–628}, R1 wrongly calculates the distribution of $d=\min\{b,1-b\}$ the scaled length of minor arc given by $AB$, represented by $b$ the scaled coordinate of $B$. In fact, \[d=\frac43\times1_{[0,1/4[}+\frac83\times1_{[1/4,1/2[},\] but R1 insists that $d>1/4$ and wrongly gives that $p=\frac{11}{36}$.

In \textbf{Steps 629–665}, R1 claims the assumption that $A=0$ dosen't effect the result without any concrete evidence. Then it checks $p=\frac{11}{36}$ is right by repeating the wrong inferences.

In \textbf{Steps 666–701}, R1 combines the contents obtained together. \[\mathbb E[k_j] = 2 * (2/3) + (j-1) * (11/36) = 4/3 + 11(j-1)/36,\] so \[E[R] = 4 + \sum_{j=1}^{25} E[k_j + 1] = 4 + \sum_{j=1}^{25} \left(1 + \frac{4}{3} + \frac{11}{36} (j-1) \right)=154.\]
In \textbf{Steps 702–760}, R1 verifies $3$ times the result for $j=1$, no contradiction because it's independent with the wrong value $p$. In \textbf{Steps 761–777}, R1 announces to check $j=2$, but gives up without reason. In \textbf{Steps 778–791}, R1 checks the result by repeating. In \textbf{Steps 792–810}, R1 dose the reasoning as $154$ is between the minimum and the maximum of division. In \textbf{Steps 811–883}, R1 checks $3$ times by repeating, taking $j=1$ and repeating.

\end{R1box}

\begin{wrongbox}
\textbf{\Large Failure Reasoning}\vspace{1\baselineskip}

The failure in R1's reasoning process can be attributed to three main issues: no ability to deal with the cases of relation between two arbitrary lines, add invalid assumptions during the analysis and no ability to do the real check but just repeat.

\end{wrongbox}

\begin{humanbox}
\textbf{\Large Human Reasoning}\vspace{1\baselineskip}

A human approaching this problem would begin by formalizing the problem. Note $D$ the disk and $d_x$, $d_y$ the two perpendicular diameters. 

Set $X_n$ the random variable representing the number of regions when $n$ independent random lines $l_1$, $\cdots$, $l_n$ are added to the disk with quadrants. So $X_0=4$ and by observation \[\mathbb E[X_{n+1}-X_n]=\mathbb E[T_{n+1}]+1,\]where $T_{n+1}$ is the random variable representing the number of intersections with $l_{n+1}$ with $l_1$, $\cdots$, $l_n$ and $p_x$, $p_y$ in the disk.

\end{humanbox}
\begin{humanbox}
Define a function \[T(l,l')=\begin{cases}1,&l\cap l'\subset D\\
0,&l\cap l'\not\subset D\end{cases}\] and then \begin{align}\mathbb E[T_{n+1}]=&\mathbb E[T(d_x,l_{n+1})]+\mathbb E[T(d_y,l_{n+1})]+\sum_{i=1}^n\mathbb E[T(l_{i},l_{n+1})]\\
=&\mathbb E[T(d_x,L)]+\mathbb E[T(d_y,L)]+n\mathbb E[T(L,L')],\end{align}
where $L$ and $L'$ are independent random lines passing through $D$.

To calculate the terms, note the quadrants in ordre as $1$, $2$, $3$, $4$, then $L\cap D$ is uniformly distributed in the quadrants \[S=\{\{1,2\},\{1,3\},\{1,4\},\{2,3\},\{2,4\},\{3,4\}\}.\]
Note $S_1=\{\{1,2\},\{2,3\},\{3,4\},\{1,4\}\}$ and $S_2=\{\{1,3\},\{2,4\}\}$, then
\begin{align}\mathbb E[T(d_x,L)]+\mathbb E[T(d_y,L)]=&\mathbb P(L\cap D\in S_1)\times 1\\
&+\mathbb P(L\cap D\in S_2)\times 2\\
=&\frac43,\end{align}
\begin{align}\mathbb E[T(L,L')]=&\sum_{s,s'\in S}\mathbb P(L\cap D\subset s,L'\cap D\subset s')E[T(L,L')|L\cap D\subset s,L'\cap D\subset s']\\
=&\frac1{36}\left(\sum_{s=s'}+\sum_{|s\cap s'|=1}+\sum_{|s\cap s'|=0}\right)E[T(L,L')|L\cap D\subset s,L'\cap D\subset s']\\
=&\frac1{36}\left(\sum_{s=s'}\frac12+\sum_{|s\cap s'|=1}\frac12+\sum_{|s\cap s'|=0,s\in S_1}0+\sum_{|s\cap s'|=0,s\in S_2}1\right) \\
=&\frac1{36}(6\times\frac12+24\times\frac12+4\times 0+2\times 1)\\
=&\frac{17}{36}.\end{align}
So \begin{align}\mathbb E[X_n]=&\mathbb E[X_0]+\sum_{i=1}^n\mathbb E[T_{n}]+n\\
=&4+\frac43n+\frac{89}{72}n(n-1)+n\\
=&\frac{17}{72}n^2+\frac{79}{72}n+4.\end{align}
When $n=25$, th expectation is $204$.
\end{humanbox}

\subsection{AIME\_1\_14}
\begin{questionbox}
\textit{Let $ABCDE$ be a convex pentagon with $AB=14,$ $BC=7,$ $CD=24,$ $DE=13,$ $EA=26,$ and $\angle B=\angle E=60^{\circ}.$ For each point $X$ in the plane, define $f(X)=AX+BX+CX+DX+EX.$ The least possible value of $f(X)$ can be expressed as $m+n\sqrt{p},$ where $m$ and $n$ are positive integers and $p$ is not divisible by the square of any prime. Find $m+n+p.$}
\end{questionbox}

\begin{R1box}
\textbf{\Large R1 Reasoning}

\vspace{1\baselineskip}

In \textbf{Steps 1–28}, in $\triangle ABC$, given $AB = 14$, $BC = 7$, and $\angle B = 60^{\circ}$, using the Law of Cosines, R1 calculates using the Law of Cosines: $AC=7\sqrt{3}$. Similarly, R1 computes $AD = 13\sqrt{3}$.

In \textbf{Steps 29–40}, R1 considers three approaches: (1) geometric means with symmetry, (2) special points including but not limited to the Torricelli point, and (3) vector-based computation upon establishing a coordinate system. Finally, R1 proceeds with Method 3: establishing a coordinate system with $B$ designated as the origin $(0,0)$, setting point $C$ at $(7,0)$, where the Pythagorean theorem confirms $\angle ACB$ is $90^{\circ}$, thus resulting in coordinates $(7, 7\sqrt{3})$ for point $A$.

In \textbf{Steps 41–44}, R1 likewise deduces from the Pythagorean theorem that $\triangle ADE$ is right-angled with $\angle D = 90^{\circ}$ and $\angle E = 60^{\circ}$.

In \textbf{Steps 45–48}, R1 considers $\angle D=90^{\circ}$ and $\angle E=60^{\circ}$ to be contradictory (although actually correct). Through numerical verification and logical inspection, R1 finds the misjudgment stems from confusing the notation.

In \textbf{Steps 49–64}, R1 finds the coordinate calculation for points $D$ and $E$ challenging and proposes establishing a new coordinate system with $D$ as the origin. However, since coordinates of points $A$ and $C$ prove equally difficult to determine, R1 thus returns to the coordinate system with $B$ as the origin.

In \textbf{Steps 65–84}, R1 assumes $\boldsymbol{DA} = (7 - dx, 7\sqrt{3} - dy)$ and $\boldsymbol{DE} = (ex - dx, ey - dy)$, obtaining the following system of equations:

\[ 
\left\{
\begin{aligned}
     & (7-dx)(ex-dx)+(7\sqrt{3}-dy)(ey-dy)= 0 \quad (\angle ADE=90^{\circ})\\
     & (ex-dx)^2+(ey-dy)^2= 169 \quad (DE=13)\\
     & (7-ex)^2+(7\sqrt{3}-ey)^2= 676 \quad (AE=26)\\
     & (7-dx)^2+(7\sqrt{3}-dy)^2= 507 \quad (AD=13\sqrt{3})\\
     & (dx-7)^2+(dy)^2= 576 \quad (CD=24)\\
\end{aligned}
\right.
\]

In \textbf{Steps 85–120}, R1 simplifies the system by subtracting Equation 5 from Equation 4, obtaining: $dy=36\sqrt{13}/7$. Substituting this result into Equation 5 yields:

\[
(dx-7)^2=\frac{24336}{49}=496+\frac{32}{49}
\]

Then, R1 verifies the correctness of the numerical computation in the above solving process.

\end{R1box}
\begin{R1box}

In \textbf{Steps 130–142}, R1 considers that the numerical expression on the right-hand side of the equation is unsuitable for direct square root computation and thus recalculates it for verification, confirming the correctness of the right-hand term. Subsequently, R1 performs the square root operation to obtain the result:

\[
dx=\frac{205}{7},-\frac{107}{7}\Rightarrow D=(\frac{205}{7},\frac{36\sqrt{13}}{7}),(-\frac{107}{7},\frac{36\sqrt{13}}{7})
\]

In \textbf{Steps 143–156}, given that pentagon $ABCED$ is convex, R1 hypothesizes that the coordinates of point $D$ should lie to the right of point $C$. Nevertheless, further categorical analysis remains necessary.

In \textbf{Steps 157–185}, R1 first considers point $D$ with coordinates $(205/7, 36\sqrt{3}/7)$. Given that $AD$ is perpendicular to $DE$, and the known vector ‌$\boldsymbol{DA‌} = (-156/7, 13\sqrt{3}/7)$, thus R1 obtains two possible directional vectors for $\boldsymbol{DE}$:

\[
\boldsymbol{a}=(\frac{13\sqrt{3}}{7}, \frac{156}{7}),(-\frac{13\sqrt{3}}{7}, -\frac{156}{7})
\]

R1 takes $\boldsymbol{a}=( 13\sqrt{3}/7,156/7,)$, since the length of DE is 13, it follows that:
\begin{align*}
    E&=D+|DE|\cdot\frac{\boldsymbol{a}}{|\boldsymbol{a}|}\\
     &=(\frac{205}{7},\frac{36\sqrt{13}}{7})+13\cdot\frac{(\frac{13\sqrt{3}}{7}, \frac{156}{7})}{13\sqrt{3}}\\
     &=(\frac{205}{7},\frac{36\sqrt{13}}{7})+13\cdot(\frac{\sqrt{3}}{7},\frac{12}{7})\quad\text{(note: R1 experiences a calculation error here.) }\\
     &=(\frac{205+13\sqrt{3}}{7},\frac{36\sqrt{13}+156}{7})
\end{align*}
In \textbf{Steps 186–206}, R1 initiates verification of whether the vector $\boldsymbol{a}$ takes $( 13\sqrt{3}/7,156/7)$ correctly (note: this check does not pertain to validating the numerical results mentioned above). The coordinates of point $E$ have been determined as $(\frac{205+13\sqrt{3}}{7},\frac{36\sqrt{13}+156}{7})$. Given that point $A$ is at $(7,7\sqrt{3})$, the distance $AE$ is calculated as $\sqrt{1014}$. This contradicts the known condition $AE=26$, prompting R1 to reject the vector $\boldsymbol{a}$ taking $( 13\sqrt{3}/7,156/7,)$. The discussion now proceeds to the case where vector $\boldsymbol{a}$ takes $(-13\sqrt{3}/7,-156/7)$.

In \textbf{Steps 207–226}, ‌R1 observes:
\begin{align*}
    E &= D -13\cdot( \frac{\sqrt{3}}{7},\frac{12}{7} )\\
      &= (\frac{205-13\sqrt{3}}{7},\frac{36\sqrt{13}-156}{7})
\end{align*}
Similarly to steps 186-206, R1 computes that the length of $AE$ is $\sqrt{1014}$, which is not equal to $26$, resulting in a contradiction; thus, it is inferred that the case with vector $\boldsymbol{a}=(-13\sqrt{3}/7,-156/7)$ is also incorrect. 

\end{R1box}
\begin{R1box}

In \textbf{Steps 227–237}, R1 concludes that establishing the coordinate system with point $B$ as the origin was flawed and began evaluating alternative approaches:
‌
\begin{enumerate}
    \item Re-establishing coordinates with D as origin, but recognizes this was conceptually identical to the B-origin approach and thus rejects it.

    \item  Decomposing the pentagon into triangles‌ and leveraging Torricelli points with segment transformations on $f(X)$, but dismissed this due to excessive technical complexity and uncertain feasibility.
    
    \item Direct vector representation of $f(X)$‌, which was deemed equivalent to coordinate-based methods in essence.

    \item ‌Testing for concyclicity‌ (5-point circle) to use the center point, but R1 considers this statistically improbable and impractical.
    
\end{enumerate}

In \textbf{Steps 238–278}, R1 revisits the coordinate system setup approach with point $B$ as the origin and ultimately identified the computational error noted in the prior annotation, successfully deriving the correct coordinates for point $E$:

\[
E=(\frac{205}{7},\frac{36\sqrt{13}}{7})+(\frac{13}{7},\frac{52\sqrt{13}}{7})=(\frac{218}{7},\frac{88\sqrt{13}}{7})
\]
Subsequently, R1 computationally verifies that the equation $|AE| = 26$ and $\angle E=60^{\circ}$ holds. In summary, the coordinates of the five vertices obtained are:
\[
A(7,7\sqrt{3}),B(0,0),C(7,0),D(\frac{205}{7},\frac{36\sqrt{13}}{7}),E(\frac{218}{7},\frac{88\sqrt{13}}{7})
\]
(Note: At this stage, R1 omits discussing two possibilities: (1) Point $D$ being positioned on the left-hand side of Point $C$; (2) Vector a adopting a reversed direction.)

In \textbf{Steps 279–316}, R1 begins by calculating the minimum value of f(‌X‌); it then verifies whether this minimum holds by evaluating the function at the coordinates of ‌five vertices‌. After tedious computations, R1 concludes that f(‌A‌) is the smallest among these five values, ‌yet not necessarily the global minimum‌ of the problem.

In \textbf{Steps 317–328}, R1 hopes to find a breakthrough point by studying the particularity of $\triangle ACD$; it attempts to prove that $\triangle ACD$ is either a right triangle or contains a $60^{\circ}$ angle, but rigorous calculations demonstrate that neither conjecture holds.

In \textbf{Steps 329–355}, R1 starts considering potential methods for solving $X$:

\begin{enumerate}
    \item The equilibrium point of pentagon $ABCDE$.
    \item Constructing an equilateral triangle using angles $B$ and $E$.   
    \item The Torricelli point of $\triangle ABC$ and $\triangle ADE$.
    \item ‌The sum of specific distances.    
\end{enumerate}

R1 considers combining Method 3 and Method 4 for the solution and lists out the following possibilities:

\begin{enumerate}
    \item $AC + AD + BE$.
    \item $AC + CD + DE + BE$.   
    \item Distances from the Torricelli point (Fermat point) of $\triangle ABC$ or $\triangle ADE$ to all five vertices.   
\end{enumerate}

\end{R1box}
\begin{R1box}

‌R1 feels that the previous guesses only yield function values and fail to pinpoint the minimum point. R1 hopes to identify special intersection points by constructing equilateral triangles, as these might represent the minimum points. However, the coordinates of points $A$, $D$, and $E$ currently appear exceedingly complex, so R1 wishes to adopt methods to simplify the numerical values.‌

In \textbf{Steps 356–372}, R1 postulates that the Cartesian coordinates of points $A$, $D$, and $E$—all containing the radical term $\sqrt{3}$—might be computationally streamlined by converting to a complex coordinate system.‌ ‌R1 meticulously calculates the complex coordinates of each point, but recognizes that this approach was overcomplicating the problem, prompting consideration of alternative approaches.‌

In \textbf{Steps 373–395}, ‌Applying the Law of Cosines, R1 computes $\angle C$ as approximately $68^{\circ}$. Subsequently, via the Pythagorean theorem, R1 confirms that $\triangle ACD$ is not a right triangle.

In \textbf{Steps 396–405}, R1 hopes to find the shortest path using the reflection principle. It attempts to reflect point $A$ onto side $BC$ or onto side $DE$, but discovers limited significance in either approach.

In \textbf{Steps 406–421}, R1 refocuses on the coordinate system with point $B$ as the origin and proposes the following conjectures:

\begin{enumerate}
    \item Calculate the Torricelli points $T_1$ and $T_2$ for $\triangle ABC$ and $\triangle ADE$ respectively. Connecting these points, it conjectures that the minimum point may lie on segment $T_1T_2$ – possibly even coinciding with segment $CD$, suggesting the minimum might reside on $CD$ itself.
    \item Given the sufficient length of boundary $CD$ and the convexity of pentagon $ABCDE$, the minimum point likely lies directly on boundary $CD$.   
    \item Conceptualizing the problem as a weighted graph, it considers finding a Steiner point connecting all five vertices, though recognizes Steiner trees for five points are highly complex.   
\end{enumerate}

In \textbf{Steps 422–427}, R1 speculates that the minimum point $X$ may lie at the intersections of special segments. It suddenly embarks on verifying whether points $A$, $C$, $D$ are collinear (Note: This violates the pentagon's convex constraint) and whether $B$, $X$, $E$ are collinear (Note: This remains unsupported by current verification, though serendipitously, $X$ ultimately lies on segment $BE$ in the human-provided solution).

In \textbf{Steps 428-456}, ‌R1 suspects the minimization point may reside near vertices $A$ or $C‌$ given the relatively small values of $f(A)$ and $f(C)$, further conjecturing it might coincide with the Torricelli point of $\triangle ABC$. Consequently, R1 proceeds to compute the Torricelli point for $\triangle ABC$.

R1 constructs an equilateral triangle for the solution methodology.‌ Using segment $BC$ as the base and oriented downward to avoid intersecting the pentagon, it obtains point $B'(7/2, 7\sqrt{3}/2)$, forming equilateral triangle $BB'C$.

Subsequently, R1 employs the rotation matrix to rotate segment $BC$ $60^{\circ}$ clockwise about point $B$.‌ Upon validating the coordinates of $B'$ using this transformation, R1 confirms their correctness per the computational verification.

R1 proceeds to construct an equilateral $\triangle AA'B$ using the side $AB$, oriented outward from the pentagon.‌ Although the Torricelli point should emerge at the intersection of segments $A'C$ and $AB'$, R1 ultimately abandons the calculation due to the prohibitively complex nature of this procedure.

R1 considers exploiting the unique geometric properties of $\triangle ABC$ to accelerate the calculation of its Torricelli point,‌ deviating from conventional construction methods.

\end{R1box}
\begin{R1box}
In \textbf{Steps 457-474}, the application of the Sine Theorem confirms that $\triangle ABC$ is a $30^{\circ}-60^{\circ}-90^{\circ}$ triangle with side ratios $1:\sqrt{3}:2$. However, computation of the Torricelli point still necessitates the standard equilateral construction approach, reverting to the initial methodology. R1 finds this process prohibitively time-consuming and computationally intensive, consequently abandoning this line of investigation.

In \textbf{Steps 475-526}, R1 proposes the following conjectures:

\begin{enumerate}
    \item The minimum value may be equal to $AC+AD+BE=20\sqrt{3}+38$, although the spatial location of the minimum point $X$ remains indeterminate.
    \item Could the Torricelli points of $\triangle ABC$ and $\triangle ADE$ coincide? If coincident, it might correspond to the minimum point, but their noncoincidence is immediately evident, thus discarded this conjecture.
    \item $X$ may coincide with the intersection of segments $AC$ and $BE$ in the geometric configuration.  
\end{enumerate}

R1 meticulously calculates the intersection of the segments $AC$ and $BE$, tentatively identifying it as the minimum point $X$ with the coordinates $(7,308\sqrt{3}/109)$. Subsequently, R1 computes $f(X)\approx 72.66$, while the expected minimum value is $AC+AD+BE\approx72.46$.  Integrating Conjecture 1 and Conjecture 3, R1 derived the following conclusion:‌

\[
X=(7,\frac{308\sqrt{3}}{109}),f(X)=AX+BX+CX+DX+EX=20\sqrt{3}+38
\]
So $m=38$, $n=20$, $p=3$, giving $m+n+p=\boxed{61}$.

\end{R1box}

\begin{wrongbox}
\textbf{\Large Failure Reasoning}

\vspace{1\baselineskip}

The failure in R1's reasoning process can be attributed to one main issue: Not forming a problem-solving approach. The primary reason for the failure of R1 to solve the problem lies in the absence of a solid strategy. R1 devotes substantial discussion to the elaboration of various conjectures, yet none is viable. All focus remains on analyzing specific points and values rather than pursuing systematic geometric deduction.

\end{wrongbox}
\begin{humanbox}
\textbf{\Large Human Reasoning}

\vspace{1\baselineskip}

\begin{figure}[H]
  \centering
  \includegraphics[width=0.8\textwidth]{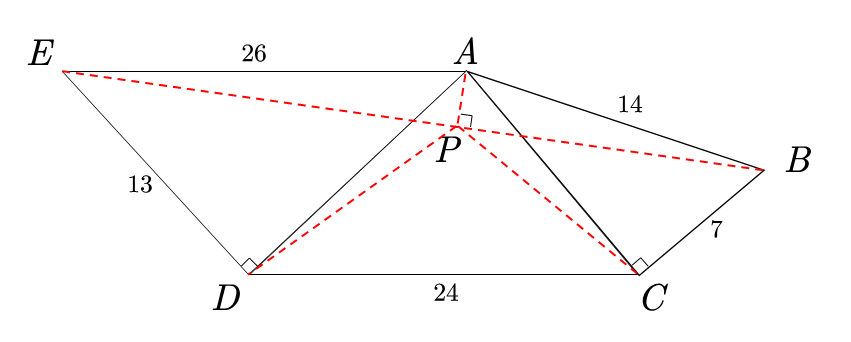} 
  \caption{Sketch map}
  \label{fig:geometric figure}
\end{figure}

Given triangle $ABC$ with: $AB = 14,BC = 7,\angle ABC = 60^\circ$.We will prove that $\triangle ABC$ is a right triangle.

Using the law of cosines: 
\[
AC^2 = AB^2 + BC^2 - 2 \cdot AB \cdot BC \cdot \cos(\angle ABC)=147
\]
So AC = $\sqrt{147} = 7\sqrt{3}$. Then we check the relationship between sides:
\[
BC^2 + AC^2 = 49 + 147 = 196 = AB^2
\]
By the converse of the Pythagorean theorem, $\angle ACB = 90^\circ$. Therefore, $\triangle ABC$ is a right triangle with a right angle at $C$. Similarly, we see that $\triangle ADE$ is a right triangle with a right angle at $D$ and $AD=13\sqrt{3}$.

Leaving $P$ as the foot of the altitude from $A$ to $BE$, we try to prove that the point $P$ is the Fermat point of $\triangle ACD$. Note that:
\[
\angle ADE=\angle APE=90^{\circ}, \angle ACB=\angle APB=90^{\circ}
\]
So $APDE$ and $APCB$ are cyclic. Then we have
\begin{align*}
        \angle APD=180^{\circ}-\angle AED=180^{\circ}-60^{\circ}=120^{\circ}\\
    \angle APC=180^{\circ}-\angle ABC=180^{\circ}-60^{\circ}=120^{\circ}
\end{align*}
So $P$ is a Fermat point of $\triangle ACD$. Now for all $X$,
\begin{align*}
    f(X)&=(AX+CX+DX)+(BX+EX)\\
    &\geq (AP+CP+DP)+(BP+EP)=f(P)
\end{align*}

\end{humanbox}
\begin{humanbox}
So $f(X)$ is minimized at $P$, we just need to evaluate $f(P)$ now. By the law of cosines:
\[
    CD^2=AD^2+AC^2-2AD\cdot AC\cdot\cos(\angle CAD) \Rightarrow  \cos(\angle CAD)=\frac{1}{7}
\]

Now $\sin(\angle CAD)=4\sqrt{3}/7$, so
\begin{align*}
\cos(\angle BAE)&=\cos(60^{\circ}+\angle CAD)\\
&=\cos(60^{\circ})\cos(\angle CAD)-\sin(60^{\circ})\sin(\angle CAD)=-\frac{11}{14}
\end{align*}
As such, by the law of cosines:
\[
BE^2 = AE^2 + AB^2 - 2 \cdot AE \cdot AB \cdot \cos(\angle BAE)=1444
\]
So $BP+PE=BE=38$. Let $AP=a,CP=b,DP=c$, by the law of cosine:
\[
\left\{
\begin{aligned}
    a^2+ab+b^2 & =  AC^2=147 \\
    a^2+ac+c^2 & =  AD^2=507 \\
    b^2+bc+c^2 & =  CD^2=576 \\
\end{aligned}
\right.
\]
So
\begin{align*}
    & (a^2+ab+b^2)+(a^2+ac+c^2)+(b^2+bc+c^2)\\
   =& 2(a^2+b^2+c^2)+(ab+bc+ca)=1230
\end{align*}
Also,
\begin{align*}
    S_{\triangle ACD}=\frac{1}{2}\cdot AD\cdot AC\cdot \sin(\angle CAD)=78\sqrt{3}
\end{align*}
Then,
\begin{align*}
    S_{\triangle ACD}&=S_{\triangle APC}+S_{\triangle CPD}+S_{\triangle APD}\\
    &=\frac{1}{2}ab\frac{\sqrt{3}}{2}+\frac{1}{2}bc\frac{\sqrt{3}}{2}+\frac{1}{2}ca\frac{\sqrt{3}}{2}\\
    &=\frac{\sqrt{3}}{4}(ab+bc+ca)=78\sqrt{3}
\end{align*}
So $ab+bc+ca=312$. Now,
\begin{align*}
    2(a^2+b^2+c^2)&=2(a^2+b^2+c^2+2ab+2bc+2ca)\\
    &=2(a^2+b^2+c^2)+(ab+bc+ca)+3(ab+bc+ca)\\
    &=1230+3\times 312=2166
\end{align*}
So $a+b+c=19\sqrt{3}$ and $f(P)=38+19\sqrt{3}$. As requested in the problem, we have obtained the final answer:
\[
m=38,n=19,p=3,m+n+p=\boxed{60}.
\]
\end{humanbox}

\subsection{AIME\_1\_15}
\begin{questionbox}
\textit{Let $N$ denote the numbers of ordered triples of positive integers $(a, b, c)$ such that $a, b, c \le 3^6$ and $a^3 + b^3 + c^3$ is a multiple of $3^7$. Find the remainder when $N$ is divided by $1000$.}
\end{questionbox}
\begin{R1box}
\textbf{\Large R1 Reasoning}\vspace{1\baselineskip}

In \textbf{Steps 1–29}, the model correctly identifies the problem and explores several advanced potential solution paths, including Hensel's lemma, generating functions, and character sums. It proposes an initial hypothesis that if the sums of cubes are uniformly distributed modulo $3^7$, the number of solutions would be $(3^6)^3 / 3^7 = 3^{11}$. However, it correctly notes that this is a simplistic assumption that needs verification.

In \textbf{Steps 30–52}, the model attempts to analyze the structure of $a^3 \pmod{3^7}$. Using a base-3 expansion, it correctly argues that for $a \in \{1, \dots, 3^6\}$, the value of $a^3 \pmod{3^7}$ depends only on the value of $a \pmod{3^6}$. Since the range of $a$ covers each residue modulo $3^6$ exactly once, it correctly reformulates the problem as counting the number of triples $(x, y, z) \in (\mathbb{Z}/3^6\mathbb{Z})^3$ such that $x^3+y^3+z^3 \equiv 0 \pmod{3^7}$.

In \textbf{Steps 53–135}, the model attempts to solve the problem by inductively counting solutions modulo $3^k$ (a form of Hensel's lifting). It correctly analyzes the base cases:
\begin{itemize}
    \item Modulo 3, the condition is $x+y+z \equiv 0 \pmod 3$, which has $3^2=9$ solutions in $(\mathbb{Z}/3\mathbb{Z})^3$.
    \item Modulo 9, it correctly determines that $x^3 \pmod 9$ depends only on $x \pmod 3$. It meticulously counts the solutions by considering which residue combinations mod 3 can have a sum of cubes equal to 0 mod 9. It correctly calculates the number of solutions in $(\mathbb{Z}/9\mathbb{Z})^3$ to be $3^3 + 6 \times 3^3 = 27 + 162 = 189$.
\end{itemize}

In \textbf{Steps 136–144}, the model observes that the number of solutions grew from 9 (for mod 3) to 189 (for mod 9), a factor of 21. It recognizes this does not match a simple scaling rule (e.g., a factor of $3^2=9$) and, unable to determine the correct, more complex lifting pattern for higher powers of 3, it abandons this valid but difficult line of reasoning.

In \textbf{Steps 145–203}, after briefly revisiting and again abandoning the character sum approach, the model reverts to its initial, simplistic hypothesis from Step 11. Despite having explicitly disproven the uniform distribution assumption with its own correct calculation for the modulo 9 case (since $189 \neq 9^3/9 = 81$), the model abandons its rigorous work. It incorrectly concludes that the answer must come from the uniform distribution assumption, yielding $3^{11} = 177,147$ solutions, and thus a final answer of $147 \pmod{1000}$. The final conclusion is based on a hypothesis the model itself proved to be false.

\end{R1box}
\begin{wrongbox}
\textbf{\Large Failure Reasoning}\vspace{1\baselineskip}

The model does not find effective method in first few steps. It finds some valuable approaches like LTE(Lifting The Exponent Lemma) But it can not make further inference. The model tried with simplified cases to find patterns but it seems that there is no pattern to follow.
\end{wrongbox}

\begin{humanbox}
    \textbf{\Large Human Reasoning}\vspace{1\baselineskip}

    \textbf{Claim.} Fix $n \geq 2$. The image of $x^3$ for $x \in (\mathbb{Z}/3^n\mathbb{Z})^\times$ covers every residue which is $\pm1 \pmod{9}$ exactly three times.\\
\textbf{Proof:} Follows immediately as $(\mathbb{Z}/3^n\mathbb{Z})^\times$ has a primitive root and $3 \mid 2 \cdot 3^{n-1}$.

\vspace{1em}

\textbf{Claim.} In fact the number of solutions is just the solutions for $a, b, c$ less than $3^7$, divided by $27$.\\
\textbf{Proof:} Observe $(x + 3^6)^3 \equiv x^3 \pmod{3^7}$.

Thus we solve for $0 \leq x, y, z < 3^7$. Since each residue is $\pm1 \pmod{9}$, we can either have all divisible by 3, or divisible by 3, $-1 \mod 9$, and $1 \mod 9$.

For the latter case, we can pick one of 3 to be divisible by 3, and have $3^6$ choices there. Then each $y$ coprime to 3, of which there are $2 \cdot 3^6$, gives rise to another coprime residue, which has exactly 3 cases. Thus here we obtain
\[
3 \cdot 3^6 \cdot 3^6 \cdot 3 = 2 \cdot 3^{14}.
\]

If all are divisible by 3, we solve mod $3^4$ for $x, y, z \leq 3^6$. The same arguments above give $(3^2)^3 \cdot (3 \cdot 3^3 \cdot 2 \cdot 3^3 \cdot 3)$ if not all are divisible by 3.

Lastly we solve mod 3 for $x, y, z \leq 3^5$. There are 9 cases normally, but we can take one of $3^4$ options for each; so we get $3^{14}$.

Our total now is $5 \cdot 3^{14}$, dividing by 27 gives $5 \cdot 3^{11}$, and this is readily computed to be
\[
735 \pmod{1000}.
\]

\end{humanbox}

\newpage
\subsection{AIME\_1\_16}

\begin{questionbox}
\textit{Six points $A, B, C, D, E,$ and $F$ lie in a straight line in that order. Suppose that $G$ is a point not on the line and that $AC=26, BD=22, CE=31, DF=33, AF=73, CG=40,$ and $DG=30.$ Find the area of $\triangle BGE.$}
\end{questionbox}
\begin{R1box}
\textbf{\Large R1 Reasoning}\vspace{1\baselineskip}

In \textbf{Steps 1–26}, the model correctly interprets the problem by establishing a 1D coordinate system, placing the line containing the points A, B, C, D, E, F on the x-axis with A at the origin. It then systematically and accurately uses the given distances (AF=73, DF=33, BD=22, AC=26, CE=31) to determine the coordinates of all six points: A(0), B(18), C(26), D(40), E(57), and F(73). The model verifies that the calculated coordinates are consistent with all given constraints.

In \textbf{Steps 27–58}, the model proceeds to find the coordinates of point G. It correctly uses the distances CG=40 and DG=30 to set up a system of two equations for the coordinates (x,y) of G:
\begin{align*} (x - 26)^2 + y^2 &= 40^2 \\ (x - 40)^2 + y^2 &= 30^2 \end{align*}
By subtracting the equations to eliminate $y^2$, it correctly solves for $x=58$. Substituting this value back, it finds $y^2 = 576$, yielding $y = \pm 24$. The model correctly notes that the sign of y does not affect the area and proceeds with G=(58, 24).

In \textbf{Steps 59–103}, the model calculates the area of the target triangle BGE using the vertices B(18, 0), G(58, 24), and E(57, 0). It first uses the shoelace formula and correctly computes the area to be 468. To verify this result, it uses a second, simpler method. It correctly identifies the segment BE on the x-axis as the base of the triangle (length $57-18=39$) and the y-coordinate of G as the height (24). The calculation $\frac{1}{2} \times 39 \times 24 = 468$ confirms the result, demonstrating a robust and accurate solution process.
\end{R1box}

\begin{humanbox}
\textbf{\Large Human Reasoning}\vspace{1\baselineskip}

Let \( AB = x \). Then, we know \( BC = 26 - x \), \( CD = x - 4 \), \( DE = 35 - x \), \( EF = x - 2 \).
$
AB + BC + CD + DE + EF + AF = x + (26 - x) + (x - 4) + (35 - x) + (x - 2) + x = 55.
$

Therefore, \( x = 18 \), so
\[
CD = x - 4 = 18 - 4 = 14
\]

Using Heron's formula to find the area of \( \triangle CDG \), with sides \( CD = 14 \), \( DG \), and \( CG \), where the semi-perimeter \( s = \frac{CD + DG + CG}{2} \). The area is
\[
\text{Area} = \sqrt{s (s - CD) (s - DG) (s - CG)}
\]
Given the area of \( \triangle CDG = 168 \), \( \triangle BGE \) has the same height as \( \triangle CDG \), so the area of \( \triangle BGE = \frac{168}{14} \times 39 = 468 \).
    \end{humanbox}

\subsection{AIME\_1\_17}
\begin{questionbox}
\textit{Find the sum of all positive integers $n$ such that $n+2$ divides the product $3(n+3)(n^2+9)$.}
\end{questionbox}

\begin{R1box}
\textbf{\Large R1 Reasoning}\vspace{1\baselineskip}

In \textbf{Steps 1–13}, the model correctly interprets the problem as a divisibility condition. It chooses an efficient and appropriate method, using modular arithmetic to state the problem as \( 3(n + 3)(n^2 + 9) \equiv 0 \pmod{n + 2} \). The key insight is to evaluate the expression modulo \( n+2 \) by substituting \( n \equiv -2 \). The model correctly computes \( 3((-2) + 3)((-2)^2 + 9) = 3 \times 1 \times 13 = 39 \). This leads to the correct conclusion that \( n + 2 \) must be a divisor of 39.

In \textbf{Steps 14–17}, the model correctly identifies the positive divisors of 39 as 1, 3, 13, and 39. It then solves for \( n \) in each case: \( n+2=1 \implies n=-1 \), \( n+2=3 \implies n=1 \), \( n+2=13 \implies n=11 \), and \( n+2=39 \implies n=37 \). It correctly discards \( n=-1 \) because the problem asks for positive integers. The remaining values are summed to get the correct answer, \( 1 + 11 + 37 = 49 \).

In \textbf{Steps 18–69}, the model performs two robust verifications of its result.

First, in \textbf{Steps 18–41}, it directly substitutes each of the found values of \( n \) (1, 11, 37) back into the original expression and confirms that the divisibility condition holds. 

Second, in \textbf{Steps 45–63}, it uses an alternative method, polynomial long division, to divide \( 3(n + 3)(n^2 + 9) \) by \( n + 2 \). It correctly finds a remainder of 39, which formally proves that the divisibility condition is equivalent to \( n+2 \) dividing 39. This second method confirms the validity of the initial modular arithmetic approach.
\end{R1box}

\begin{humanbox}
    \textbf{\Large Human Reasoning}\vspace{1\baselineskip}

     First note \( n + 2 \mid n + 3 \) always so it's just asking \( n + 2 \mid (3n^2 + 9) \).

\textbf{Case 1:} \( n \equiv 3k - 2 \)
Then \( k \mid (3k - 2)^2 + 9 \) so \( k \mid 13 \). Then \( k = 1, 13 \) and \( n = 1, 37 \).

\textbf{Case 2:} \( 3 \mid n \)
Then \( n + 2 \mid n^2 + 9 \), so \( n + 2 \mid -2n + 9 \), so \( n + 2 \mid 13 \). So \( n = 11 \).

In total they are \( 1, 11, 37 \) which sum to 49.
    \end{humanbox}

\newpage
\subsection{AIME\_1\_18}
\begin{questionbox}
\textit{Four unit squares form a $2 \times 2$ grid. Each of the $12$ unit line segments forming the sides of the squares is colored either red or blue in such way that each square has $2$ red sides and blue sides. One example is shown below (red is solid, blue is dashed). Find the number of such colorings.}
\end{questionbox}
\begin{R1box}

\textbf{\Large R1 Reasoning}\vspace{1\baselineskip}

In \textbf{Steps 1–6}, the model correctly deconstructs the 2x2 grid, identifying that it consists of 12 total edges. It accurately classifies these into 8 boundary edges (each belonging to a single square) and 4 internal edges (each shared between two squares).

In \textbf{Steps 7–143}, the model develops a sound and systematic strategy. It correctly formulates the problem as a set of constraints: for each of the four squares, the sum of its four edge colors (represented as 0 for blue, 1 for red) must be 2. The key insight is to partition the problem based on the coloring of the 4 internal edges. The model correctly deduces that for any of the $2^4=16$ possible colorings of the internal edges, the number of required red edges for the boundary part of each square is determined. Let $N(s)$ be the number of ways to color two boundary edges to have a sum of $s$ red edges, where $N(0)=1$, $N(1)=2$, and $N(2)=1$. The total number of colorings is the sum over all 16 internal colorings of the product of the corresponding $N(s_i)$ for each of the four squares.

In \textbf{Steps 144–344}, the model meticulously executes this strategy. It performs a case-by-case analysis of all 16 possible colorings for the 4 internal edges. For each case, it correctly calculates the number of valid ways to color the 8 boundary edges. For example:
\begin{itemize}
    \item If all 4 internal edges are blue, each square requires 2 red boundary edges. This can be achieved in $N(2) \times N(2) \times N(2) \times N(2) = 1 \times 1 \times 1 \times 1 = 1$ way.
    \item If two internal edges forming a diagonal are red and the other two are blue (e.g., case 7), then each square has one red and one blue internal edge. Thus, each square requires 1 red boundary edge. This can be achieved in $N(1) \times N(1) \times N(1) \times N(1) = 2 \times 2 \times 2 \times 2 = 16$ ways.
\end{itemize}
The model correctly calculates the contributions from all 16 cases and sums them to get the total number of colorings: $1 \times 2 + 12 \times 4 + 2 \times 16 = 2 + 48 + 32 = 82$.

In \textbf{Steps 345–498}, the model performs verification. It correctly applies its method to a simpler 1x2 grid and obtains the correct answer of 18, which increases confidence in its approach. It briefly explores a dual graph approach but becomes confused and abandons it. It concludes, based on its detailed and verified case analysis, that the answer is 82.

\end{R1box}

\begin{humanbox}
      \textbf{\Large Human Reasoning}\vspace{1\baselineskip}

    \textbf{Case 1: 4 red lines strictly inside the square} \\
There is only 1 structure. The 4 red lines are in the center.

\textbf{Case 2: 3 red lines strictly inside the square} \\
There are 4 choices to choose the segment strictly within the square that is not a red line. From that, there are 2 choices to place another red segment for each square containing only 1 internal red segment, so we have
\[
4 \cdot 2 \cdot 2 = 16 \text{ cases.}
\]

\textbf{Case 3: 2 red lines strictly inside the square, the lines are parallel} \\
2 ways the red lines can be formed inside the square.

There are 2 choices to place another red segment for each square containing only 1 internal red segment, similar to above, so we have
\[
2 \cdot 2 \cdot 2 \cdot 2 \cdot 2 = 32 \text{ cases.}
\]

\textbf{Case 4: 2 red lines strictly inside the square, the lines are perpendicular} \\
There are 4 ways to orient these 2 lines.

For the square with no internal red segments, the other sides of that square must be red. For the squares with 1 internal red segment, there are 2 ways to place it. Thus we have
\[
4 \cdot 2 \cdot 2 = 16 \text{ cases.}
\]

\textbf{Case 5: 1 internal red segment} \\
4 ways to orient this.

Similar to the reasons above, we have the squares with no internal red segments, the other two sides must be red. Thus, we have
\[
4 \cdot 2 \cdot 2 = 16 \text{ cases.}
\]

\textbf{Case 6: 0 internal} \\
1 case for this.

\vspace{1em}
Thus, we have our answer as
\[
1 + 16 + 32 + 16 + 16 + 1 = 82
\]
\end{humanbox}

\newpage
\subsection{AIME\_1\_19}
\begin{questionbox}
\textit{The product\[\prod^{63}_{k=4} \frac{\log_k (5^{k^2 - 1})}{\log_{k + 1} (5^{k^2 - 4})} = \frac{\log_4 (5^{15})}{\log_5 (5^{12})} \cdot \frac{\log_5 (5^{24})}{\log_6 (5^{21})}\cdot \frac{\log_6 (5^{35})}{\log_7 (5^{32})} \cdots \frac{\log_{63} (5^{3968})}{\log_{64} (5^{3965})}\]is equal to $\tfrac mn,$ where $m$ and $n$ are relatively prime positive integers. Find $m + n.$}
\end{questionbox}

\begin{R1box}
\textbf{\Large R1 Reasoning}

\vspace{1\baselineskip}

In \textbf{Steps 1–9}, R1 begins by analyzing the given product and identifies the need to simplify each term. It breaks down the logarithmic expressions and rewrites them using logarithm properties, such as $\log_a (b^c) = c \log_a b$. This sets the foundation for further simplification.

In \textbf{Steps 10–12}, R1 separates the product into two distinct parts: one involving polynomial fractions and the other involving logarithmic ratios. This separation is crucial for simplifying the problem into manageable components.

From \textbf{Steps 13–272}, R1 focuses on simplifying the first product, $\prod_{k=4}^{63} \frac{k^2 - 1}{k^2 - 4}$. It factors the numerator and denominator as differences of squares, rewrites the terms, and meticulously tracks the cancellation of factors across the product. After confusion and multiple steps, R1 eventually arrives at the simplified form $\frac{5 \times 62 \times 64}{2 \times 61 \times 65} = \frac{1984}{793}$.

In \textbf{Steps 273–285}, R1 simplifies the second product, $\prod_{k=4}^{63} \frac{\ln(k+1)}{\ln k}$. Recognizing the telescoping pattern, it quickly reduces the product to $\frac{\ln 64}{\ln 4}$.

In \textbf{Steps 286–307}, R1 combines the results of the two products to form the final expression $\frac{1984}{793} \times 3 = \frac{5952}{793}$. It verifies that the fraction is irreducible and computes the sum $m + n = 5952 + 793 = 6745$.

From \textbf{Steps 308–418}, R1 performs extensive verification of its calculations. It revisits the telescoping logic, checks numerical computations, and cross-references the simplified forms with the original problem statement. Despite initial doubts about the size of the result, R1 confirms the correctness of the solution through multiple validation steps.

In \textbf{Step 419}, R1 concludes with the final answer:
\[
\boxed{6745}.
\]

\end{R1box}

\begin{wrongbox}
\textbf{\Large Failure Reasoning}

\vspace{1\baselineskip}

In \textbf{Steps 39-40}, the remaining terms in the numerator and in the denominator are fault

In \textbf{Steps 72}, it forget that the last term is k=63, which influences the later steps

The remaining numerators and denominators after cancellation are wrong in many steps after Step 72.

\end{wrongbox}
\begin{humanbox}
\textbf{\Large Human Reasoning}

\vspace{1\baselineskip}

We can move the exponents to the front of the logarithms like this:
$$\frac{\log_{4}(5^{15})}{\log_{5}(5^{12})} \cdot \frac{\log_{5}(5^{24})}{\log_{6}(5^{21})} \cdot \frac{\log_{6}(5^{35})}{\log_{7}(5^{32})} \ldots = \frac{15 \log_{4}(5)}{12 \log_{5}(5)} \cdot \frac{24 \log_{5}(5)}{21 \log_{6}(5)} \cdot \frac{35 \log_{6}(5)}{32 \log_{7}(5)} \ldots$$
Now we multiply the logs and fractions separately. Let's do it for the logs first:
$$
\frac{\log_{4}(5)}{\log_{5}(5)} \cdot \frac{\log_{5}(5)}{\log_{6}(5)} \cdot \frac{\log_{6}(5)}{\log_{7}(5)} \ldots \frac{\log_{63}(5)}{\log_{64}(5)} = \frac{\log_{4}(5)}{\log_{64}(5)} = 3$$
Now fractions:
$$\frac{15}{12} \cdot \frac{24}{21} \cdot \frac{35}{32} \ldots = \frac{3 \cdot 5}{2 \cdot 6} \cdot \frac{4 \cdot 6}{3 \cdot 7} \cdot \frac{5 \cdot 7}{4 \cdot 8} \ldots \frac{62 \cdot 64}{61 \cdot 65} = \frac{5}{2} \cdot \frac{62}{65} = \frac{31}{13}$$

Multiplying these together gets us the original product, which is $\frac{31}{13} \cdot 3 = \frac{93}{13}$. Thus $m + n = \boxed{106}$.

\end{humanbox}

\newpage
\subsection{AIME\_1\_20}
\begin{questionbox}
\textit{Suppose $\triangle ABC$ has angles $\angle BAC = 84^\circ, \angle ABC=60^\circ,$ and $\angle ACB = 36^\circ.$ Let $D, E,$ and $F$ be the midpoints of sides $\overline{BC}, \overline{AC},$ and $\overline{AB},$ respectively. The circumcircle of $\triangle DEF$ intersects $\overline{BD}, \overline{AE},$ and $\overline{AF}$ at points $G, H,$ and $J,$ respectively. The points $G, D, E, H, J,$ and $F$ divide the circumcircle of $\triangle DEF$ into six minor arcs, as shown. Find $DE+2\cdot HJ + 3\cdot FG,$ where the arcs are measured in degrees.}
\end{questionbox}
\begin{R1box}
\textbf{\Large R1 Reasoning}\vspace{1\baselineskip}

In \textbf{Steps 1–8}, the model correctly identifies the key geometric figures. It recognizes that triangle DEF is the medial triangle of ABC and correctly deduces that its circumcircle is the nine-point circle of ABC.

In \textbf{Steps 9–87}, the model explores two main solution paths. It briefly considers a coordinate geometry approach but notes it would be tedious. It then explores a more elegant geometric approach. In \textbf{Steps 82–83}, it incorrectly questions whether the medial triangle is similar to the original, but immediately self-corrects, correctly concluding that triangle DEF is similar to ABC and thus has the same angles (84°, 60°, 36°). This leads to the crucial and correct deduction in \textbf{Step 87} that the arc subtended by side DE on the circumcircle of DEF is given by the inscribed angle theorem: $\text{arc DE} = 2 \times \angle DFE = 2 \times \angle C = 2 \times 36^\circ = 72^\circ$.

In \textbf{Steps 88–288}, despite having found a promising geometric path, the model abandons it and commits fully to the tedious coordinate geometry approach using floating-point approximations. It assigns approximate coordinates to the vertices A, B, C, then calculates the approximate coordinates for D, E, F. It then finds the approximate equation of the circumcircle of DEF. It proceeds to find the approximate coordinates of the intersection points G, H, and J by parametrizing the medians and solving the resulting quadratic equations. This is a very long and error-prone series of numerical calculations.

In \textbf{Steps 289–296}, using the approximate coordinates of the six points on the circle, the model calculates the central angles for the required arcs by computing dot products of vectors from the approximate center. Its calculations yield:
\begin{itemize}
    \item $\text{ DE} \approx 72^\circ$ (which fortuitously matches its earlier exact geometric deduction)
    \item $\text{ HJ} \approx 24^\circ$
    \item $\text{ FG} \approx 72^\circ$
\end{itemize}
Using these values, it calculates the final sum: $\text{DE} + 2 \cdot \text{HJ} + 3 \cdot \text{FG} \approx 72 + 2(24) + 3(72) = 72 + 48 + 216 = 336^\circ$.

In \textbf{Steps 297–345}, the model expresses doubt about its approximate numerical method. In a final leap of faith, it spots a pattern in \textbf{Step 342}, testing if the answer could be a weighted sum of the original angles of triangle ABC. It discovers that $\angle A + 2\angle C + 3\angle B = 84^\circ + 2(36^\circ) + 3(60^\circ) = 84 + 72 + 180 = 336^\circ$. Since this elegant formula matches its tedious numerical result, it adopts this as the justification for its answer, even though it has not formally proven why the arcs correspond to the angles in this specific way.
\end{R1box}

\begin{humanbox}
    \textbf{\Large Human Reasoning}\vspace{1\baselineskip}

    Firstly, connect FG and JE. The corresponding diagram can be easily drawn. Now FJDE is a cyclic quadrilateral. From median point we can derive three parallel relationships. Then apparently $\angle AFE = 60^\circ $ and $\angle EFD = 36^\circ$. By cyclic quadrilateral's property we know that $\angle AFD + \angle JED = 180^\circ $. Since $\angle FED = 60^\circ$, we have 
    \begin{itemize}
        \item $\angle JEF = 24^\circ$
        \item $\angle FJE = 96^\circ$
         \item $\angle AEJ = 12^\circ$
    \end{itemize}

    Furthermore, we can derive that $\angle FGD = 120^\circ$ by cyclic quadrilateral FGDE. Then we have:
    $$FG= DE \quad \text{and the inscribed angle} \angle EFD = 36^\circ
    $$

    $$HJ \text{'s inscribed angle}\quad \angle AEJ = 12^\circ
    $$
    Then we have our final answer:
    $$arc DE + 2*arc HJ + 3*arc FG = 2 * 36 + 2*(2 * 12) + 3 * (36 * 2) = 336 $$
    
\end{humanbox}

\newpage
\subsection{AIME\_1\_21}
\begin{questionbox}
\textit{Circle $\omega_1$ with radius $6$ centered at point $A$ is internally tangent at point $B$ to circle $\omega_2$ with radius $15$. Points $C$ and $D$ lie on $\omega_2$ such that $\overline{BC}$ is a diameter of $\omega_2$ and $\overline{BC} \perp \overline{AD}$. The rectangle $EFGH$ is inscribed in $\omega_1$ such that $\overline{EF} \perp \overline{BC}$, $C$ is closer to $\overline{GH}$ than to $\overline{EF}$, and $D$ is closer to $\overline{FG}$ than to $\overline{EH}$, as shown. Triangles $\triangle DGF$ and $\triangle CHG$ have equal areas. The area of rectangle $EFGH$ is $\frac{m}{n}$, where $m$ and $n$ are relatively prime positive integers. Find $m+n$.}
\end{questionbox}

\begin{R1box}
\textbf{\Large R1 Reasoning}\vspace{1\baselineskip}

In \textbf{Steps 1–23}, the model correctly interprets the geometric setup. It establishes a coordinate system with the center of circle $\omega_1$ at A=(0,0). By using the properties of internal tangency, it correctly deduces the key coordinates of the problem:
\begin{itemize}
    \item Center of $\omega_1$: A at (0,0) with radius $r_1=6$.
    \item Center of $\omega_2$: O at (-9,0) with radius $r_2=15$.
    \item Point of tangency: B at (6,0).
    \item The diameter of $\omega_2$ is BC, so C is at (-24,0).
    \item D lies on $\omega_2$ and AD is perpendicular to BC (the x-axis), so D is at (0,12).
\end{itemize}

In \textbf{Steps 24–40}, the model analyzes the constraints on the rectangle EFGH. It correctly deduces that since its sides are parallel/perpendicular to BC, it must be an axis-aligned rectangle centered at A=(0,0). It correctly parametrizes the vertices as E(a, -b), F(a, b), G(-a, b), H(-a, -b). Since the rectangle is inscribed in $\omega_1$, it correctly establishes the relationship $a^2 + b^2 = 6^2 = 36$. The goal is to find the area, $4ab$.

In \textbf{Steps 41–94}, the model uses the crucial condition that the areas of triangles DGF and CHG are equal. It sets up the coordinates for the vertices of each triangle:
\begin{itemize}
    \item $\triangle DGF$: Vertices D(0,12), G(-a, b), F(a, b).
    \item $\triangle CHG$: Vertices C(-24,0), H(-a, -b), G(-a, b).
\end{itemize}
Using the shoelace formula (or base times height), it correctly calculates the areas:
\begin{itemize}
    \item Area($\triangle DGF$) = $a(12-b)$
    \item Area($\triangle CHG$) = $b(24-a)$
\end{itemize}
By setting the two areas equal, $a(12-b) = b(24-a)$, it correctly simplifies the expression to find the key relationship $12a = 24b$, or $a = 2b$.

In \textbf{Steps 95–147}, the model combines the two derived equations to solve the system. It substitutes $a=2b$ into $a^2+b^2=36$ to get $(2b)^2+b^2=36$, which correctly yields $5b^2=36$. It solves for $a$ and $b$, finding $a=12/\sqrt{5}$ and $b=6/\sqrt{5}$. It then correctly calculates the area of the rectangle as $4ab = 4(12/\sqrt{5})(6/\sqrt{5}) = 288/5$. This gives $m=288$ and $n=5$. Finally, it computes the required sum $m+n = 288+5 = 293$. The model also performs a sanity check against coordinates mentioned in the problem's source code, correctly concluding the diagram was illustrative and not to scale, which reinforces the confidence in its analytical solution.
\end{R1box}
\begin{humanbox}
      \textbf{\Large Human Reasoning}\vspace{1\baselineskip}

    Note that the radius are 6 and 15. Then:
    $$AC = 24 , AB = 6$$ Let AD = h, then:
    $$h^2 + 576 + h^2 + 36 = 900$$and finally
    $$h = 12$$
    Let GF = 2b, GH = 2a. It is easy to know that:
    $$4a^2 + 4b^2 = 144$$ 
    From the equity of two areas of the respective triangle, we can derive  such relationship between $a$ and $b$:
    \begin{align}
        2b(12-a) &= 2a(24-b) \\
        b &= 2a
    \end{align}

By cyclic quadrilateral GHEF we can derive $a^2 + b^2 = 36$,then :
$$a = \frac{6}{\sqrt{5}} ,\quad b= \frac{12}{\sqrt{5}}$$
The area of the desired rectangle is $4ab$,then:
$$4ab = \frac{288}{5} = \frac{m}{n} \quad\text{since 288 and 5 is coprime}$$ and $$ m + n = 288 + 5 = 293$$ 
    
\end{humanbox}

\newpage
\subsection{AIME\_1\_22}
\begin{questionbox}
\textit{Let $A$ be the set of positive integer divisors of $2025$. Let $B$ be a randomly selected subset of $A$. The probability that $B$ is a nonempty set with the property that the least common multiple of its element is $2025$ is $\frac{m}{n}$, where $m$ and $n$ are relatively prime positive integers. Find $m+n$.}
\end{questionbox}

\begin{R1box}
\textbf{\Large R1 Reasoning}

\vspace{1\baselineskip}

In \textbf{Steps 1--7}, R1 begins by understanding the problem structure. It  computes the prime factorization ($3^4 \times 5^2$) of $A$, and determines $|A| = 15$. R1 then identifies the sample space as all $2^{15}$ subsets of $A$, with favorable cases being non-empty subsets where $\text{LCM}(B) = 2025$.

In \textbf{Steps 8--14}, R1 recognizes that $\text{LCM}(B) = 2025$ iff $B$ contains at least one divisor with exponent $4$ for prime $3$ (i.e., divisible by $81$) and one with exponent $2$ for prime $5$ (i.e., divisible by $25$). It lists the relevant divisors: $3$ multiples of $81$ ($\{81, 405, 2025\}$) and $5$ multiples of $25$ ($\{25, 75, 225, 675, 2025\}$), noting $2025$ is common to both sets.

From \textbf{Steps 15--31}, R1 sets up the inclusion-exclusion principle to count subsets satisfying both conditions. After resolving concerns about empty-set handling, it derives the expression:  
\[
2^{15} - 2^{12} - 2^{10} + 2^{8}
\]
where $2^{12}$ excludes multiples of $81$, $2^{10}$ excludes multiples of $25$, and $2^{8}$ corrects for over-subtraction of subsets excluding both.

In \textbf{Steps 32--44}, R1 computes:  
$2^{15} = 32768$, $2^{12} = 4096$, $2^{10} = 1024$, $2^{8} = 256$  
yielding $32768 - 4096 - 1024 + 256 = 27904$ favorable subsets. The probability simplifies to $\frac{27904}{32768} = \frac{109}{128}$, giving $m + n = 237$.

Through \textbf{Steps 45--60}, R1 verifies that the inclusion-exclusion formula inherently excludes the empty set and correctly counts non-empty valid subsets.

In \textbf{Steps 61--100}, R1 validates the approach with a small case ($A = \text{divisors of }6$). Manual listing confirms the inclusion-exclusion count for subsets with $\text{LCM}=6$ matches the derived formula, resolving an initial discrepancy in auxiliary reasoning.

From \textbf{Steps 101--112}, R1 considers an alternative method treating prime exponents independently but rejects it after finding multiplicative counts violate subset dependencies.

In \textbf{Steps 113--116}, R1 concludes that the inclusion-exclusion result is correct and finalizes the answer \boxed{237}.

\end{R1box}

\begin{wrongbox}

\textbf{\Large Failure Reasoning}

\vspace{1\baselineskip}

In \textbf{Steps 89}, miscalculating the number of subsets with 2 and 3 but not 6 as 4 instead of 2.
\end{wrongbox}

\begin{humanbox}
\textbf{\Large Human Reasoning}

\vspace{1\baselineskip}

We split into different conditions:

Note that the numbers in the set need to have a least common multiple of 2025, so we need to ensure that the set has at least 1 number that is a multiple of $3^4$ and a number that is a multiple of $5^2$.

Multiples of $3^4$: 81, 405, 2025

Multiples of $5^2$: 25, 75, 225, 675, 2025

If the set $B$ contains 2025, then all of the rest 14 factors is no longer important. The valid cases are $2^{14}$.

If the set $B$ doesn't contain 2025, but contains 405, we just need another multiple of $5^2$. It could be 1 of them, 2 of them, 3 of them, or 4 of them, which has $2^4 - 1 = 15$ cases. Excluding 2025, 405, 25, 75, 225, 675, the rest 9 numbers could appear or not appear. Therefore, this case has a valid case of $15 \cdot 2^9$.

If set $B$ doesn't contain 2025 nor 405, it must contain 81. It also needs to contain at least 1 of the multiples from $5^2$, where it would be $15 \cdot 2^8$.

The total valid cases are $2^{14} + 15 \cdot (2^9 + 2^8)$, and the total cases are $2^{15}$.

The answer is $\frac{2^8 \cdot (64 + 30 + 15)}{2^8 \cdot 2^7} = \frac{109}{128}$.

Desired answer: $109 + 128 = \boxed{237}$.

\end{humanbox}

\newpage
\subsection{AIME\_1\_23}
\begin{questionbox}
\textit{From an unlimited supply of 1-cent coins, 10-cent coins, and 25-cent coins, Silas wants to find a collection of coins that has a total value of $N$ cents, where $N$ is a positive integer. He uses the so-called greedy algorithm, successively choosing the coin of greatest value that does not cause the value of his collection to exceed $N.$ For example, to get 42 cents, Silas will choose a 25-cent coin, then a 10-cent coin, then 7 1-cent coins. However, this collection of 9 coins uses more coins than necessary to get a total of 42 cents; indeed, choosing 4 10-cent coins and 2 1-cent coins achieves the same total value with only 6 coins. In general, the greedy algorithm succeeds for a given $N$ if no other collection of 1-cent, 10-cent, and 25-cent coins gives a total value of $N$ cents using strictly fewer coins than the collection given by the greedy algorithm. Find the number of values of $N$ between $1$ and $1000$ inclusive for which the greedy algorithm succeeds.}
\end{questionbox}

\begin{R1box}
\textbf{\Large R1 Reasoning}

\vspace{1\baselineskip}

In \textbf{Steps 1–27}, R1 begins by converting the problem statement into mathematical formulation. R1 defines a series of variables according to the following principles:
In the initial phase of the derivation, the greedy algorithm's coin count is determined by decomposing $N$ into $q_{25}=\lfloor N/25 \rfloor$, $r_1=N-25 \times q_{25}$, $q_{10}=\lfloor r_1/10 \rfloor$, and $r_2=r_1-10 \times q_{10}$, yielding a total of $q_{25}+q_{10}+q_{1}$ coins where $q_1 = r_2$. Subsequent analysis indicates that for a fixed $q_{25}$, the selection strategies for 10-cent and 1-cent coins remain identical between the greedy and optimal algorithms. This suggests that any divergence in optimality must stem from the over-selection of 25-cent coins.

In \textbf{Steps 28–103}, R1 posits that the success pattern of the greedy algorithm may be linked to a periodicity of 25, thus expressing the total value as $N = 25k + r$, where $0 \le r < 25$. Under the greedy paradigm, the number of coins is given by $k + \lfloor r/10 \rfloor + (r \bmod 10)$. Assuming that the greedy algorithm fails when one fewer 25-cent coin is taken, the reasoning trajectory leads to the following inequality:
\begin{equation}
    (k - 1) + \lfloor (25 + r)/10 \rfloor + ((25 + r) \bmod 10) < k + \lfloor r/10 \rfloor + (r \bmod 10)
\end{equation}
which simplifies to
\begin{equation}
    \lfloor (25 + r)/10 \rfloor + ((25 + r) \bmod 10) - \lfloor r/10 \rfloor - (r \bmod 10) < 1. \label{eq:inequality_1}
\end{equation}

To facilitate solving this inequality, R1 parameterises $r$ as $10a + b$, where $a = \lfloor r/10 \rfloor \in \{0, 1, 2\}$ and $b = r \bmod 10$. Rewriting the sum as $25 + r = 10(a + 2) + (b + 5)$, the analysis branches into two exhaustive cases:
\begin{description}
    \item[Case 1:] If $b + 5 < 10$, then $\lfloor (25 + r)/10 \rfloor = a + 2$ and $(25 + r) \bmod 10 = b + 5$.
    \item[Case 2:] If $b + 5 \ge 10$, then $\lfloor (25 + r)/10 \rfloor = a + 3$ and $(25 + r) \bmod 10 = b - 5$.
\end{description}
For case 1, solving inequality (1) obtains:
\[
(a+2)+(b+5)-a-b<1\Rightarrow7<1
\]
\end{R1box}
\begin{R1box}
Therefore, case 1 is not valid. For case 1, solving inequality (1) obtains:
\[
(a+3)+(b-5)-a-b<1\Rightarrow-2<1
\]
Therefore, when $b \geq5$, replacing one $25$ cent with dimes and pennies reduces the total coins by $3$. This implies that the greedy algorithm succeeds when $b<5$.

In \textbf{Steps 104–123}, R1 computes specific examples to verify the correctness of the conclusions derived in the preceding derivation.

In \textbf{Steps 124–171}, within the range of $0 \leq r < 25$, R1 enumerates all values of $r$ that satisfy $b= (r \text{ mod } 10) < 5$ as follows:
\[
0,1,2,3,4,10,11,12,13,14,20,21,22,23,24
\]
‌There are $15$ such $r$ in total. For $1\leq N\leq1000,N=25k+r,0\leq r<25$,‌ R1 finds that the complete set of remainders r from 0 to 24 repeats exactly 40 times. Therefore, R1 concludes the final answer is $40\times15=600$.

In \textbf{Steps 172–226}, R1 recognizes that when $0 < N < 25$, the inability to use a 25-cent coin naturally precludes any attempt to optimize the coin acquisition method by reducing one such coin. This scenario requires separate consideration, thus prompting R1 to categorize the approach into two cases:
\begin{enumerate}
    \item Case 1: For $0<N<25$, The greedy algorithm is optimal when using only 1-cent and 10-cent coins, with $24$ numbers involved in this case.
    \item Case 2: For $25\leq N\leq1000$, by applying the above conclusion, R1 counts the number of remainders that make the greedy algorithm succeed:
    \begin{itemize}
        \item $r=0$, there are $40$ cases in total;
        \item $r=1:4$, there are $39\times4$ cases in total;
        \item $r=10:14$, there are $39\times5$ cases in total;
        \item $r=20:24$, there are $39\times5$ cases in total;
    \end{itemize}
\end{enumerate}
In summary, R1 yields a final result of $\boxed{610}$.

In \textbf{Steps 227–311}, R1 validates‌ the correctness of answer $610$ through repeated verification iterations and case demonstrations.

\end{R1box}

\begin{wrongbox}
\textbf{\Large Failure Reasoning}

\vspace{1\baselineskip}

The failure in R1's reasoning process can be attributed to two main issues:

\textbf{1. The assumptions lack sufficient consideration.} Introducing the supplemental assumption $k\geq1$ when reducing a 25-cent coin enables an error-free solution process during optimal solution exploration.

\textbf{2. The solution process is unnecessarily lengthy.} R1 conducts numerous repeated verifications, and exemplification cannot serve as a means for conclusion validation—only as a starting point for problem comprehension.

\end{wrongbox}
\begin{humanbox}
\textbf{\Large Human Reasoning}

\vspace{1\baselineskip}

‌When solving this problem, humans mainly adopt the enumeration method. It should be noted that the problem can be divided into three cases for discussion:‌

Case 1:‌ when $1\leq N\leq25$, the greedy algorithm is clearly successful for all integers in this range — there are exactly $25$ such numbers.

Case 2: when $26\leq N\leq50$, the cases are presented in the following table:

\begin{table}[H]
\centering
\begin{tabular}{cccc}
  \toprule
  $N$ & \textbf{Greedy Algorithm} & \textbf{‌Minimum Coin Selection} &\textbf{‌Success / Failure}\\
  \midrule
  $26$ & $25\times1+1\times1$ & $25\times1+1\times1$ & Success\\
  $27$ & $25\times1+1\times2$ & $25\times1+1\times2$ & Success\\
  $28$ & $25\times1+1\times3$ & $25\times1+1\times3$ & Success\\
  $29$ & $25\times1+1\times4$ & $25\times1+1\times4$ & Success\\
  $30$ & $25\times1+1\times5$ & $10\times3+1\times0$ & Failure\\
  $31$ & $25\times1+1\times6$ & $10\times3+1\times1$ & Failure\\
  $32$ & $25\times1+1\times7$ & $10\times3+1\times2$ & Failure\\
  $33$ & $25\times1+1\times8$ & $10\times3+1\times3$ & Failure\\
  $34$ & $25\times1+1\times9$ & $10\times3+1\times4$ & Failure\\
  $35$ & $25\times1+10\times1+1\times0$ & $25\times1+10\times1+1\times0$ & Success\\
  $36$ & $25\times1+10\times1+1\times1$ & $25\times1+10\times1+1\times1$ & Success\\
  $37$ & $25\times1+10\times1+1\times2$ & $25\times1+10\times1+1\times2$ & Success\\
  $38$ & $25\times1+10\times1+1\times3$ & $25\times1+10\times1+1\times3$ & Success\\
  $39$ & $25\times1+10\times1+1\times4$ & $25\times1+10\times1+1\times4$ & Success\\
  $40$ & $25\times1+10\times1+1\times5$ & $25\times1+10\times1+1\times5$ & Failure\\
  $41$ & $25\times1+10\times1+1\times6$ & $10\times4+1\times1$ & Failure\\
  $42$ & $25\times1+10\times1+1\times7$ & $10\times4+1\times2$ & Failure\\
  $43$ & $25\times1+10\times1+1\times8$ & $10\times4+1\times3$ & Failure\\
  $44$ & $25\times1+10\times1+1\times9$ & $10\times4+1\times4$ & Failure\\
  $45$ & $25\times1+10\times2+1\times0$ & $25\times1+10\times2+1\times0$ & Success\\
  $46$ & $25\times1+10\times2+1\times1$ & $25\times1+10\times2+1\times1$ & Success\\
  $47$ & $25\times1+10\times2+1\times2$ & $25\times1+10\times2+1\times2$ & Success\\
  $48$ & $25\times1+10\times2+1\times3$ & $25\times1+10\times2+1\times3$ & Success\\
  $49$ & $25\times1+10\times2+1\times4$ & $25\times1+10\times2+1\times4$ & Success\\
  $50$ & $25\times2$ & $25\times2$ & Success\\
  \bottomrule
\end{tabular}
\end{table}

In this case, a total of $15$ numbers enable the greedy algorithm to succeed.

Case 3: when $N>50$, subtracting an appropriate multiple of $25$ from $N$ causes it to fall into case 2. This indicates that within ‌case 3‌, $N$ exhibits periodic behavior with a period of $25$, where ‌$15$ numbers per period‌ enable the greedy algorithm to succeed.

In conclusion, the final answer to the problem is:
\[
\frac{15}{25}\times975+25=\boxed{610}.
\]

\end{humanbox}

\subsection{AIME\_1\_24}
\begin{questionbox}
\textit{There are $n$ values of $x$ in the interval $0<x<2\pi$ where $f(x)=\sin(7\pi\cdot\sin(5x))=0$. For $t$ of these $n$ values of $x$, the graph of $y=f(x)$ is tangent to the $x$-axis. Find $n+t$.}
\end{questionbox}
\begin{R1box}
\textbf{\Large R1 Reasoning}

\vspace{1\baselineskip}

In \textbf{Steps 1}, ‌By definition, R1 obtains $n$ as the number of roots and $t$ as the number of double roots (where both the function value and its first derivative are zero) for the same equation. The final answer is $n+t$.

In \textbf{Steps 1–40}, R1 first solves the equation $f(x)=\sin(7\pi\sin(5x))=0$. This function equals zero when $7\pi\sin(5x)$ is an integer multiple of $\pi$. Specifically, it holds when:
\[
7\pi\sin(5x)=k\pi,\text{where } k\text{ is an integer}\Rightarrow\sin(5x)=\frac{k}{7}
\]
Since the range of $\sin(5x)$ is $[-1,1]$, $k$ must satisfy $|k|\leq7$ and be an integer. Therefore, $k$ has $15$ possible integer values. Each integer value corresponds to solutions for $x$.

Core Problem: Solving $\sin(5x)=k/7$ for $x\in(0,2\pi)$ (equivalent to $\sin\theta=k/7$ for $\theta\in(0,10\pi)$).

General Case ($0<|k/7|<1$): 
The sine function has period $2\pi$. The interval $10\pi$ spans $5$ full periods. Within each period, $\sin\theta=c(0<|c|<1)$ has $2$ solutions. Total solutions: $5$ periods $\times$ $2$ solutions/period = $10$ solutions. 

Special Cases ($|k/7|=0,1$):

\begin{enumerate}
    \item $k=\pm7$ ($\sin\theta=\pm1$): $\sin\theta=\pm1$ has only $1$ solution per period. Total solutions: $5$ periods $\times$ $1$ solutions/period = $5$ solutions (each for $k=7$ and $k=-7$).
    \item $k=0$ ($\sin\theta=0$): Solutions occur where $\theta=n\pi$ ($n$ is an integer). Within the open interval $(0,10\pi)$, the solutions are at $\theta=\pi,2\pi,\cdots,9\pi$, giving $9$ solutions. 
\end{enumerate}

Based on the synthesis of the general and special cases above, R1 concludes:
\begin{align*}
    n &=5(k=-7) + 10\times6(k=-6:-1)+9(k=0)+10\times6(k=1:6)+5(k=7)\\
      &=139
\end{align*}
‌Then R1 recounts the number of roots and confirms that $n=139$.

In \textbf{Steps 41–50}, R1 attempts to verify the results once again by calculating specific examples. For $k=7$, $\sin(5x)=1$. So $5x = \frac{\pi}{2} + 2\pi n$. So $x = \frac{\pi}{10} + \frac{2\pi}{5}n$. Then $x$ must be in $(0,2\pi)$. Then R1 solves for $n$:
\begin{align*}
0 < \frac{\pi}{10} + \frac{2\pi}{5}n < 2\pi \Rightarrow -\frac{1}{4} < n < \frac{19}{4}
\end{align*}
So, $n$ can be $0,1,2,3,4$. So, $5$ solutions. 
‌R1 determines that the equation has five roots when $k=-7$, following the same approach.
\end{R1box}
\begin{R1box}
In \textbf{Steps 51–73}, R1 begins to solve for t, first requiring the first derivative of f(x):
\[
f^{'}(x)=35\pi\cdot\cos(5x)\cdot\cos(7\pi\cdot\sin(5x))
\]
So for the derivative to be zero, either $\cos(5x)=0$ or $\cos(7\pi \sin(5x))=0$. R1 wants to find points where $f(x)=0$ and $f^{'}(x)=0$ simultaneously. This requires:

\[ 
\left\{
\begin{aligned}
    & \sin(7\pi\sin(5x)) = 0 \\
    & \cos(5x) = 0\text{ or }\cos(7\pi\sin(5x))=0 \\
\end{aligned}
\right.
\]
From $\sin(7\pi\sin(5x)) = 0$, $\sin(5x)=k/7$, where $k$ is an integer. If $\cos(7\pi\sin(5x))=0$, then $7\pi\sin(5x)=(m+1/2)\pi$, where $m$ is an integer. Then, R1 gets $k=m+1/2$, which contradicts the assumption that $k$ is an integer. So when $\sin(7\pi \sin(5x))=0$, ‌$\cos(7\pi \sin(5x))=0$ is impossible‌. 

For $f(x)=0,f^{'}(x)=0$ necessarily requires $\cos(5x)=0$, which implies:
\[
5x=\frac{\pi}{2}+n\pi\Rightarrow x=\frac{\pi}{10}+\frac{n\pi}{5},n\in\mathbb{Z} 
\]
Given $x\in (0, 2\pi)$, R1 deduces the inequality:
\[
0<\frac{\pi}{10}+\frac{n\pi}{5}<2\pi\Rightarrow -0.5<n<9.5
\]
So $x= \frac{\pi}{10} + \frac{n\pi}{5}$, $n=0$ to $9$. So, $10$ solutions. Then R1 verifies by substitution whether the solutions obtained satisfy $\sin(7\pi\sin(5x))=0$:
\begin{align*}
    \sin(7\pi\sin(5x)) &= \sin(7\pi\sin(\frac{\pi}{2}+n\pi))\\
    &=\sin(7\pi\cdot(-1)^n)=0
\end{align*}
Therefore, all the solutions of $\cos(5x)=0$ in $(0,2\pi)$ are also solutions of $f(x)=0$. So $t=10$. After thorough verification of the procedure, R1 gives the final solution $n+t=\boxed{149}$.

\end{R1box}

\begin{humanbox}
\textbf{\Large Human Reasoning}

\vspace{1\baselineskip}

Let $f(x)=\sin(7\pi\cdot\sin(5x))=0,x\in(0,2\pi)$, then we have:
\begin{align*}
    & \text{case 1}:\sin(5x)=0 \\
    & \text{case 2}:\sin(5x)=\pm\frac{t}{7},t=1,2,3,4,5,6,7
\end{align*}
Then we draw the graph of $y=\sin(5x)$ to help determine the number of roots of the equation $f(x)=0,x\in(0,2\pi)$.
\end{humanbox}
\begin{humanbox}
\begin{figure}[H]
  \centering
  \includegraphics[width=0.8\textwidth]{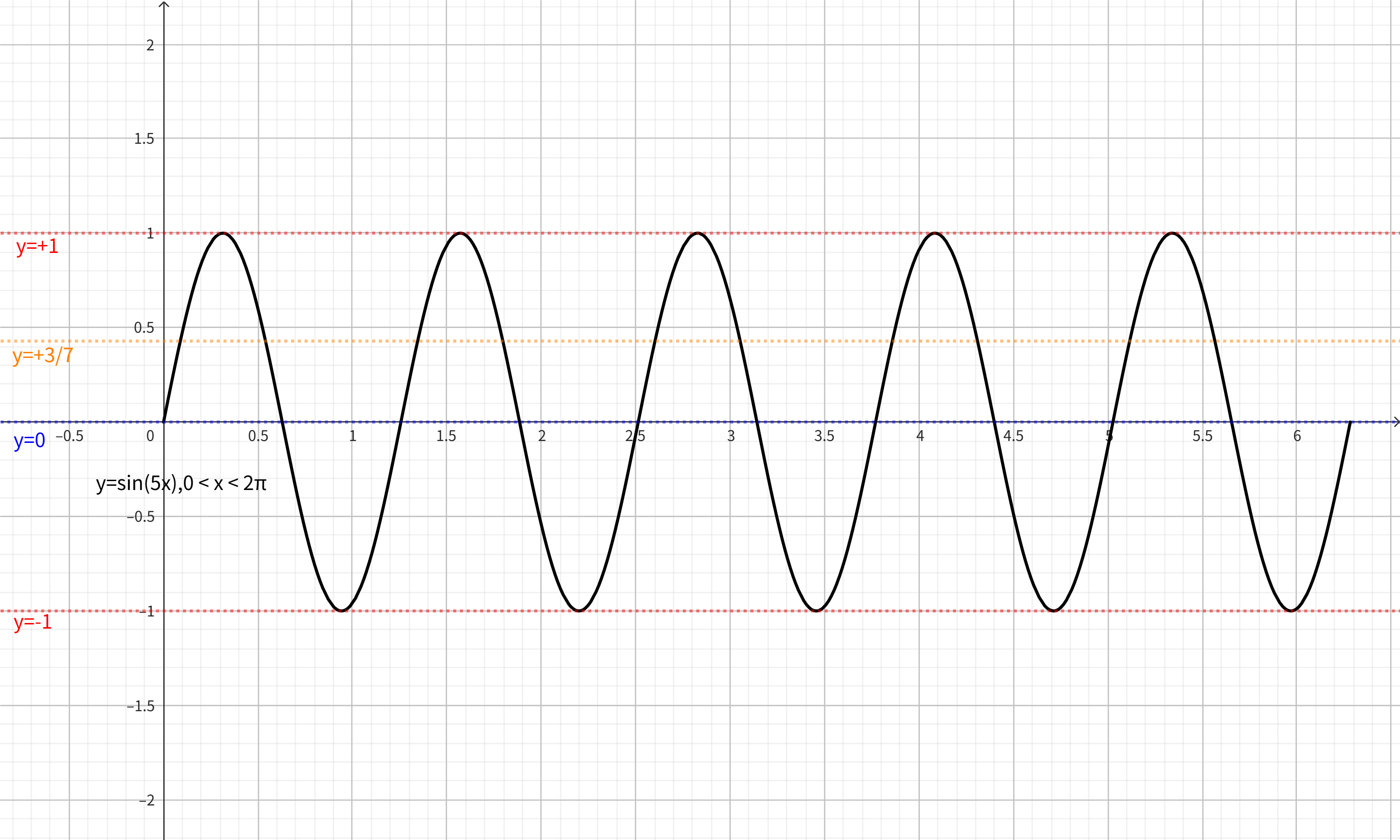} 
  \caption{Sketch map}
\end{figure}

For case 1, we observe that $y=\sin(5x)$ and $y=0$ have $9$ intersection points within $0<x<2\pi$.

For case 2, we discuss two scenarios:
\begin{itemize}
    \item For the curves $y=\sin(5x)$ and $y=\pm\frac{7}{7}$ in the interval $0<x<2\pi$, each pair has $5$ intersection points, resulting in a total of $10$ intersection points.
    \item For each $t\in{1,2,3,4,5,6}$, the curve $y=\sin(5x)$ and the lines $y=\pm\frac{t}{7}$ each have $10$ points in the same interval. Thus, the total number of intersection points is $120$ (an example in the figure 1 shows the intersection between $y=\sin(5x)$ and $y=+3/7$.
\end{itemize}

To summarize, the equation $f(x)=0$ whit $x\in(0,2\pi)$ has $139$ roots, that is, $n=139$.

Next, we need to select $t$ roots from these $n$ roots such that the graph of $y=\sin(5x)$ is tangent to the $x$-axis at these points. For this purpose, we must first compute the derivative of $f(x)$, set $f^{'}(x)=0$, and analyze the resulting equation.
\[
f^{'}(x)=35\pi\cdot\cos(7\pi\cdot\sin(5x))\cdot\cos(5x)=0
\]

Then we have:
\begin{align*}
    & \text{case 1}:\cos(5x)=0 \\
    & \text{case 2}:\cos(7\pi\cdot\sin(5x))=\cos(7\pi\cdot\pm\frac{t}{7})=\cos(\pm t\pi)=0
\end{align*}

For case 1, when $\cos(5x)=0,\sin(5x)=\pm1$, thus corresponding to $t=\pm7$. There are exactly $10$ roots that satisfy this condition. 

For Case 2, when $t\in\{0,1,2,3,4,5,6,7\},\cos(\pm t\pi)\in\{-1,1\}$. Therefore, the equation $\cos(7\pi\sin(5x))=0$ has no solution. 

‌To conclude, $t=10$ and $n+t=\boxed{149}$.
\end{humanbox}

\newpage
\subsection{AIME\_1\_25}
\begin{questionbox}
\textit{Sixteen chairs are arranged in a row. Eight people each select a chair in which to sit so that no person sits next to two other people. Let $N$ be the number of subsets of $16$ chairs that could be selected. Find the remainder when $N$ is divided by $1000$.}
\end{questionbox}

\begin{R1box}
\textbf{\Large R1 Reasoning}\vspace{1\baselineskip}

In \textbf{Steps 1–6}, the model correctly interprets the problem. It initially struggles with the condition "no person sits next to two other people," but after consideration, it correctly deduces this means that no three consecutive chairs can be selected. It correctly frames the problem as counting the number of subsets of 8 chairs from 16, with no three chairs being consecutive.

In \textbf{Steps 7–22}, the model reframes the problem as counting binary strings of length 16 with exactly eight 1s and no occurrence of the substring '111'. It then derives a correct recurrence relation for $f(n,k)$, the number of such strings of length $n$ with $k$ ones. By considering the possible endings of a valid string (0, 10, or 110), it arrives at:
\[f(n, k) = f(n - 1, k) + f(n - 2, k - 1) + f(n - 3, k - 2)\]

In \textbf{Steps 23–68}, the model verifies the recurrence relation with small test cases ($f(3,2)$ and $f(4,3)$), confirming its validity. It correctly establishes the necessary base cases, such as $f(n,0)=1$ and $f(n,k)=0$ for $n<k$. It briefly explores alternative methods like generating functions and inclusion-exclusion but correctly dismisses them as more complex than the dynamic programming approach suggested by the recurrence.

In \textbf{Steps 69–207}, the model systematically applies the recurrence relation to build a table of values for $f(n,k)$ in a dynamic programming fashion. It computes all necessary values step-by-step, starting from $n=0$ up to $n=16$. The process is long and involves many additions, but it is carried out correctly. The final target value is calculated as $f(16,8) = f(15,8) + f(14,7) + f(13,6) = 1107 + 1016 + 784 = 2907$.

In \textbf{Steps 208–257}, the model finds the required remainder, $2907 \pmod{1000} = 907$. Crucially, it then performs a thorough and systematic verification of its calculations, re-computing several of the key values that led to the final answer to ensure no arithmetic errors were made. This extensive self-checking confirms the accuracy of its result.

\end{R1box}
\begin{humanbox}
     \textbf{\Large Human Reasoning}\vspace{1\baselineskip}

     Denote the chairs eight people sit as $1, 2, 3, 4, 5, 6, 7, 8$. Firstly, if none of two chairs are together, it creates 9 total blocks, but we need to have 1 space between adjacent chairs. Since there are 7 spaces, we want to fill $8 - 7 = 1$ more chair among the nine blocks, which yields $9$ cases.

If two of the chairs are adjacent to each other, there are $\binom{7}{1}$ ways to select such pair. Then we need to fill the six spaces between the chairs by this condition. There are 8 spaces created by such way and we want to fill other 2 chairs in, so
\[
x_1 + x_2 + \dots + x_8 = 2 \quad \text{has} \quad \binom{9}{7} = \binom{9}{2} \quad \text{ways},
\]
so there are
\[
\binom{7}{1} \cdot \binom{9}{2} \quad \text{cases}.
\]

If two pairs of chairs are adjacent to each other, we could have $\binom{7}{2}$ ways to select but we have to subtract six cases where three chairs are adjacent to each other. By the similar reason above, for each case there are $\binom{9}{3}$ cases, so in this case we have
\[
\binom{9}{3} \cdot 15 \quad \text{cases}.
\]

If three pairs: $12, 34, 56$; $12, 34, 78$; $12, 45, 67$; $12, 45, 78$; $12, 56, 78$; $23, 45, 67$; $23, 45, 78$; $23, 56, 78$; $34, 56, 78$ -- in total 10 cases and for each there are
\[
\binom{9}{4} \quad \text{ways}.
\]

If four pairs, only one way and there are $\binom{9}{4}$ ways as well.

\vspace{1em}
Sum everything up the answer is:
\[
2907
\]
\end{humanbox}

\newpage
\subsection{AIME\_1\_26}
\begin{questionbox}
\textit{Let $S$ be the set of vertices of a regular $24$-gon. Find the number of ways to draw $12$ segments of equal lengths so that each vertex in $S$ is an endpoint of exactly one of the $12$ segments.}
\end{questionbox}

\begin{R1box}
\textbf{\Large R1 Reasoning}\vspace{1\baselineskip}

In \textbf{Steps 1–5}, R1 begins by understanding the problem, identify the length of segments with the distances of points in cyclic order, varing from $1$ to $12$.

In \textbf{Steps 6–11}, R1 looks through some examples to figure out the rule to groupify the vertices when the pace $k=1$.

In \textbf{Steps 12–41}, R1 looks for the possible valuesof the pace $k$. R1 firstly consider the divisibility $2k\mid n$ and then deny it. Then R1 finds that the key is the cycles activated by the pace $k$, i.e. the orbits induced by the action $+k$ on $\mathbb Z/24\mathbb Z$. So $k=8$ is impossible because each orbit consists odd number of points. 

In \textbf{Steps 42–80}, R1 understand the problem on a $24$-gon with vertices labeled, so each cycle is equipped $2$ different ways of matching. So the pace $k$ deduces $\mathrm{gcd}(24,k)$ cycles and then \[2^{\mathrm{gcd}(24,k)}\] matchings. At \textbf{Steps 59}, R1 notice the excpetion when the cycle is a $2$-cycle, relating to the case $k=12$.

In \textbf{Steps 81–106}, R1 misunderstands the word "crossing" introduced by itself, it claims it counts only the non-crossing cases in above, but the problem doesn't suppose that it's forced to be non-crossing. After a series of logicless nonsense, R1 confirms that the appropriate matchings are always non-crossing and the calculation is right.

In \textbf{Steps 81–167}, R1 sums the first time the numbers of matchings and gives $113$. TO verify, R1 sums the second time the numbers of matchings and gives $113$. R1 comments that $113$ is prime and weird so calculates it the third time, still $113$.

In \textbf{Steps 168–173}, R1 raises the famous Catalan number as a reflection, and then denies this thought because the Catalan number dosen't suppose the segments to be same lengths.
In \textbf{Steps 174–177}, R1 does this problem on square and hexagon, which is meaningless. 

In \textbf{Steps 178–222}, R1 sums the fourth time and fifth time the number of matchings, still $113$. And finally conclude.

\end{R1box}

\begin{wrongbox}
\textbf{\Large Failure Reasoning}\vspace{1\baselineskip}

R1 dose it in a different way from human, because the problem is more meaningful on the $24$-gon with vertices unlabeled. Even in R1's comprehension, it repeats plural times of sommation, and raise weird doubts (crossing cases, Catalan number).

\end{wrongbox}

\begin{humanbox}
\textbf{\Large Human Reasoning}\vspace{1\baselineskip}

A human approaching this problem would begin by noting \[S=\{s_k=(\cos\frac{k\pi}{12},\sin\frac{k\pi}{12});0\leq k\leq23\}\cong\mathbb Z/24\mathbb Z\] and calculating the length of segment connecting two vertices $s_k$ and $s_l$: \[|s_ks_l|=2\sin\frac{|k-l|\pi}{24}.\]So the lengths of segments connecting two vertices is bijective to $\{1,\cdots,11,12\}$ which represent the actions of addition on $\mathbb Z/24\mathbb Z$.

When $n=1$, $5$, $7$, $11$, $\mathrm{gcd}(n,24)=1$, the orbit of $+n$ is unique and each of them contribute one way to draw.

When $n=8$, each orbit of $+n$ has $3$ points, so the solution doesn't exist.

When $n=12$, each orbit of $+n$ has only $2$ points, so the solution is unique.

When $n=2$, $10$, there are $2$ orbits of $+n$ of $12$ points, so the solution is unique because by the symmetry, $s_5s_{19}$ ($s_1s_{23}$ for $n=2$) and $s_5s_{15}$ ($s_1s_3$ for $n=2$) are the same if we connect $s_0s_{10}$ ($s_0s_2$ for $n=2$) without loss of generality.

When $n=3$, $9$, there are $3$ orbits of $+n$ of $8$ points. On each orbit, we have two directions of a unique way to draw. So by the symmetry, there are two ways to draw for each $n$.

When $n=4$, there are $4$ orbits of $+n$ of $6$ points. Without loss of generality, suppose that we connect $s_0s_{4}$ and $s_2s_6$. Then by the symmetry, it leaves only two possibilities of $s_1s_5$ and $s_1s_21$.

When $n=6$, there are $6$ orbits of $+n$ of $4$ points. By the symmetry, we can combine each orbit $B$ with the orbit $B+3=\{b+3,b\in B\}$, then like the case of $n=3$ or $9$, it provides $2$ ways to draw.

In conclusion, we have $4\times1+0+1+2\times1+2\times2+2+2=15$ different ways to draw.

\end{humanbox}

\newpage
\subsection{AIME\_1\_27}
\begin{questionbox}
\textit{Let $A_1A_2\dots A_{11}$ be a non-convex $11$-gon such that
- The area of $A_iA_1A_{i+1}$ is $1$ for each $2 \le i \le 10$,
- $\cos(\angle A_iA_1A_{i+1})=\frac{12}{13}$ for each $2 \le i \le 10$,
- The perimeter of $A_1A_2\dots A_{11}$ is $20$.
If $A_1A_2+A_1A_{11}$ can be expressed as $\frac{m\sqrt{n}-p}{q}$ for positive integers $m,n,p,q$ with $n$ squarefree and $\gcd(m,p,q)=1$, find $m+n+p+q$.}
\end{questionbox}

\begin{R1box}
\textbf{\Large R1 Reasoning}

\vspace{1\baselineskip}

In \textbf{Steps 1–15}, let $|A_{1}A_{i}|=l_i,2\leq i\leq 11$, $l_2=a$, $l_3=b$. Based on the angle conditions and area constraints provided in the problem, R1 draws the following conclusions:
\begin{enumerate}
    \item $ab=\frac{26}{5}$.
    \item $l_2=l_4=\cdots=l_{10}=a$ and $l_3=l_5=\cdots=l_{11}=b$.
\end{enumerate}
In \textbf{Steps 16–89}, based on the edge length conditions, R1 derives the following:
\begin{align*}
    20 & = |A_1A_2|+(|A_2A_3|+|A_3A_4|+\cdots+|A_{9}A_{10}|)+|A_{10}A_{11}|\\
       & = a + \frac{26}{5a} + \sum_{i=1}^{9}|A_{i}A_{i+1}|
\end{align*}
let $x=a$, thus, based on ‌area relations‌ and the ‌cosine theorem‌, R1 derives the following conclusions:
\[
|A_{i}A_{i+1}|^2 = x^2 + \frac{676}{25x^2} - \frac{48}{5},2\leq i\leq9
\]
let $c=|A_{i}A_{i+1}|$, so $20=x+26/(5x)+9c$, so R1 writes:
\begin{align*}
    x+\frac{26}{5x}+9\times \sqrt{x^2+\frac{676}{25x^2}-\frac{48}{5}} = 20
\end{align*}
R1 denotes $t = x + 26/(5x)$, then
\begin{align*}
    t + 9\times \sqrt{t^2 - 20} = 20
\end{align*}
R1 denotes $s = \sqrt{t^2 - 20}$, then
\[ 
\left\{
\begin{aligned}
     & t^2=s^2+20 \\
     & t+9s=20 \\
\end{aligned}
\right.
\]

\end{R1box}
\begin{R1box}
R1 solves the above system of equations to obtain:
\[
t_1=\frac{-1-9\sqrt{5}}{4}<0\text{ (discard)},\quad t_2=\frac{9\sqrt{5}-1}{4}>0
\]
Since $t=x+26/(5x)$, so R1 has:
\[
20x^2 + (5 - 45\sqrt{5})x + 104 = 0
\]

In \textbf{Steps 90–131}, by applying the quadratic formula, R1 sets $a=20,b=5-45\sqrt{5},c=104$ and obtains:
\[
x=\frac{-b\pm\sqrt{\Delta}}{2a},\Delta=b^2-4ac=1830-450\sqrt{5}>0
\]

To find the square root of $\Delta$, R1 assumes there exist integers $A$ and $B$ such that:
\[
(\sqrt{A}-\sqrt{B})^2=A+B-2\sqrt{AB}=1830-450\sqrt{5}=1830-2\sqrt{253125}
\]
$A$ and $B$ are evidently two roots of the equation $x^2-1830x+253125=0$. However, the modified discriminant $\Delta^{'}=30\times(61-15\sqrt{5})$, and $A$ and $B$ obviously cannot be integer solutions. Thus, R1 abandons this method for computing $\sqrt{\Delta}$. R1 now attempts to solve for $x$ using numerical estimation, but this approach does not satisfy the answer specifications of the problem.

In \textbf{Steps 132–152}, R1 recalls that the problem requires calculating $|A_1A_2|+|A_1A_{11}|$, which precisely equals the value represented by $t$, thus arriving at the solution:
\[
|A_1A_2|+|A_1A_{11}|=t=\frac{ 9\sqrt{5}-1}{4}\Rightarrow m=9,n=5,p=1,q=4,m+n+p+q=\boxed{19}.
\]
R1 performs three independent verifications to rigorously confirm the validity of the solution.

\end{R1box}

\begin{wrongbox}
\textbf{\Large Failure Reasoning}

\vspace{1\baselineskip}

The failure in R1's reasoning process can be attributed to two main issues:

\textbf{1. Losing sight of the problem's objective, one proceeds with blind calculations.} R1 forgets that the problem requires finding $t$, thus dedicating the majority of the effort to solving for $x$ and producing substantial nonproductive computations. 

\textbf{2. Habitually uses numerical estimation for verification.} During numerical verification, $\sqrt{5}$ is often approximated as a decimal to verify the equality of the equations; this approach lacks mathematical rigor.

\end{wrongbox}
\begin{humanbox}
\textbf{\Large Human Reasoning}

\vspace{1\baselineskip}

Let $|A_1A_2|=a$ and $|A_1A_3|=b$. For $\cos(A_iA_1A_{i+1})=\frac{12}{13},2\leq i\leq10$, we have 
\[
\sin(\angle A_iA_1A_{i+1})=\frac{5}{13},2\leq i\leq10
\]
For $S_{\triangle A_iA_1A_{i+1}}=1,2\leq i\leq10$, we have
\begin{align*}
    \frac{1}{2}|A_{i}A_1|\cdot|A_1A_{i+1}|\cdot\sin\angle A_{i}A_1A_{i+1}=\frac{1}{2}|A_{i+1}A_{i+2}|\cdot|A_1A_{i+2}|\cdot\sin\angle A_{i+1}A_1A_{i+2}
\end{align*}
So, $|A_{1}A_{i}|=|A_1A_{i+2}|,2\leq i\leq9$, that is
\begin{align*}
    &|A_1A_2|=|A_1A_4|=\cdots|A_1A_{10}|=a\\
    &|A_1A_3|=|A_1A_5|=\cdots|A_1A_{11}|=b
\end{align*}
By sine area,
\[
\frac{1}{2}ab\cdot\frac{5}{13}=1\Rightarrow ab=\frac{26}{5}
\]
Using the law of cosines, the perimeter condition can be expressed as
\begin{align*}
    20 & = |A_1A_2|+(|A_2A_{3}|+|A_3A_4|\cdots+|A_8A_9|+|A_9A_{10}|)+|A_1A_{11}|\\
       & = |A_1A_2|+|A_1A_{11}|+9\sqrt{a^2+b^2-2ab\cdot\frac{12}{13}}\\
       & = a+b+9\sqrt{(a+b)^2-\frac{50}{13}ab}\\
       & = a+b+9\sqrt{(a+b)^2-20}
\end{align*}
Set $x=a+b$, we have
\[
(20-x)^2=81(x^2-20)\Rightarrow x=\frac{-1\pm 9\sqrt{5}}{4}
\]
Clearly $x$ is positive, so the final answer is
\[
x=\frac{ 9\sqrt{5}-1}{4}\Rightarrow m=9,n=5,p=1,q=4,m+n+p+q=\boxed{19}.
\]

\end{humanbox}

\newpage
\subsection{AIME\_1\_29}
\begin{questionbox}
\textit{Let ${\triangle ABC}$ be a right triangle with $\angle A = 90^\circ$ and $BC = 38.$ There exist points $K$ and $L$ inside the triangle such\[AK = AL = BK = CL = KL = 14.\]The area of the quadrilateral $BKLC$ can be expressed as $n\sqrt3$ for some positive integer $n.$ Find $n.$}
\end{questionbox}

\begin{R1box}
\textbf{\Large R1 Reasoning}\vspace{1\baselineskip}

In \textbf{Steps 1–5}, R1 begins by understanding the conditions given.

In \textbf{Steps 6–24}, R1 sets up an (x,y)-coordinate system and set the coordinates of points with unknowns \[A=(0,0),B=(c,0),C=(0,b),K=(x_1,y_1),L=(x_2,y_2)\]and lists the equations \[\begin{cases}b^2+c^2=38^2\\
x_1^2+y_1^2=14^2  &(1)\\
(x_1-c)^2+y_1^2=14^2  &(2)\\
x_2^2+y_2^2=14^2  &(3)\\
x_2^2+(y_2-b)^2=14^2  &(4)\\
(x_1-x_2)^2+(y_1-y_2)^2=14^2  &(5)
\end{cases}.\]

In \textbf{Steps 25–153}, R1 tries to solve the equations.

\textbf{Steps 25–39}, $(1)-(2)$ shows $x_1=c/2$ and $(3)-(4)$ shows $y_2=b/2$.

\textbf{Step 44}, R1 does a long step, trying to guess directly the value of b and c, losing completely the logic. It thinks wrongly that $660>784$, but then corrects it.

\textbf{Steps 45–50}, R1 substitutes the first time the values into $(5)$, but it thinks the equation \[(c/2-x_2)^2+(y_1-b/2)^2=14^2\] is too complicated.

\textbf{Steps 51–88}, R1 goes back to \textbf{Step 39} and substitutes the second time the values into $(5)$, this time R1 introduces $p=c/2$ and $q=b/2$, and the equations are simplified as \[\begin{cases}p^2+q^2=19^2\\
p\sqrt{14^2-q^2}+q\sqrt{14^2-p^2}=98
\end{cases}.\]

\textbf{Steps 89–110}, R1 tries to guess the answer if $p$ and $q$ are integers, which is useless.

\end{R1box}
\begin{R1box}
\textbf{Steps 111–128}, R1 tries to solve the simplified equation. It introduces $m=\sqrt{14^2-q^2}$ and $n=\sqrt{14^2-p^2}$, then it returns to the equations we had \[\begin{cases}m^2+q^2=14^2  &(1')\\
n^2+p^2=14^2  &(2')\\
pm+qn=98  &(3')\\
p^2+q^2=19^2  &(4')
\end{cases}.\]$(4')-(1')$, so $(p-m)(p+m)=165$; $(4')-(2')$, so $(q-n)(q+n)=165$.

\textbf{Steps 129–143}, R1 tries to guess the answer if $p$ and $m$ are integers, and concludes that $p=13$, $m=2$, $q=8\sqrt3$, $n=3\sqrt3$.

\textbf{Steps 144–151}, R1 gives all the coordinates \[B=(26,0),C=(0,16\sqrt3),K=(13,3\sqrt3),L=(2,8\sqrt3).\]

In \textbf{Steps 152–169}, R1 calculates the result by the shoelace formula: \begin{align}S_{BKLC}=&\frac12\left(x_By_K+x_Ky_L+x_Ly_C+x_Cy_B-x_Ky_B-x_Ly_K-x_Cy_L-x_By_C\right)\\
=&-104\sqrt3,\end{align}so the answer is $104$.

In \textbf{Steps 170–196}, R1 verifies first time the result by the same formula in different order of calculating, and the answer is the same.

In \textbf{Steps 197–211}, R1 verifies second time the result by a new way to calculate. At first it wants to break $BKLC$ into two triangles, the plan $BKL$ and $CKL$ is wrong, then the plan $BKL$ and $BLC$ is right, but it gives up and turns to substract $AKB$ and $ALC$ from $ABC$. At \textbf{Steps 205}, it finds that $AKL$ has been negletted, then it gives $S_{BKLC}=S_{ABC}-S_{AKB}-S_{ACL}-S_{AKL}=104\sqrt3$ and finally concludes.

\end{R1box}

\begin{wrongbox}
\textbf{\Large Failure Reasoning}\vspace{1\baselineskip}

R1 dosen't fail. But the the action to guess answer for the second equation system is lack of logic and dangerous.
\end{wrongbox}

\begin{humanbox}
\textbf{\Large Human Reasoning}\vspace{1\baselineskip}

A human approaching this problem would begin by reforming the logic of the graph. Consider an equilateral $\triangle AKL$ and $\angle BAK=\alpha$, $\angle CAL=\frac\pi6-\alpha$ such that $AK=BK$, $AL=CL$. Note $|AK|=a$, $|BC|=2b$, then by the Pythagorean theorem \begin{align*}b^2=&\left(a\cos\alpha\right)^2+\left(a\cos(\frac\pi6-\alpha)\right)^2\\
=&\left(\frac{1+\cos2\alpha}2+\frac{1+\cos(\frac\pi3-2\alpha)}2\right)a^2\\
=&\left(1+\cos\frac\pi6\cos(\frac\pi6-2\alpha)\right)a^2,
\end{align*}with the value provided, it's just
\[\cos(\frac\pi6-2\alpha)a^2=\frac{2(b-a)(b+a)}3\sqrt3.\]
While \[S_{\triangle ABC}=2\cdot a\cos\alpha\cdot a\cos(\frac\pi6-\alpha)=\left(\cos\frac\pi6+\cos(\frac\pi6-2\alpha)\right)a^2,\]
\[S_{\triangle AKB}=\frac12\sin(2\alpha)a^2,\]
\[S_{\triangle ALC}=\frac12\sin(\frac\pi3-2\alpha)a^2\]
and so that \begin{align}S_{BKLC}&=S_{\triangle ABC}-S_{\triangle AKB}-S_{\triangle ALC}-S_{\triangle AKL}\\
&=\left(\cos\frac\pi6+\cos(\frac\pi6-2\alpha)-\frac12\sin(2\alpha)-\frac12\sin(\frac\pi3-2\alpha)-\frac12\sin\frac\pi3\right)a^2\\
&=\left(\frac{\sqrt3}4+\cos(\frac\pi6-2\alpha)-\sin\frac\pi6\cos(\frac\pi6-2\alpha)\right)a^2\\
&=\frac14\left(\sqrt3+2\cos(\frac\pi6-2\alpha)\right)a^2\\
&=\left(\frac{a^2}4+\frac{(b-a)(b+a)}3\right)\sqrt3.\end{align}
We conclude then $n=\frac{a^2}4+\frac{(b-a)(b+a)}3=104$ since $a=14$ and $b=38/2=19$.
\end{humanbox}

\newpage
\subsection{AIME\_1\_30}
\begin{questionbox}
\textit{Let
\[f(x)=\frac{(x-18)(x-72)(x-98)(x-k)}{x}.\]There exist exactly three positive real values of $k$ such that $f$ has a minimum at exactly two real values of $x$. Find the sum of these three values of $k$.}
\end{questionbox}

\begin{R1box}
\textbf{\Large R1 Reasoning}\vspace{1\baselineskip}

In \textbf{Steps 1–10}, R1 begins by understanding the meaning of the problem.

In \textbf{Steps 11–61}, R1 makes a plan to analyse the derivative $f'$, but it's ugly, so R1 turns to calculate the logarithm \[\frac{f'(x)}{f(x)}=\frac{1}{x-18}+\frac{1}{x-72}+\frac{1}{x-98}+\frac{1}{x-k}-\frac{1}x.\]So the two minimum $x_1$, $x_2$ of $f$ satisfying the equation \[S(x):=\frac{1}{x-18}+\frac{1}{x-72}+\frac{1}{x-98}+\frac{1}{x-k}-\frac{1}x=0\] since the zero point of $f$ are evidently not minimal.

In \textbf{Steps 62–141}, R1 tries to solve the equation $S=0$ and design plural times the plan. First practiced plan rewrites $S=0$ as \[3x^4 - 2(k + 188)x^3 + (188k + 10116)x^2 - 127008k = 0\] and supposes that this equation can be factored as \[(x^2 + ax + b)(3x^2 + cx + d)=0.\] Comparing the coefficients, and it gives \[\begin{cases}\;\,c + 3a  = -2(k + 188),(1)\\
d + a c + 3b = 188k + 10116,(2)\\
a d +b c = 0,(3)\\
b d = -127008k.(4)\end{cases}\]
Second practiced plan is collecting the constraints on $x_1$ and $x_2$ that \[\begin{cases}\frac1{x_1 - 18} + \frac1{x_1 - 72} + \frac1{x_1 - 98} + \frac1{x_1 - k} = \frac1{x_1},\\
\frac1{x_2 - 18} + \frac1{x_2 - 72} + \frac1{x_2 - 98} + \frac1{x_2 - k} = \frac1{x_2},\\
\frac{(x_1 - 18)(x_1 - 72)(x_1 - 98)(x_1 - k)}{x_1}=\frac{(x_2 - 18)(x_2 - 72)(x_2 - 98)(x_2 - k)}{x_2}.\end{cases}\]
But ther are all given up because not concrete enough.

\end{R1box}
\begin{R1box}
In \textbf{Steps 62–179}, R1 repeats and combines the plans raised above. In \textbf{Steps 180–216}, R1 wrongly transforms the equation $S=0$ to \[\frac{18}{x - 18} + \frac{72}{x - 72} + \frac{98}{x - 98} + \frac k{x - k} = 0.\] But it's still useless. In \textbf{Steps 217–225}, R1 begins to guess $x_1x_2=18\times98$. Evidently it's helpless. In \textbf{Steps 226–268}, R1 expands the wrong edition of $S=0$ and gives up.

In \textbf{Steps 269–289}, R1 returns to $f$ itself. R1 guesses that $x_1x_2=18\times72\times98\times k$ and gives up. In \textbf{Steps 290–295}, R1 returns again to $f$ itself.

In \textbf{Steps 296–299}, R1 sets $C$ the minimum and gives \[(x-18)(x-72)(x-98)(x-k) - C x = (x - x1)^2 (x - x2)^2,\]where $x_1$, $x_2$ are the two only minimal points. In \textbf{Steps 300–334}, R1 expands it and compares the coefficients \[\begin{cases}k + 188 = 2(x_1 + x_2),\\
188k + 10116 = x_1^2 + 4x_1x_2 + x_2^2,\\
-(10116k + 127008 + C) = -2x_1x_2(x_1 + x_2),\\
127008k = x_1^2x_2^2.\end{cases}\]
with the constraints for $x_1$, $x_2$ that $S=0$ i.e. \[3x^4 - 2(k + 188)x^3 + (188k + 10116)x^2 - 127008k = 0.\]
\textbf{Steps 335–451}, R1 substituts $x_1x_2$ and $x_1+x_2$ in the equations, it leaves on an equation for $k$ \[188k + 10116 = [(k + 188)/2]^2 + 2*sqrt(127008k)\] equivalent to \[t^4 - 752 t^2 + 8064 t - 20480 = 0\] where $t=\sqrt{2k}$, it factors then \[0=(t - 4)(t - 8)(t - 20)(t + 32),\] so $t=4$, $8$, $20$ and then $k=8$, $32$, $200$. R1 concludes that the result is $8+32+200=240$.
\end{R1box}

\begin{wrongbox}
\textbf{\Large Failure Reasoning}\vspace{1\baselineskip}

The failure in R1's reasoning process can be attributed to three main issues: meaningless guessing and wrong calculation. But R1 gives up them quickly and finally find the correct method to solve the problem.

\end{wrongbox}

\begin{humanbox}
\textbf{\Large Human Reasoning}\vspace{1\baselineskip}

A human approaching this problem would begin by considering a new polynomial \[F(x)=xf(x)-mx,\]where $m=\min\limits_{x>0}f(x)$. Then $F\geq0$ on $\mathbb R_+$ and the equality is reached if and only if $x=x_1$ or $x_2$. $x_1$ and $x_2$ are minimal implies that $(x-x_1)^2|F(x)$ and $(x-x_2)^2|F(x)$, so $F(x)=(x-x_1)^2(x-x_2)^2$ since $F$ is unitary.

Then the equation \[(x-x_1)^2(x-x_2)^2+mx=0\] admits $4$ roots $18$, $72$, $98$ and $k$. Let $t=\sqrt x$, then the equations \begin{align}(t^2-x_1)(t^2-x_2)=&\sqrt{-m}t\\
(t^2-x_1)(t^2-x_2)=&-\sqrt{-m}t\end{align} admit respectively roots as quadruple \[(i_1\sqrt{18},i_2\sqrt{72},i_3\sqrt{98},i_4\sqrt k)\] and \[(j_1\sqrt{18},j_2\sqrt{72},j_3\sqrt{98},j_4\sqrt k),\] where $\{i_n,j_n\}=\{-1,1\}$ for any $1\leq n\leq4$.

So $k$ is bound to be included in the set \[\{(\pm\sqrt{18}\pm\sqrt{72}\pm\sqrt{98})^2\}=\{8,32,200,512\}.\] We claim that $512$ is impossible because \[0<x_1x_2=-\sqrt{18}\sqrt{72}\sqrt{98}\sqrt{k}<0.\]
We conclude that $k=8$, $32$ or $200$ and the sum is $240$.

\end{humanbox}

\newpage
\section{Logic Graph for R1 COT}
\label{sec:logicgraph}
\begin{figure}[htbp]
    \centering
    \includegraphics[width=0.65\linewidth]{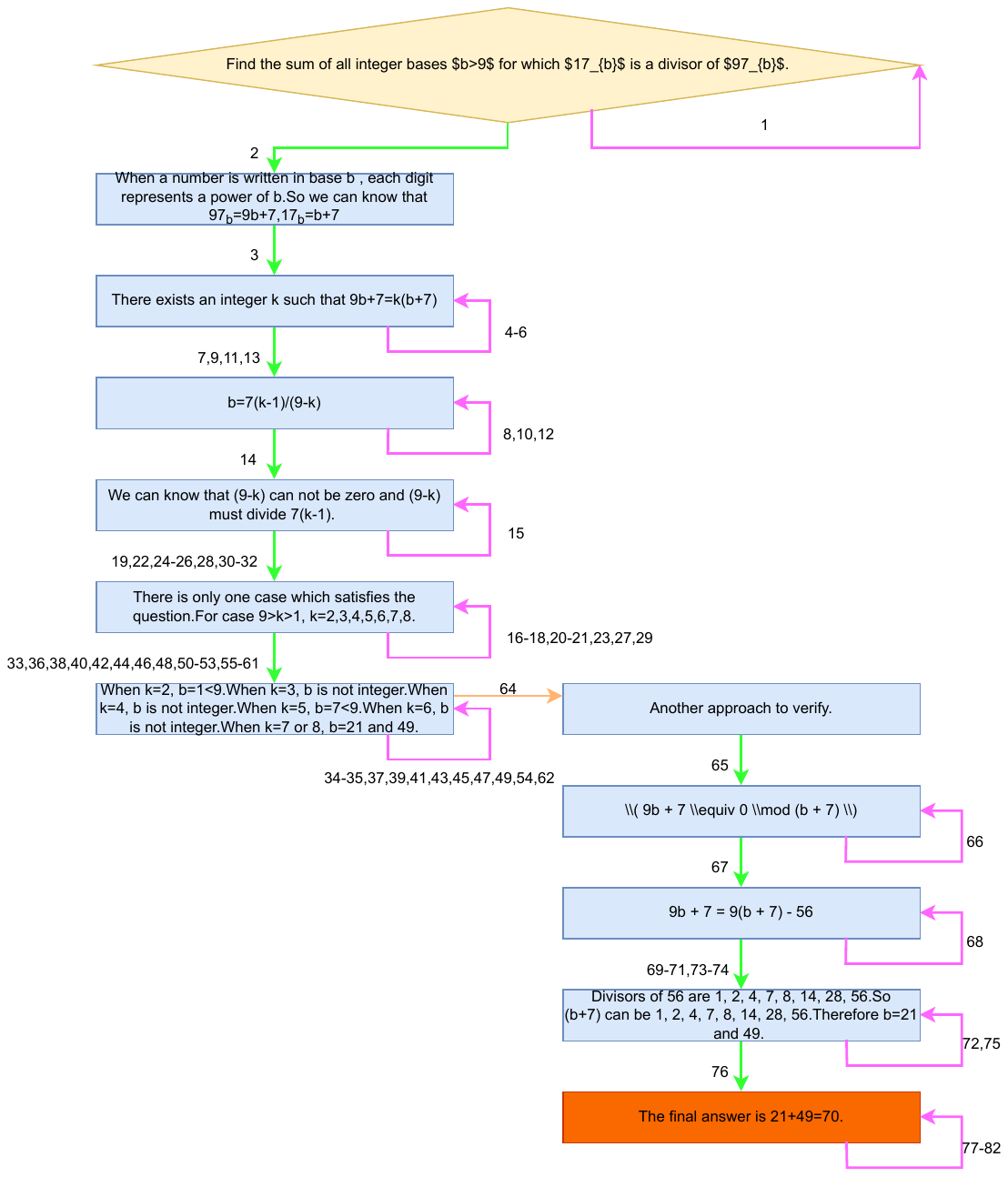}
    \caption{Logic Graph of Case 1}
    \label{fig:AIME_1}
\end{figure}

\begin{figure}[htbp]
    \centering
    \includegraphics[width=0.8\linewidth]{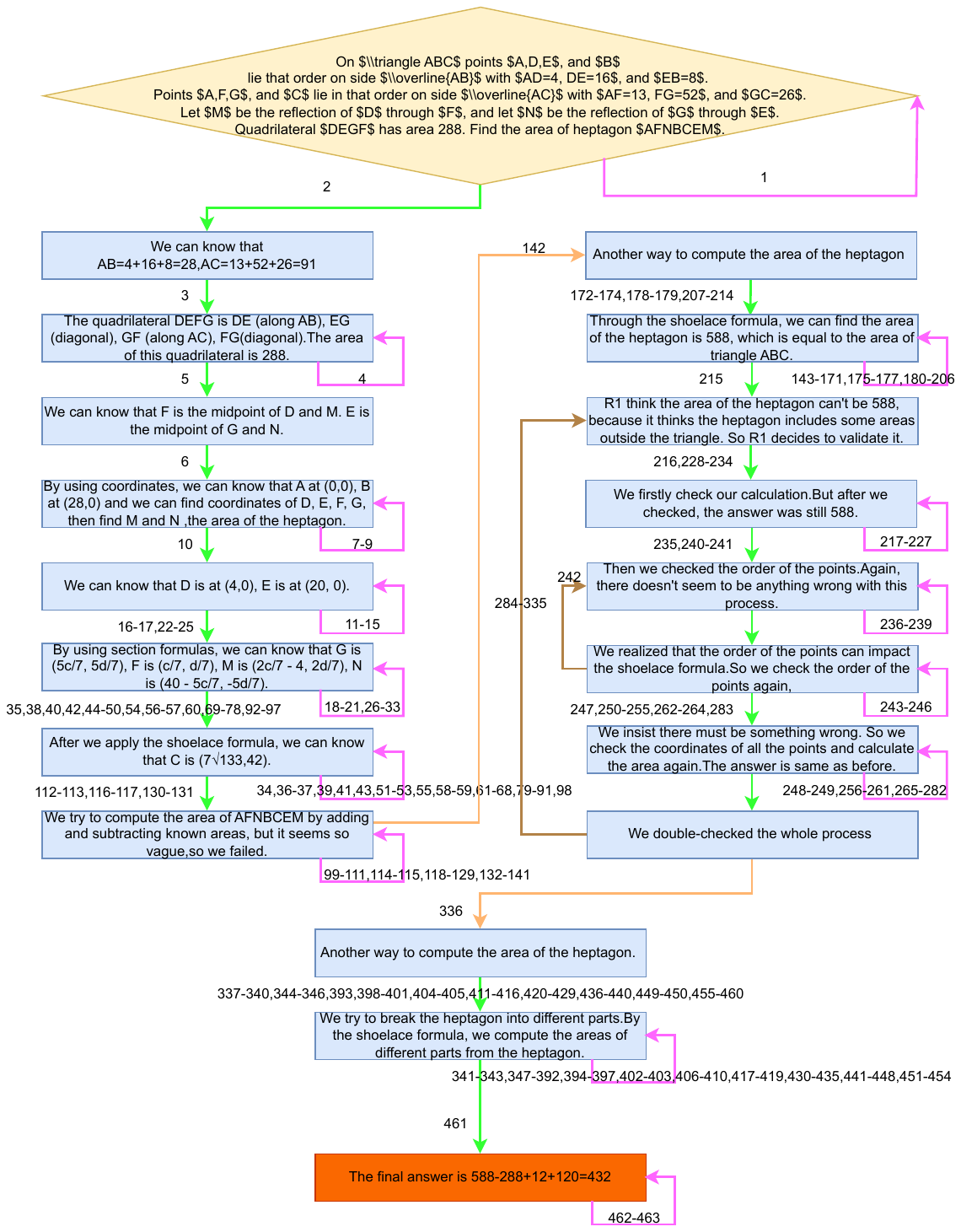}
    \caption{Logic Graph of Case 2}
    \label{fig:AIME_2}
\end{figure}

\begin{figure}[htbp]
    \centering
    \includegraphics[width=0.8\linewidth]{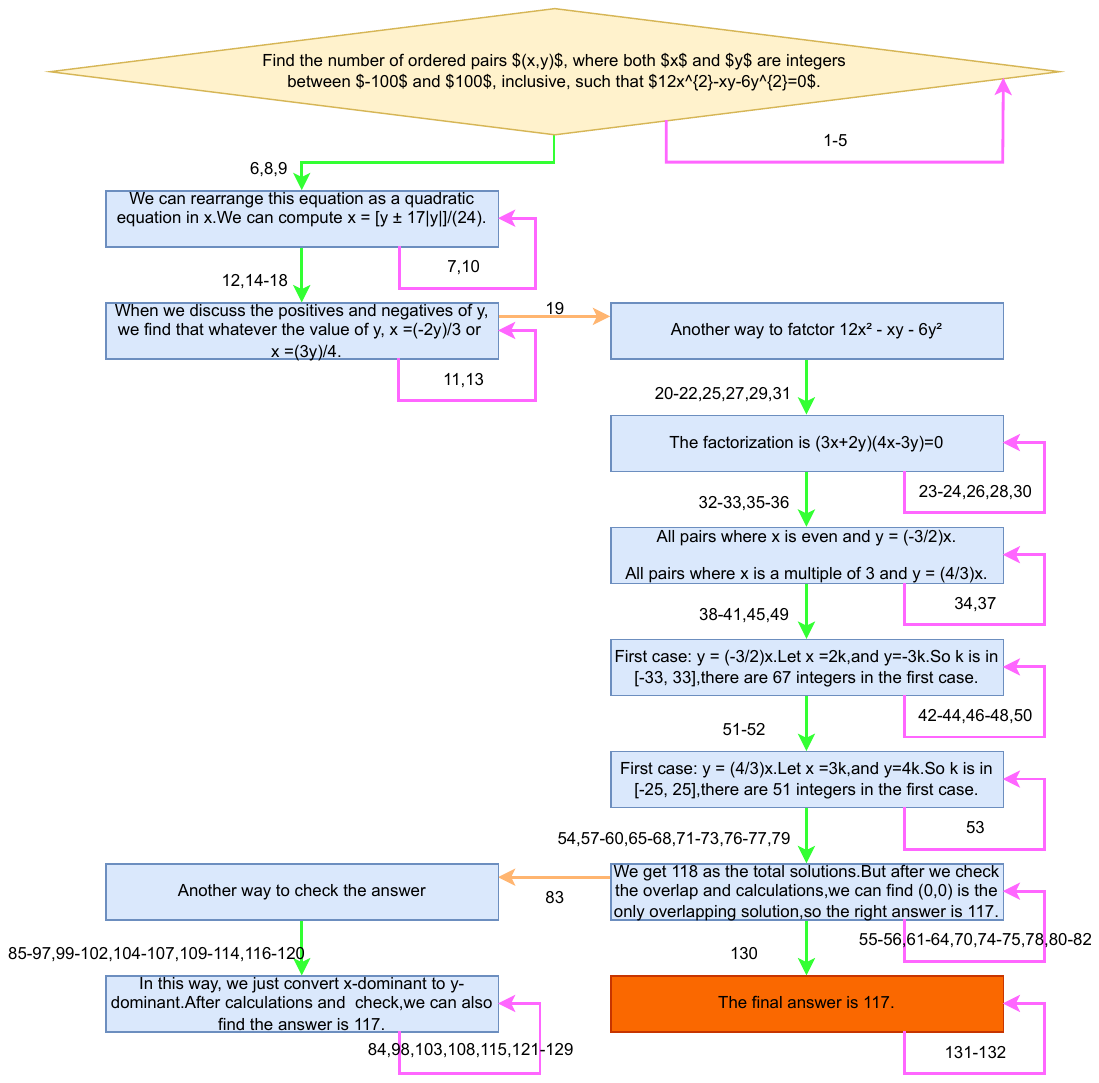}
    \caption{Logic Graph of Case 4}
    \label{fig:AIME_4}
\end{figure}

\begin{figure}[htbp]
    \centering
    \includegraphics[width=0.8\linewidth]{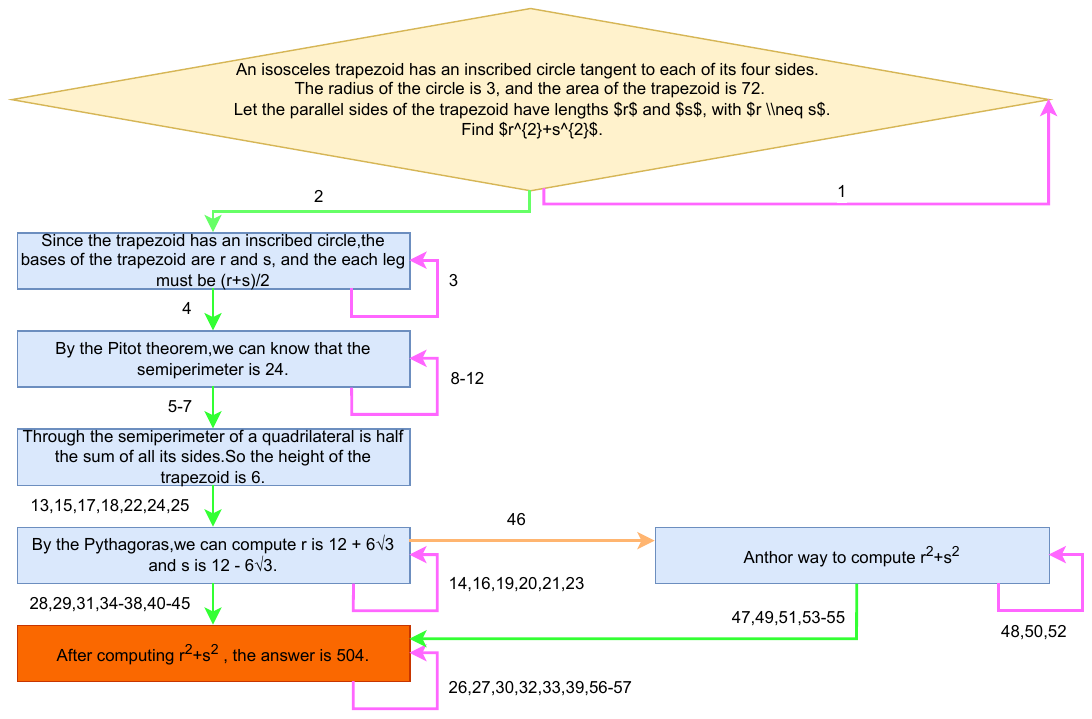}
    \caption{Logic Graph of Case 7}
    \label{fig:AIME_7}
\end{figure}

\begin{figure}[htbp]
    \centering
    \includegraphics[width=0.8\linewidth]{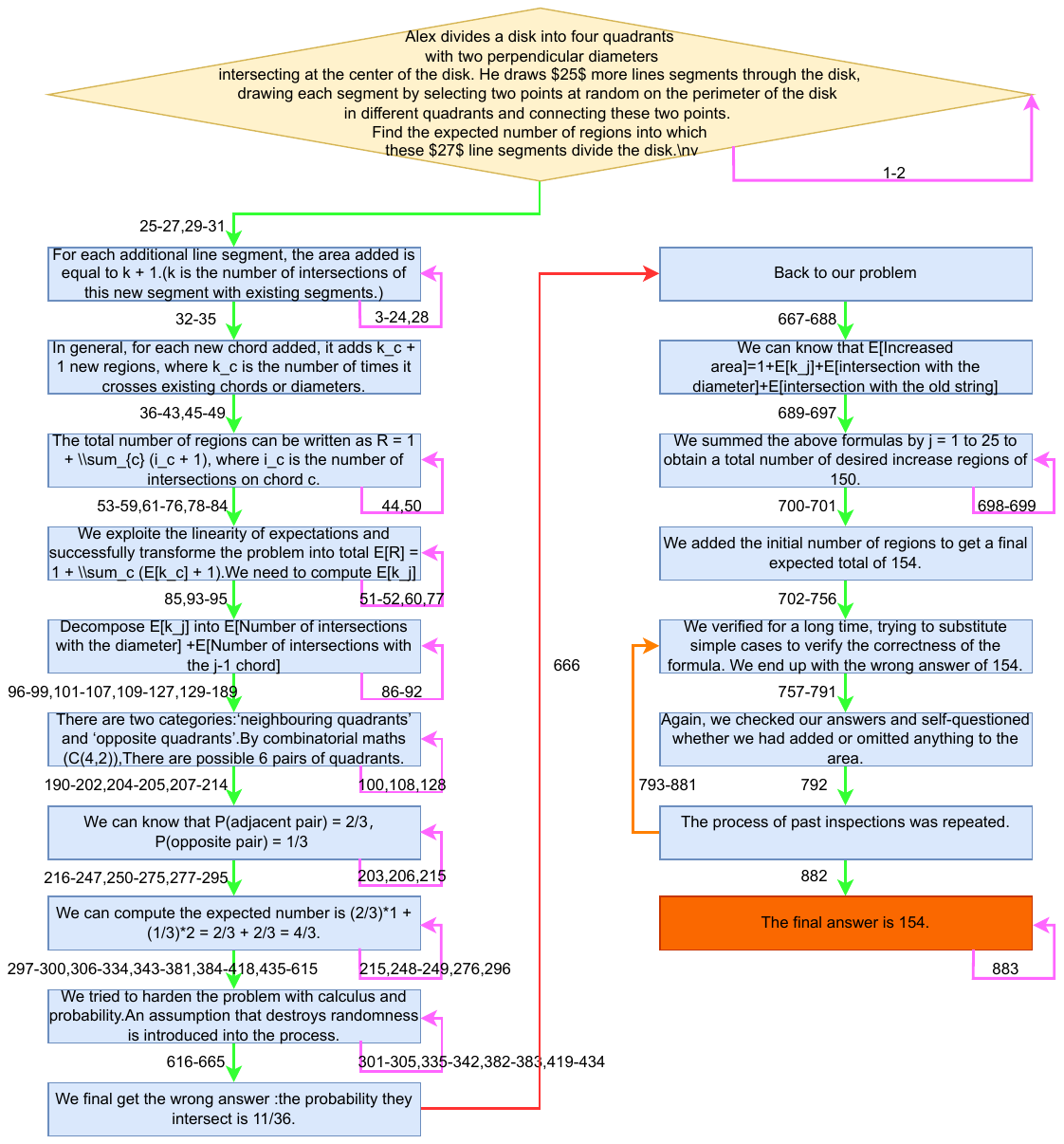}
    \caption{Logic Graph of Case 13}
    \label{fig:AIME_13}
\end{figure}

\begin{figure}[htbp]
    \centering
    \includegraphics[width=0.8\linewidth]{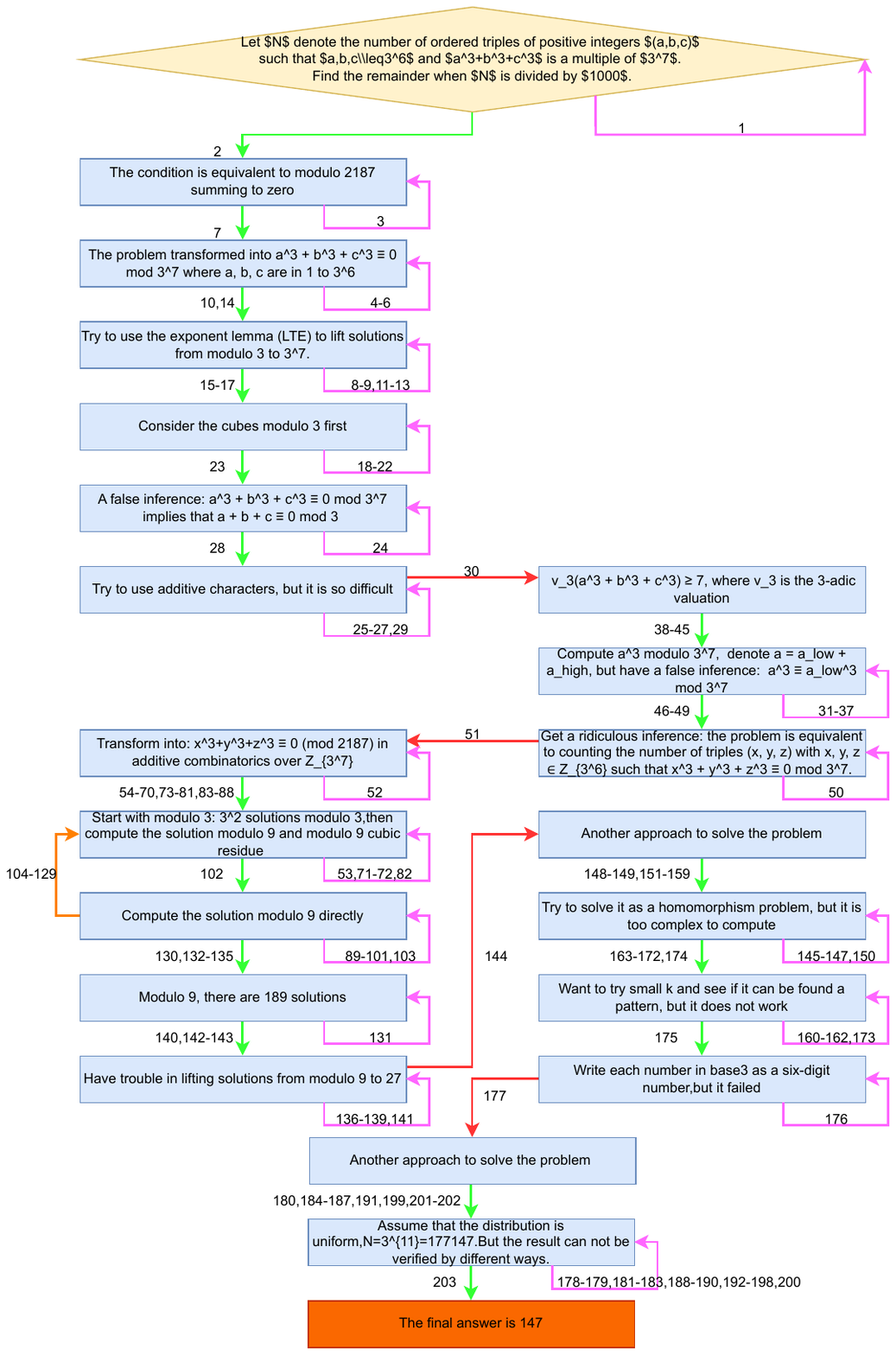}
    \caption{Logic Graph of Case 15}
    \label{fig:AIME_15}
\end{figure}

\begin{figure}[htbp]
    \centering
    \includegraphics[width=0.8\linewidth]{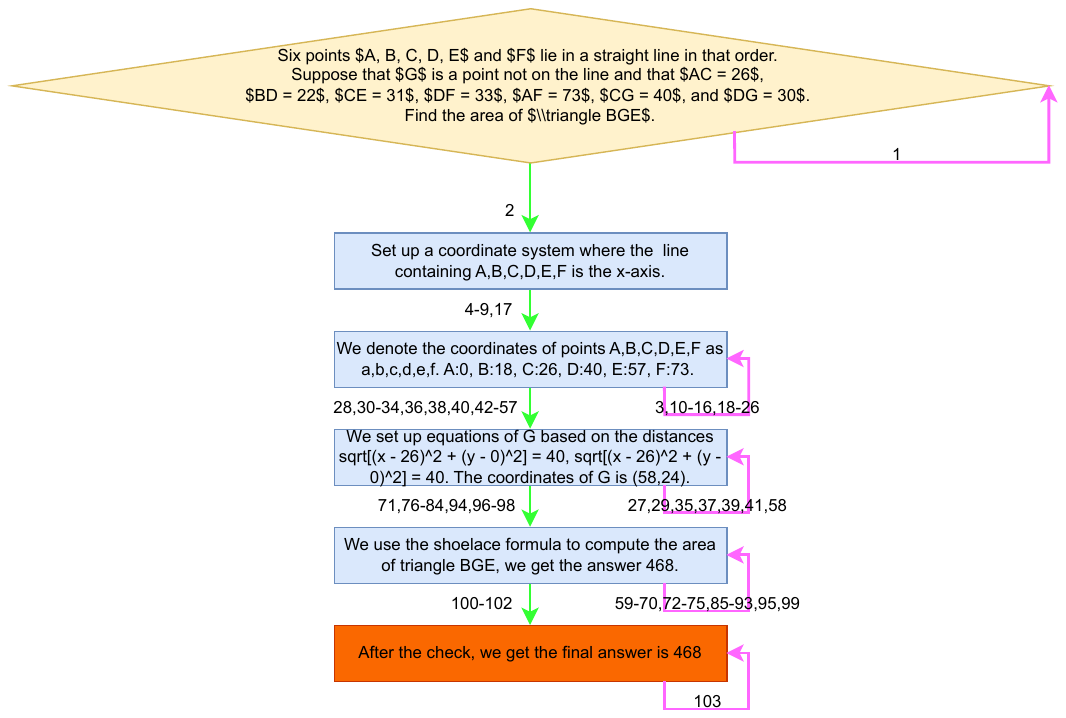}
    \caption{Logic Graph of Case 16}
    \label{fig:AIME_16}
\end{figure}

\begin{figure}[htbp]
    \centering
    \includegraphics[width=0.8\linewidth]{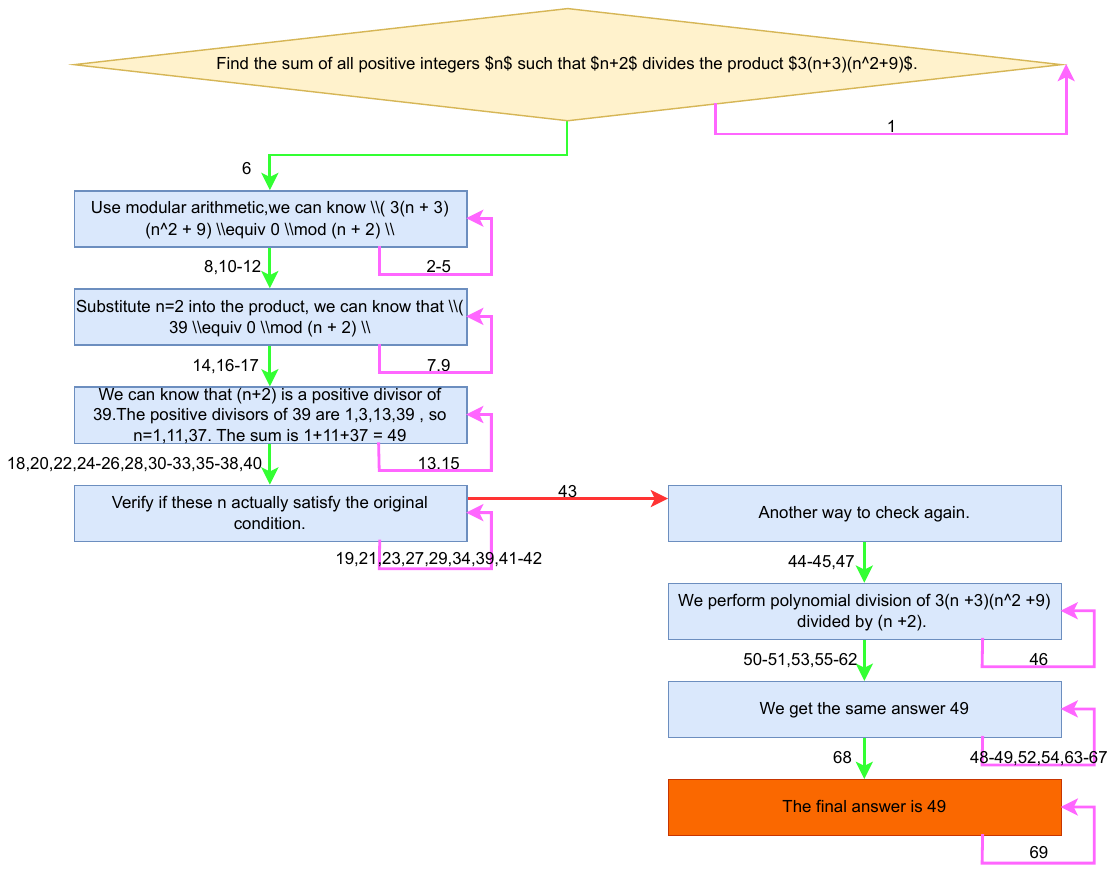}
    \caption{Logic Graph of Case 17}
    \label{fig:AIME_17}
\end{figure}

\begin{figure}[htbp]
    \centering
    \includegraphics[width=0.8\linewidth]{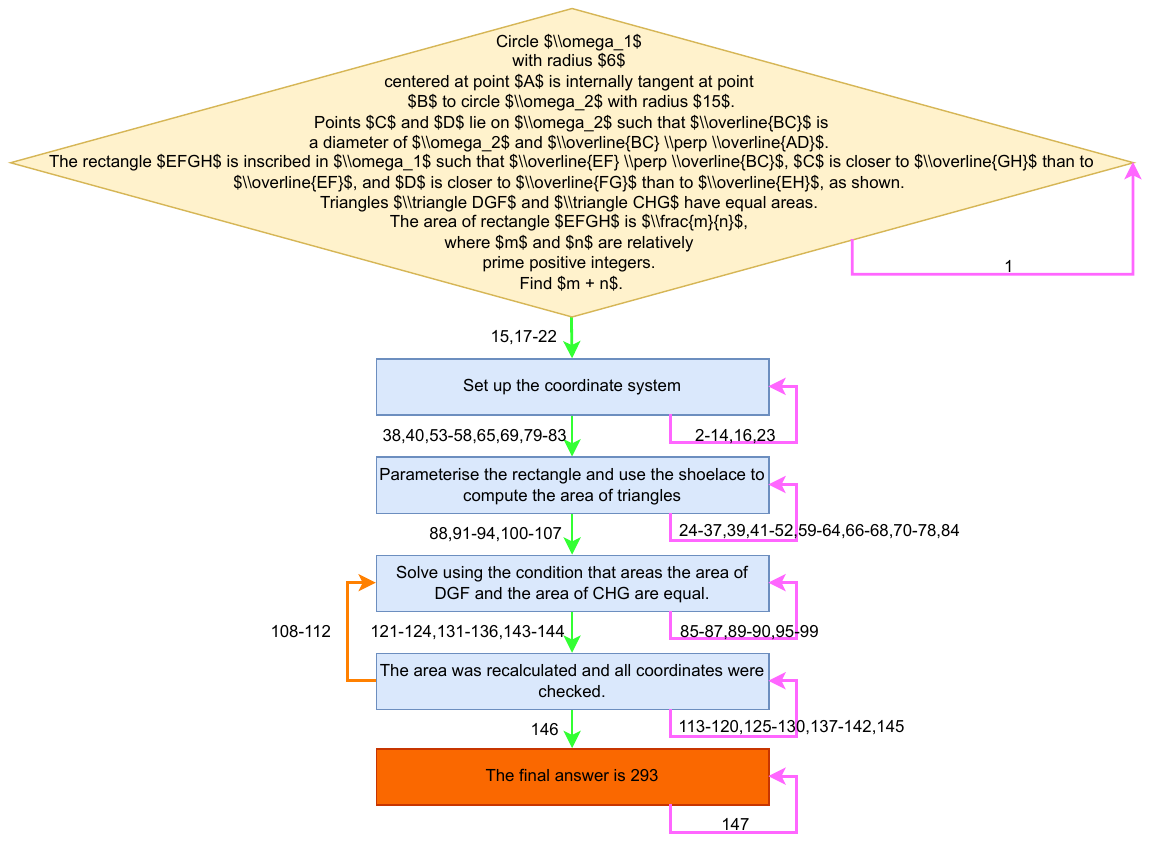}
    \caption{Logic Graph of Case 21}
    \label{fig:AIME_21}
\end{figure}

\begin{figure}[htbp]
    \centering
    \includegraphics[width=0.8\linewidth]{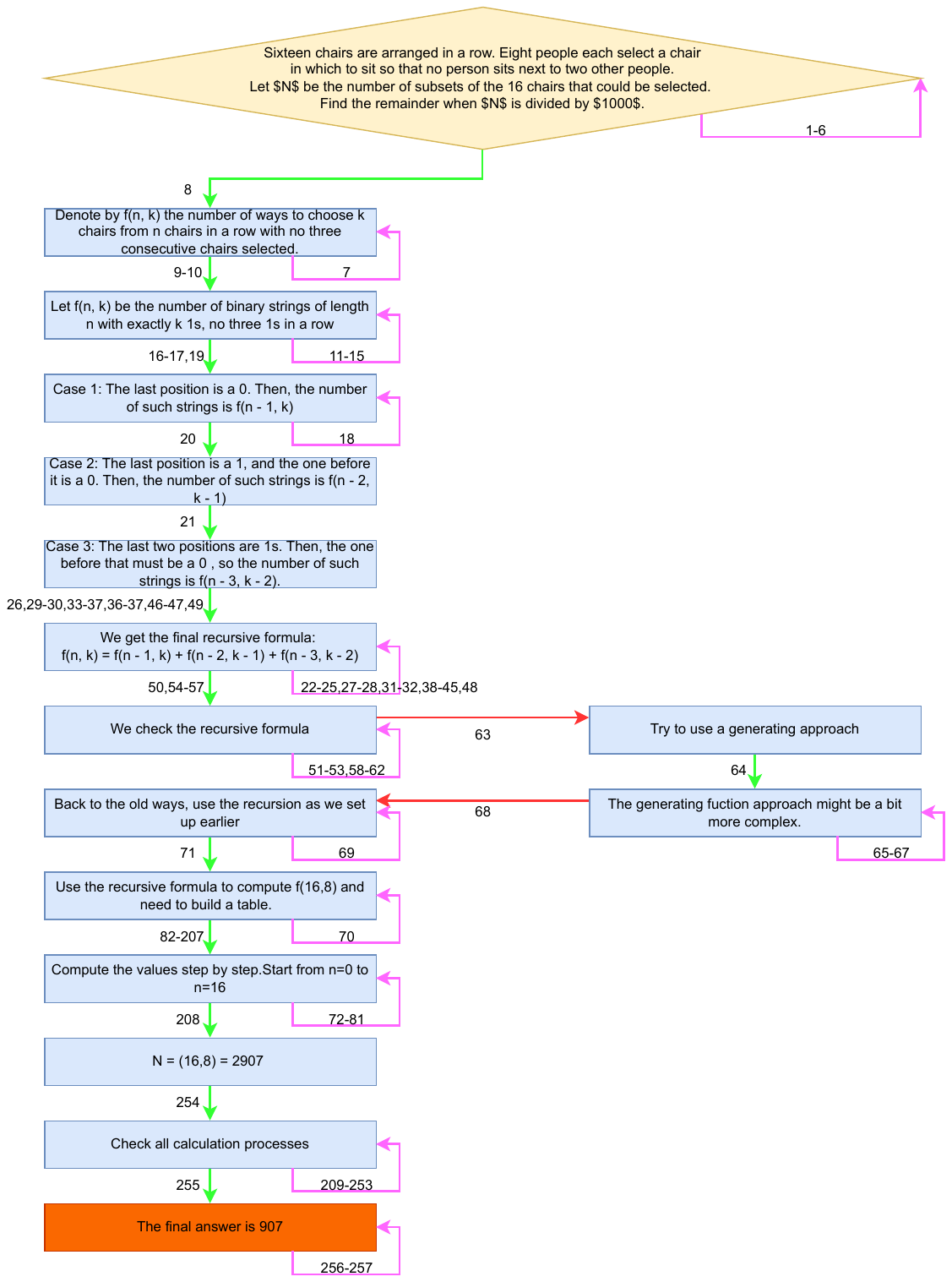}
    \caption{Logic Graph of Case 25}
    \label{fig:AIME_25}
\end{figure}

\begin{figure}[htbp]
    \centering
    \includegraphics[width=0.8\linewidth]{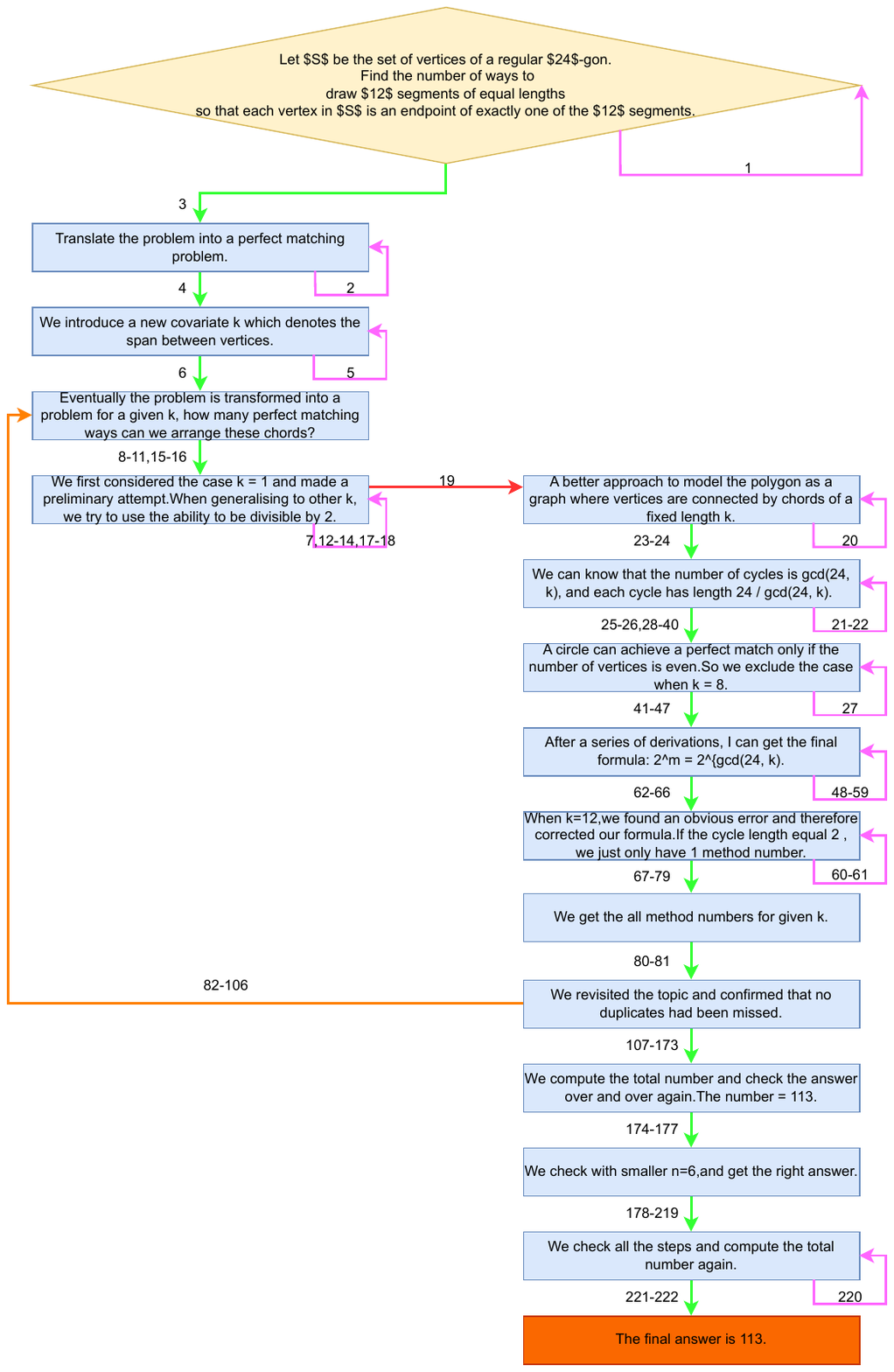}
    \caption{Logic Graph of Case 26}
    \label{fig:AIME_26}
\end{figure}

\begin{figure}[htbp]
    \centering
    \includegraphics[width=0.8\linewidth]{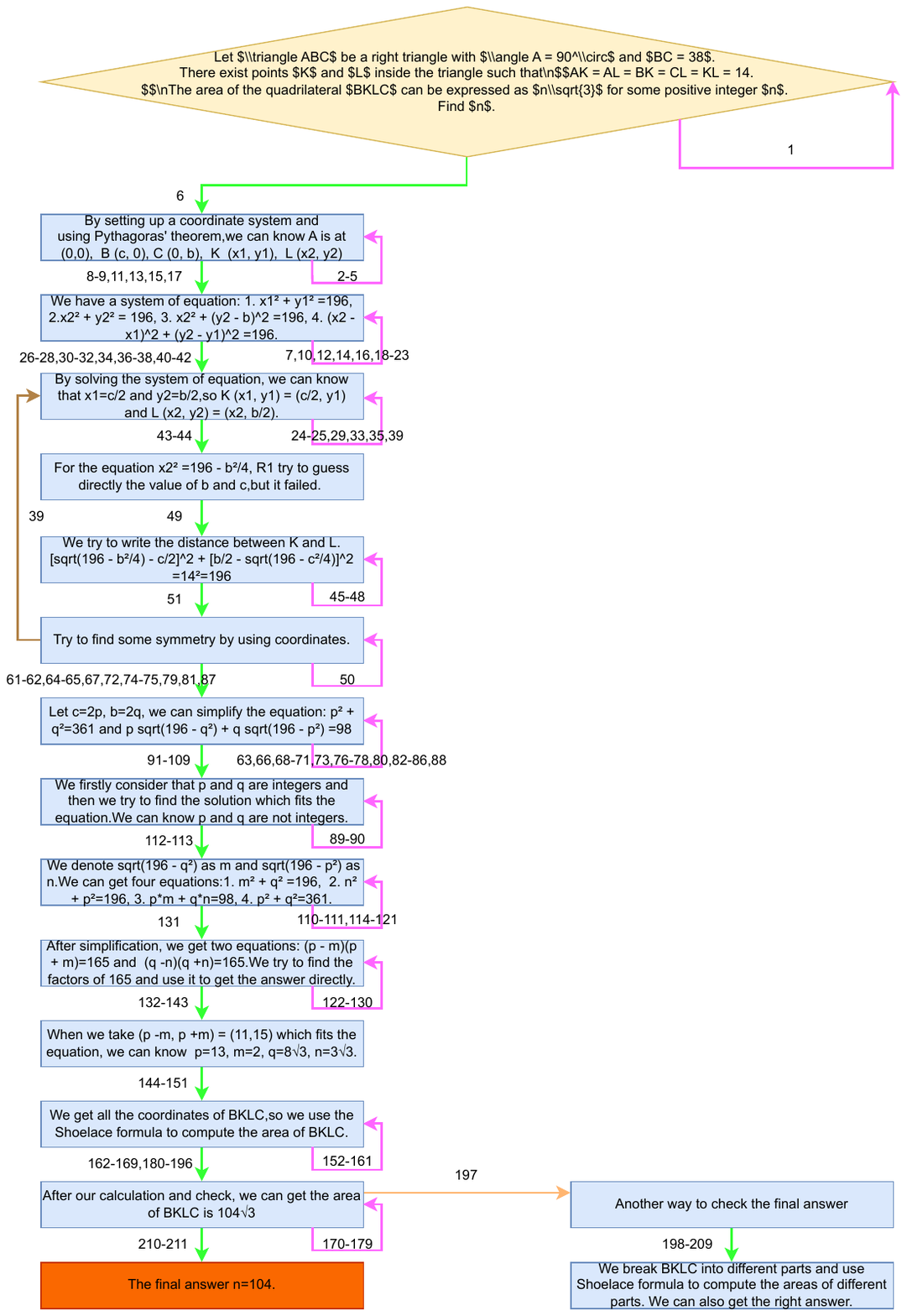}
    \caption{Logic Graph of Case 29}
    \label{fig:AIME_29}
\end{figure}

\begin{figure}[htbp]
    \centering
    \includegraphics[width=0.8\linewidth]{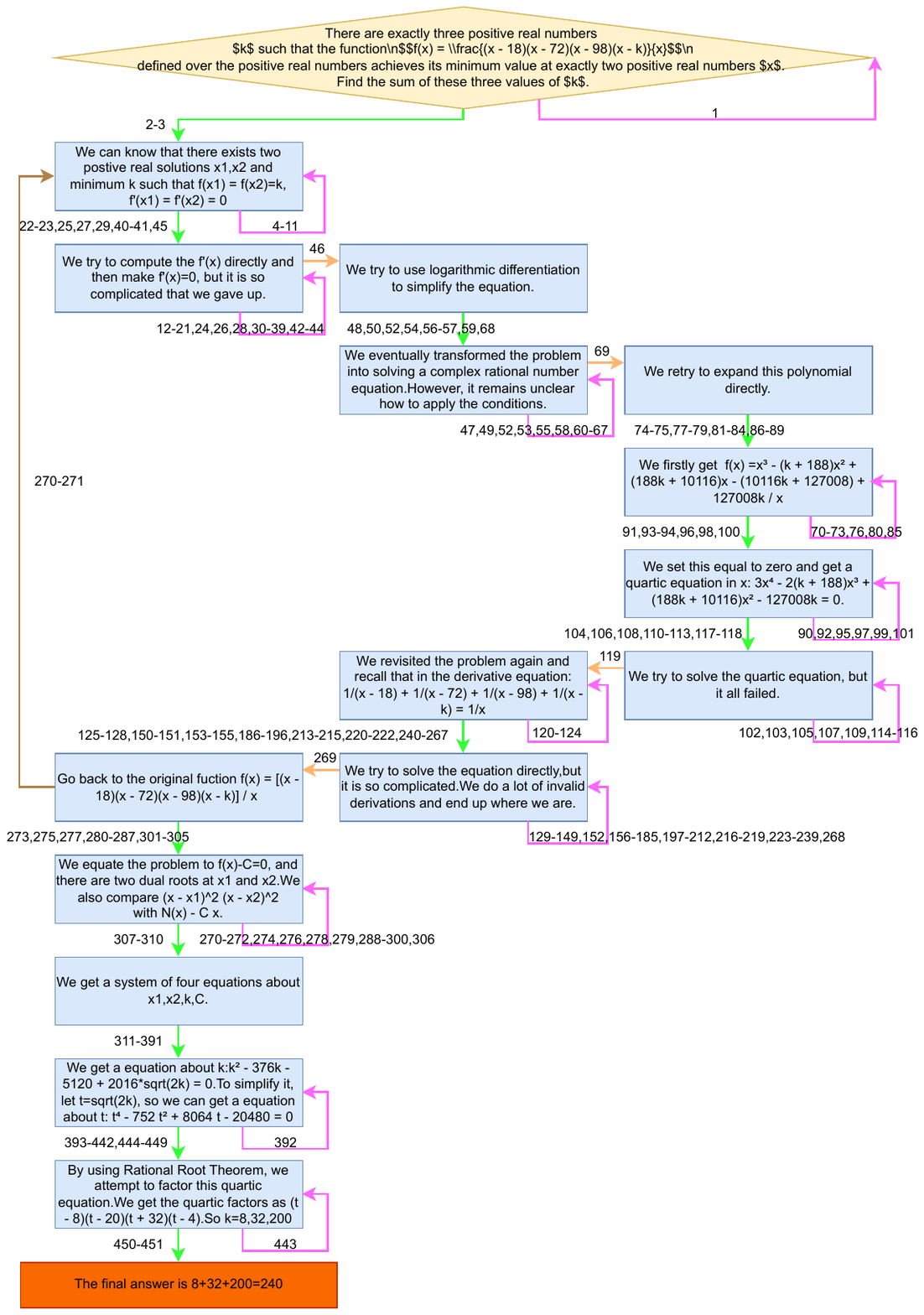}
    \caption{Logic Graph of Case 30}
    \label{fig:AIME_30}
\end{figure}

\end{document}